\documentclass{article}

\usepackage{hyperref}
\usepackage{url}
\usepackage{microtype}
\usepackage{booktabs} 
\usepackage{subcaption}
\usepackage{graphicx}
\usepackage{algorithm}
\usepackage{algorithmic}
\usepackage{textcomp}
\usepackage{xfrac}
\usepackage{xspace}
\usepackage{mathtools}
\usepackage{multirow}
\usepackage{wrapfig}
\usepackage{url}
\usepackage{xcolor}
\usepackage{enumitem}
\usepackage{capt-of}
\usepackage{booktabs}
\usepackage{subcaption}
\usepackage{calrsfs}
\usepackage{amsfonts}

\newcommand{\system}{\textsc{\mbox{CHAI}}\xspace}
\newcommand{\dejavu}{\textsc{\mbox{DejaVu}}\xspace}
\newcommand{\spatten}{\textsc{\mbox{SpAtten}}\xspace}
\newcommand{\llama}{\textsc{\mbox{LLaMa}}\xspace}
\newcommand{\opt}{\textsc{OPT}\xspace}

\usepackage{xspace}
\usepackage{soul}
\usepackage[noabbrev]{cleveref}

\newcommand{\eg}{{\it e.g.}, }
\newcommand{\ie}{{\it i.e.}, }
\newcommand{\real}{\mathbb{R}}

\newcommand{\weight}{\mathbf{W}}
\newcommand{\key}{\mathbf{K}}
\newcommand{\query}{\mathbf{Q}}
\newcommand{\valuev}{\mathbf{V}}
\DeclarePairedDelimiter{\opair}{\langle}{\rangle}

\newenvironment{myitemizeleft}
{
   \vspace{0pt}
    \begin{list}{$\bullet$ }{\leftmargin=1em \itemindent=0em}
        \setlength{\topsep}{0em}
        \setlength{\parskip}{0pt}
        \setlength{\partopsep}{0pt}
        \setlength{\parsep}{0pt}
        \setlength{\itemsep}{0.3mm}
}
{
    \end{list}
}

\usepackage{graphicx}
\graphicspath{ {./images/} }

\usepackage{bigstrut}
\newsavebox{\algleft}
\newsavebox{\algright}
\usepackage{pifont}

\usepackage[accepted]{icml2023}

\icmltitlerunning{Runtime Clustered Head Attention for Efficient Inference}
\begin{document}

%

\vskip 0.3in

\twocolumn[
\icmltitle{\system: Clustered Head Attention for Efficient LLM Inference}








\icmlsetsymbol{equal}{*}

\begin{icmlauthorlist}
\icmlauthor{Saurabh Agarwal}{wisc}
\icmlauthor{Bilge Acun}{meta}
\icmlauthor{Basil Hosmer}{meta}
\icmlauthor{Mostafa Elhoushi}{meta}
\icmlauthor{Yejin Lee}{meta}
\icmlauthor{Shivaram Venkataraman}{wisc}
\icmlauthor{Dimitris Papailiopoulos}{wisc}
\icmlauthor{Carole-Jean Wu}{meta}
\end{icmlauthorlist}

\icmlaffiliation{wisc}{University of Wisconsin-Madison}
\icmlaffiliation{meta}{Meta-FAIR}

\icmlcorrespondingauthor{Saurabh Agarwal}{agarwal@cs.wisc.edu}

\vspace{0.3 in}
]
\printAffiliationsAndNotice{}
\begin{abstract}
    Large Language Models (LLMs) with hundreds of billions of parameters have transformed the field of machine learning. However, serving these models at inference time 
    is both compute and memory intensive, where a single request can require multiple GPUs and tens of Gigabytes of memory. 
    Multi-Head Attention is one of the key components of LLMs,  which can account for over 50\% of LLMs memory and compute requirement. We observe that there is a high amount of redundancy across heads on which tokens they pay attention to. Based on this insight, we propose Clustered Head Attention (\system). \system combines 
    heads with a high amount of correlation for self-attention at runtime, thus reducing both memory and compute.
    In our experiments, we show that \system is able to reduce the memory requirements for storing K,V cache by up to 21.4\% and inference time latency by up to $1.73\times$ without any fine-tuning required. \system achieves this with a maximum 3.2\% deviation in accuracy across 3 different models (i.e. \opt-66B, \llama-7B, \llama-33B) and 5 different evaluation datasets.
\end{abstract}

\section{Introduction}

LLMs have demonstrated remarkable performance on 
language modelling tasks ranging from question answering, text summarizing, language translation. 
However, such performance has been achieved by scaling models to trillions of parameters, and existing works~\cite{hoffmann2022training,touvron2023llama, kaplan2020scaling} show that increasing the model size may lead to even higher model quality.

Inference on LLMs introduce several new challenges. Beyond just the quadratic computation cost of self-attention~\cite{vaswani2017attention} with increasing context and large model sizes, LLMs also store intermediate Key (K) and Value (V) pairs for subsequent next word prediction. This K,V caching introduces additional memory related challenges as K,V cache size increases with increase in sequence length. 
The architecture of widely used LLMs like GPT~\cite{brown2020language} and \llama~\cite{touvron2023llama, touvron2023llama2} use Multi-Head Attention (MHA)~\cite{vaswani2017attention}. MHA uses several attention heads to look at a sequence. As models grow bigger, the number of heads increases as well. For example, \llama-7B uses 32 attention heads in each layer, while \llama-65B uses 64 attention heads per layer~\cite{touvron2023llama}.
The use of MHA exacerbates bottlenecks for serving LLMs. 
First, it increases compute pressure due to repeated application of the attention operation. Second, it increases the memory pressure due to requiring storage of Key (K), Value (V) caches that comes with the additional attention heads.
To alleviate these bottlenecks, prior works have introduced primarily two types of methods - (i) pruning of LLMs to utilize sparsity based on the input context~\cite{liu2023deja, voita2019analyzing} and (ii) Co-designing of the Attention module to reuse components across multiple heads like Multi-Query Attention (MQA)~\cite{shazeer2019fast} and Grouped-Query Attention (GQA)~\cite{ainslie2023gqa}.

\begin{figure}[t]
    \centering
    \includegraphics[width=0.8\linewidth]{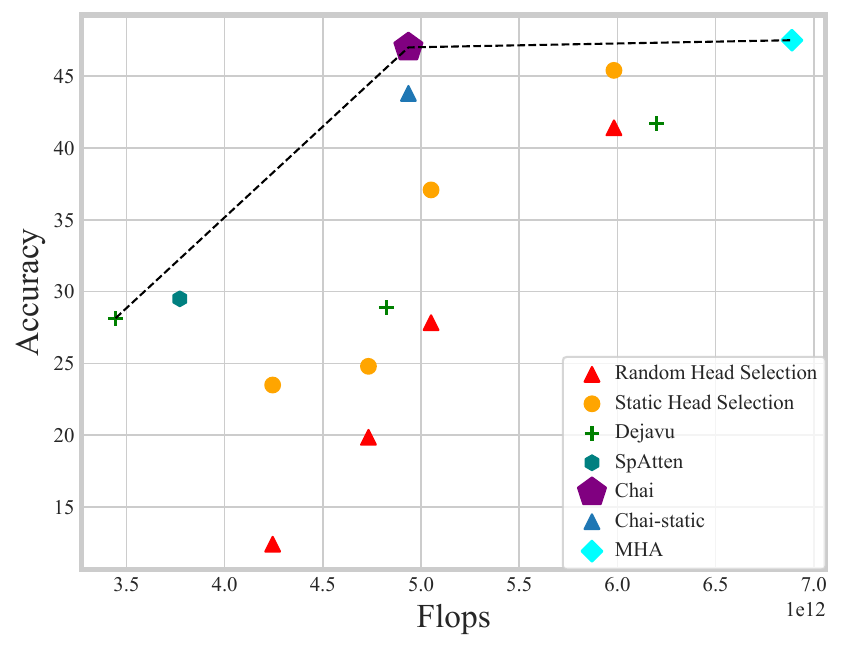}
    \vspace{-0.45cm}
    \caption{\small{\textbf{Accuracy vs Flops:} We study various methods of clustering attention heads, and plot the runtime for sequence length of 2048. For random head selection we randomly choose heads to combine in increasing number of 4, 8, 16 and 24. For \textit{Static Head Selection}, we choose the heads to combine based on activations. \system is our proposed method.}}
    \label{fig:pareto_optimality_flops}
    \vspace{-0.45cm}
\end{figure}

Pruning LLMs can potentially ease the compute bottleneck, however it is challenging as the classical methods for pruning~\cite{frankle2018lottery,chen2020earlybert,you2019drawing, waleffe2020principal} require fine-tuning or iterative training which is prohibitively expensive for LLMs due to massive memory and compute cost. There have been recent pruning works such as \dejavu~\cite{liu2023deja} which perform pruning based on the context at inference time without requiring fine-tuning. However, 
we observe that methods like \dejavu are 
primarily designed for large  parameter-inefficient models such as OPT~\cite{zhang2022opt} and the insights used to build \dejavu
are not directly applicable on newer parameter efficient models like \llama-7B (Section~\ref{sec:background}). In Figure~\ref{fig:pareto_optimality_flops}, we show that \system achieves the best trade-off between flops and accuracy compared to the state-of-the-art methods.
Furthermore, 
 runtime pruning  methods like \dejavu only reduce the compute cost and have no effect on the large memory requirements of K,V cache.
 
The Attention module co-design methods like GQA~\cite{ainslie2023gqa} require re-training of LLMs,
\eg \llama-2~\cite{touvron2023llama2} trained the models from scratch to utilize the benefits of GQA, making it quite expensive. Even in the case where users are willing to perform retraining, accuracy trade-off between GQA and MHA will not be known prior to multiple rounds of training.
Further, Attention module co-design methods only reduce the K,V cache size and do not reduce computational complexity. 
Therefore, there is a need for a method, which can reduce both the compute and K,V cache overhead for attention and is - (i) Applicable on a  wide range of models (from \llama--7B to OPT-66B).  (ii) Does not require any fine-tuning or re-training. 

\begin{figure}[t]
    \begin{center}
    \begin{subfigure}[t]{0.49\textwidth}
    \centering
    \includegraphics[width=0.7\textwidth]{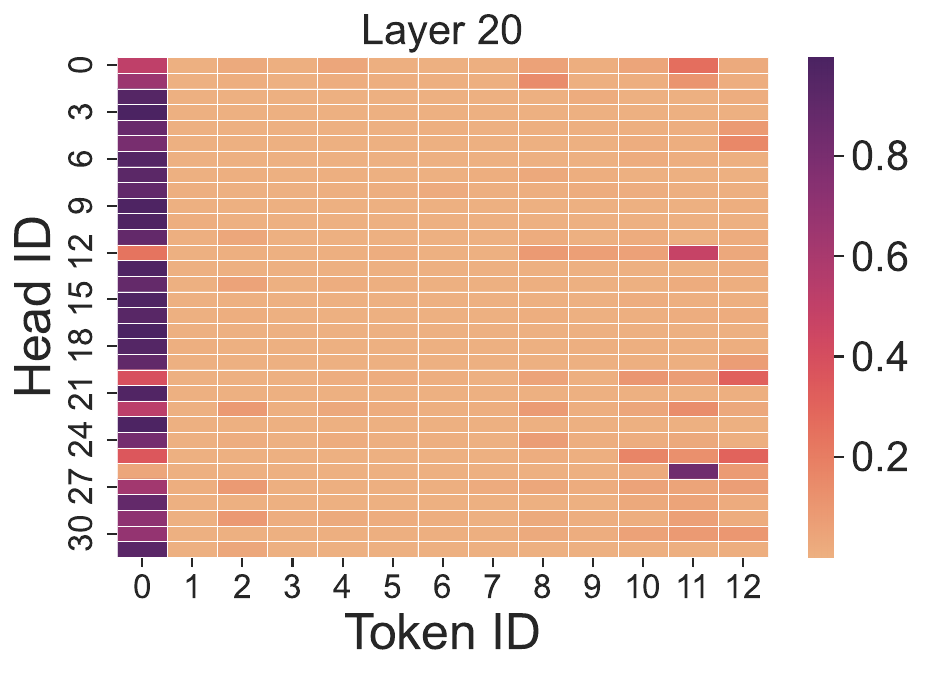}
    \vspace{-5pt}
    \caption{\small{\textbf{Activations of Multi Head Attention:} Figure shows activation scores for each token for each head. We observe that several heads give similar scores to the sequence.}}
    \label{fig:attention_scores}
    \end{subfigure}
    \begin{subfigure}[t]{0.49\textwidth}
    \centering
    \includegraphics[width=0.7\textwidth]{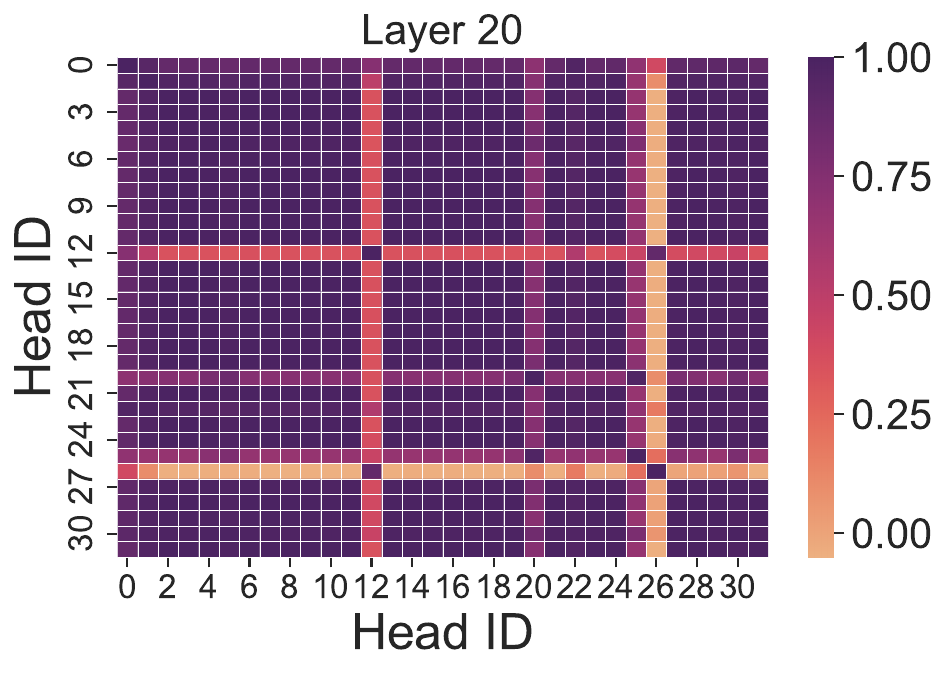}
    \vspace{-5pt}
    \caption{\small{\textbf{Pairwise cross correlation:} Pairwise cross-correlations show existence of three clusters- Heads [12,26] show strong correlation forming one cluster, Heads [20,25] form another, and the remaining heads form a third cluster.}}
    \label{fig:correlation_score}
    \end{subfigure}
    \end{center}
    \vspace{-5pt}
    \caption{\small{\textbf{Redundancy across heads for \llama--7B}}}
    \vspace{-15pt}
\end{figure}

In this work we present Clustered Head Attention for efficient LLM Inference (\system), a dynamic inference time method for efficient LLM inference that does not require fine-tuning. 
\system is inspired by two observations. First, several heads in multi-head attention give similar weight to each token in a given sequence, indicating redundant compute. 
In Figure~\ref{fig:attention_scores} we show attention scores for a single layer of \llama -7B for an auto-regressive decoding step of a sentence. We observe that several heads output similar scores, \ie giving similar weight to each token in the sequence.   
Figure~\ref{fig:correlation_score} highlights the similarity in attention score by plotting correlation for the activation for \llama-7B. In Figure~\ref{fig:correlation_score} we observe that there are three clusters and within these clusters the correlation is greater than 0.95. 

\begin{figure}[t]
    \begin{center}
    \begin{subfigure}[t]{0.22\textwidth}
    \includegraphics[width=\textwidth]{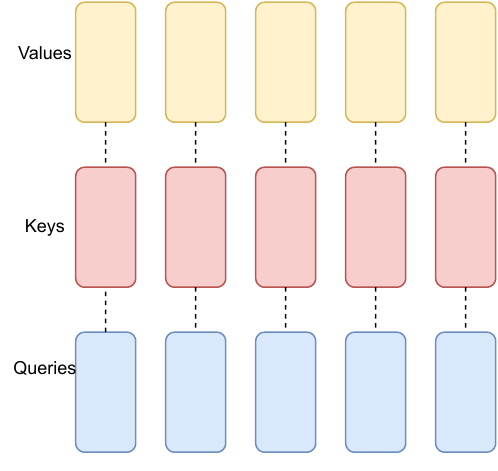}
    \caption{\small{\textbf{Multi-Head Attention}}}
    \end{subfigure}
    \begin{subfigure}[t]{0.23\textwidth}
    \includegraphics[width=\textwidth]{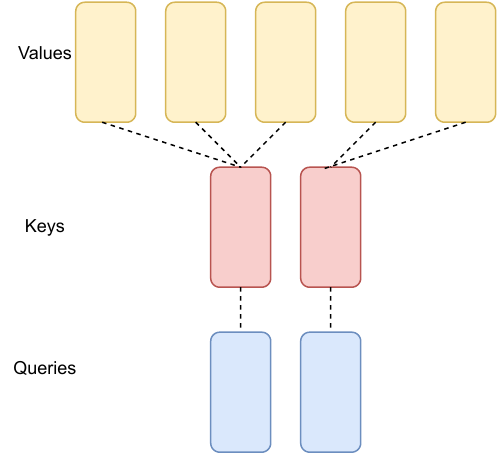}
    \caption{\small{\textbf{Clustered Head Attention}}}
    \end{subfigure}
    \end{center}
\vspace{-10pt}
    \caption{\small{\textbf{Clustered Head Attention:} Schematic of clustered head attention, comparing it with popular Multi-Head Attention. In clustered head attention, we remove the query and key vectors which produce similar attention scores.}}
    \vspace{-10pt}
        \label{fig:cha_schematic}
\end{figure}

This indicates that by identifying attention heads with similar attention scores and clustering them together we can reduce the number of self-attention operations for MHA by calculating self-attention only for a single head within a cluster. 
Secondly, we observe that for each request to an LLM we can accurately determine the heads which are going to give similar (attention) weight to the tokens in a sequence after running a few decoding steps on the sequence (Section~\ref{sec:cluster_membership}). Schematic in Figure~\ref{fig:cha_schematic} depicts both Multi-Head and Clustered-Head Attention.  

Our contributions in this paper are as follows:
\begin{myitemizeleft}
    \item We show that there is high level of redundancy across several different heads of multi head attention, 
    and the redundancy varies differently across layers with increasing redundancy towards later layers.
    \item We introduce \system, a practical and principled inference time pruning method which clusters attention heads that have similar output together with dynamic determination of clusters. \system reduces both compute and  K,V cache size for self attention.
    \item  We show that \system is capable of reducing the inference time by up to 1.73$\times$ and K,V cache memory size by up to 21.4\% compared to MHA for \llama models with minimal accuracy trade-off (maximum of 3.2\%).
    \item Compared to other runtime pruning methods like \dejavu, which only works well for OPT models, \system outperforms \dejavu and performs well for wider class of models.
\end{myitemizeleft}


\section{Background and Related Work}
\label{sec:background}
We first provide background on inference process for decoder only transformers like GPT~\cite{radford2019language, brown2020language}, LLaMa~\cite{touvron2023llama,touvron2023llama2} and the bottlenecks in performing inference. 
Further, we discussed several prior lines of work which have tried to tackle the inference bottlenecks for transformer based model.

\textbf{Decoder-only Transformer} A decoder-only transformer forms the building block of popular LLMs. A single decoder block consists of a self attention layer and a MLP. An input token is fed into the decoder block, to perform next-word prediction. The self attention block uses prior query (Q), key (K) and value (V) vectors associated with current token. These tokens are extracted by performing a linear projection with query, key and value weight matrices associated with a transformer. 

To precisely define Multi-Head Attention (MHA), let $H$, $T$, $d$ be positive integers,  where $H$ denotes number of heads, $T$  denotes sequence length, $d$ denotes model dimension.  Let $x \in \real^{T \times d}$ be input to the MHA layer. For a single head $h$, then $\key^h = x\weight_{K}^{h}$, $\query^h = x\weight_{Q}^h$ and $\valuev^h = x\weight_{V}^h$ denote the corresponding key, query and value vector. The attention matrix for head $h$ is calculated as follows: 
$$A_h = \sigma(\frac{1}{\sqrt{d}} Q^h K{^h}{^T}) $$
Output of MHA is denoted by:
$$ y= A_0V_0 \oplus A_1V_1 \oplus A_2V_2 \oplus \cdot \cdot \cdot \oplus A_HV_H$$
For performing inference, self attention needs access to the query, key and values associated with prior tokens. In order to avoid re-computation, inference serving systems cache the prior tokens in a sequence. 

Compute cost required for multiple attention heads and memory capacity required for storing key and value vectors associated with each head during inference form two primary bottlenecks for LLM inference. In this work, we focus on reducing both memory and compute requirements via clustering multiple attention heads with similar output.

\textbf{Building Efficient Transformers.} Improving efficiency of transformer models has been of major focus in recent years. Prior work can be broadly categorized in the following fields 
 - (i) Hardware-software co-design~\cite{dao2022flashattention, dao2023flashattention,ham20203, ham2021elsa, tambe2021edgebert, fang2022algorithm, qin2023fact, wang2021spatten}, (ii) Knowledge distillation~\cite{hsieh2023distilling, jiao2019tinybert, sanh2019distilbert, wang2020linformer} (iii) Neural Architecture Search (NAS)~\cite{zhou2023brainformers,kitaev2020reformer,lagunas2021block} and (iv) Pruning~\cite{voita2019analyzing,liu2023deja} and Quantization~\cite{frantar2022gptq, xiao2023smoothquant, kim2021bert, shen2020q, dettmers2022llm, dettmers20158,dettmers2023case}. In this work our focus is on pruning , which we discuss next.

\textbf{LLM Quantization.} Recently several methods have been proposed to perform post training quantization allowing models to be quantized to a lower precision~\cite{frantar2022gptq,xiao2023smoothquant, dettmers2023case}. The goal of these methods is to perform quantization so as to minimize the error, \system is orthogonal to quantization based mechanisms as it depends on the insight of several attention heads focusing on the same tokens. The goal of quantization methods is to keep the same properties of original models, therefore we believe \system can be used to further accelerate post training quantized neural networks.

\textbf{LLM Pruning.} Pruning is a widely studied method to improve inference time by removing unused weights post training. 
Several prior works have looked at pruning for language models~\cite{chen2020earlybert, prasanna2020bert,chen2020lottery}. For example, oBERT is a second order method to reduce the number of weights~\cite{kurtic2022optimal}. Although these approaches can compress a model,  they rarely yield inference speedups due to lack of hardware support for sparse operations on modern GPUs. To overcome the challenges, low rank decomposition methods~\cite{wang2023cuttlefish, wang2021pufferfish, wang2019structured}, attention head pruning~\cite{michel2019sixteen, voita2019analyzing}, layer dropping~\cite{sajjad2023effect,fan2019reducing, dai2023efficient} were proposed. However, these methods are infeasible for LLMs due to the use of iterative gradient calculations or fine-tuning leading to high resource requirements.

\begin{figure}[t]
    \begin{center}
    \begin{subfigure}[t]{0.5\textwidth}
    \centering
    \includegraphics[width=0.7\textwidth]{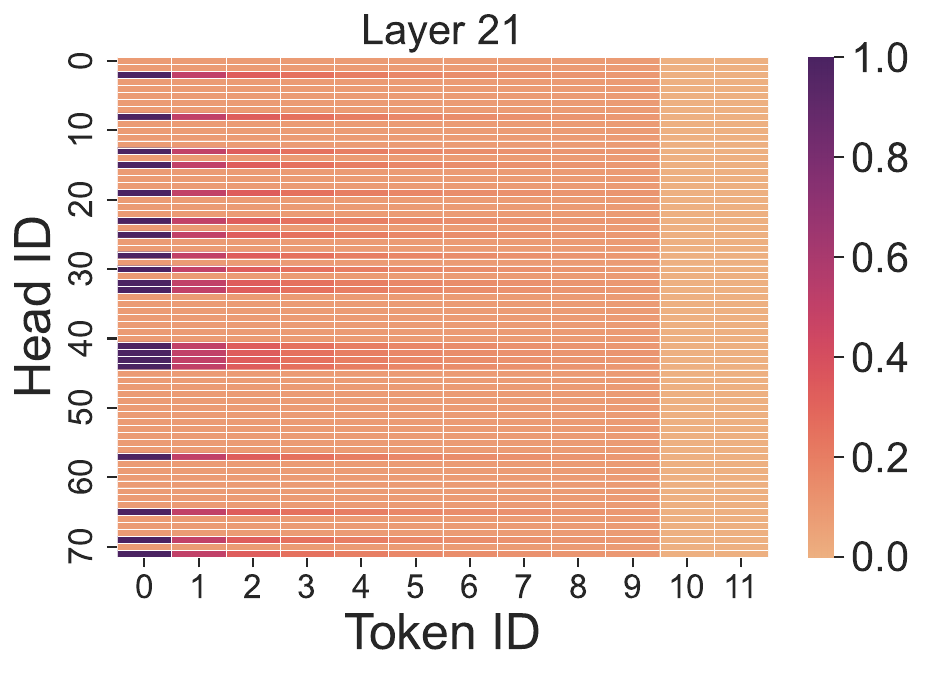}
    \vspace{-8pt}
    \caption{\small{\textbf{\opt-66B}: For several heads the activation scores are uniform, \ie the heads given close to equal importance to each input token.}}
    \end{subfigure}
    \begin{subfigure}[t]{0.5\textwidth}
    \centering
    \includegraphics[width=0.7\textwidth]{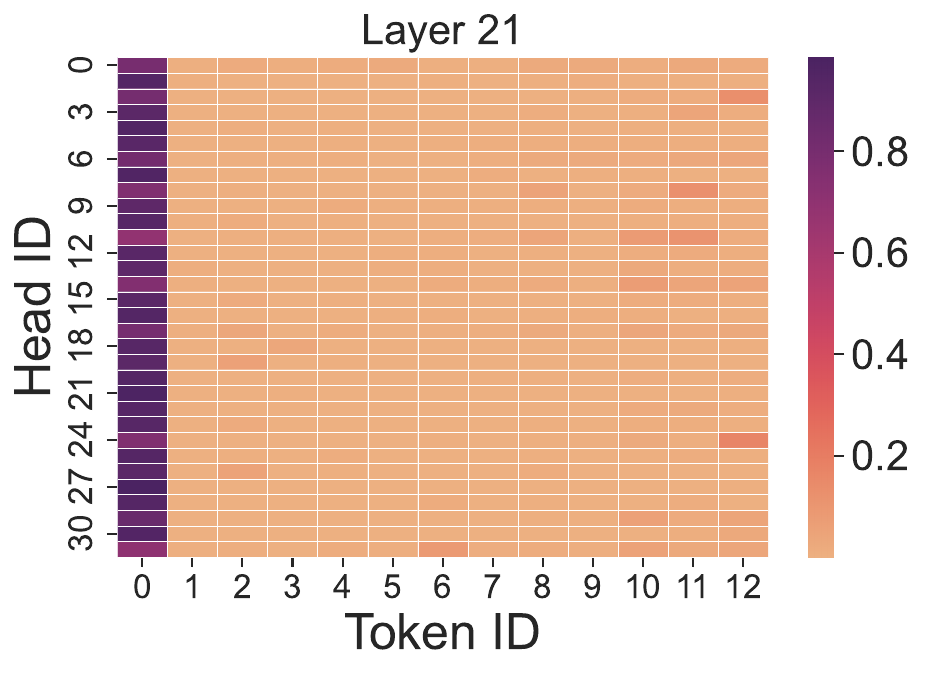}
    \vspace{-8pt}
    \caption{\small{\textbf{\llama-7B:} Heads in \llama-7B specifically pay attention to a specific token. However, multiple heads are attending to same token, in this case the first token.}}
    \end{subfigure}
    \end{center}
    \vspace{-10pt}
    \caption{\small{\textbf{Activations for \opt-66B and \llama-7B for an  exemplary sentence:} We observe that \opt-66B has several heads which give uniform attention scores to tokens whereas \llama-7B does not. However, both models have redundancies across heads, \ie groups of heads are give similar attention to each token.}}
    \label{fig:activation_opt_llama}
    \vspace{-20pt}
\end{figure}

To overcome these issues, a recently proposed method, \dejavu~\cite{liu2023deja}, identifies portions of the model which are unused for a given context. To reduce the overhead of self-attention, \dejavu prunes attention heads which give \emph{uniform weight across tokens}. 
We plot the activations for an exemplary sentence used by \dejavu for both \opt-66B and \llama-7B in Figure~\ref{fig:activation_opt_llama}. We observe that while there are heads which give uniform weight to each token in \opt-66B model, there are no such heads in more parameter efficient models like \llama-7B, indicating that for smaller parameter efficient models like \llama  \dejavu might not be applicable. (Additional plots for different layers can be found in Appendix-\ref{sec:additional_plots}.)
The primary difference between OPT and \llama activation patterns could be attributed to the fact that \llama models are trained significantly longer and with more data.

We observe that \system's insight about redundancy in the output of multiple heads in the attention holds across both \opt and \llama family of models. In our evaluation (Section~\ref{sec:eval}), we perform quantitative comparison between \system and \dejavu. 

\begin{figure}[t]
    \centering
    \includegraphics[width=0.8\linewidth]{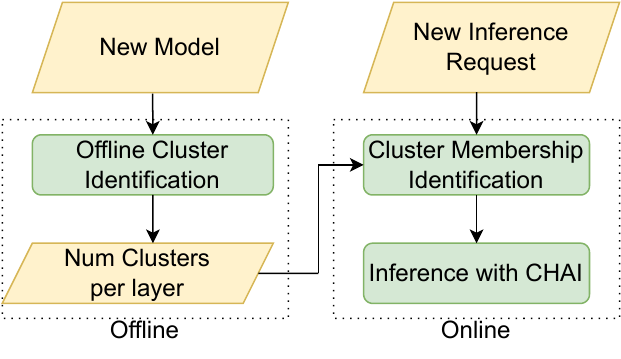}
    \vspace{-10pt}
    \caption{\small{\textbf{\system Flow:} In the offline phase, we run clustering and perform elbow plot analysis for each new model. Then, for each new inference request we only perform cluster membership identification based on online performance.}}
    \vspace{-20pt}
    \label{fig:chai_flow}
\end{figure}

\begin{figure*}[t]
    \begin{center}
    \begin{subfigure}[b]{0.24\textwidth}
        \includegraphics[width=\textwidth]{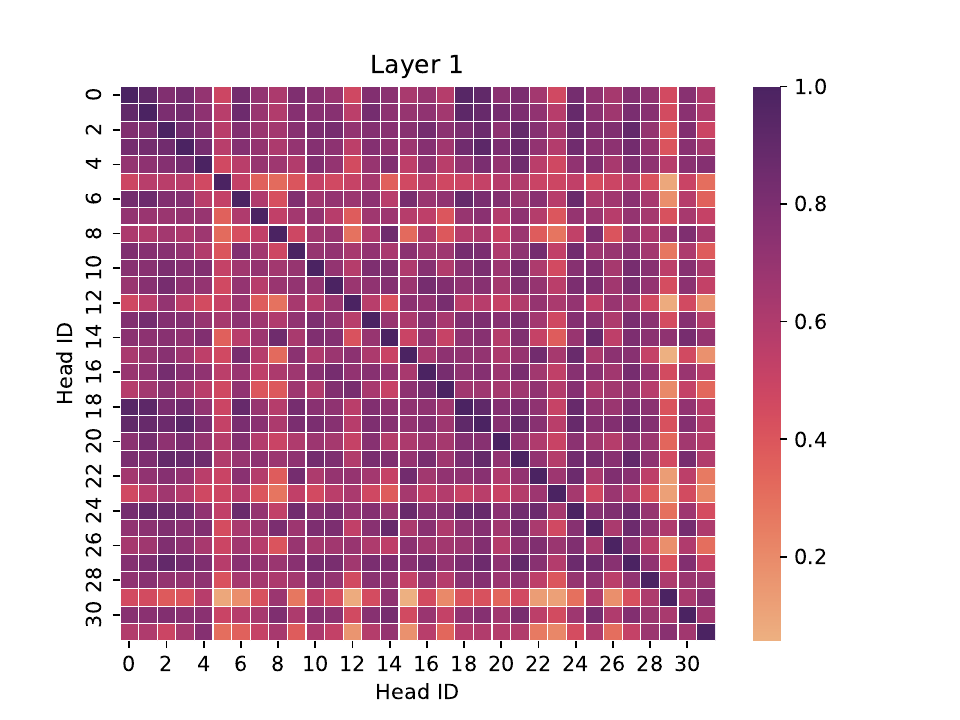}
        \vspace{-0.2in}
    \caption{Layer 1}
     \label{fig:layer_1_avg}
    \end{subfigure}
    \begin{subfigure}[b]{0.24\textwidth}
    \includegraphics[width=\textwidth]{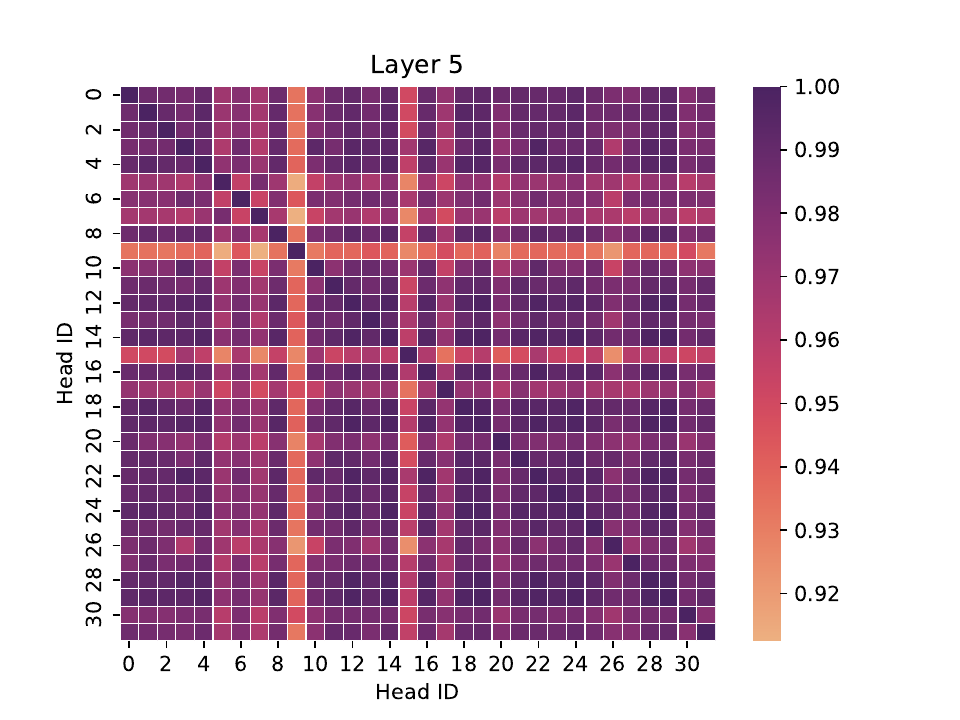}
    \vspace{-0.2in}
    \caption{Layer 5}
    \label{fig:layer_5_avg}
    \end{subfigure}
    \begin{subfigure}[b]{0.24\textwidth}
    \includegraphics[width=\textwidth]{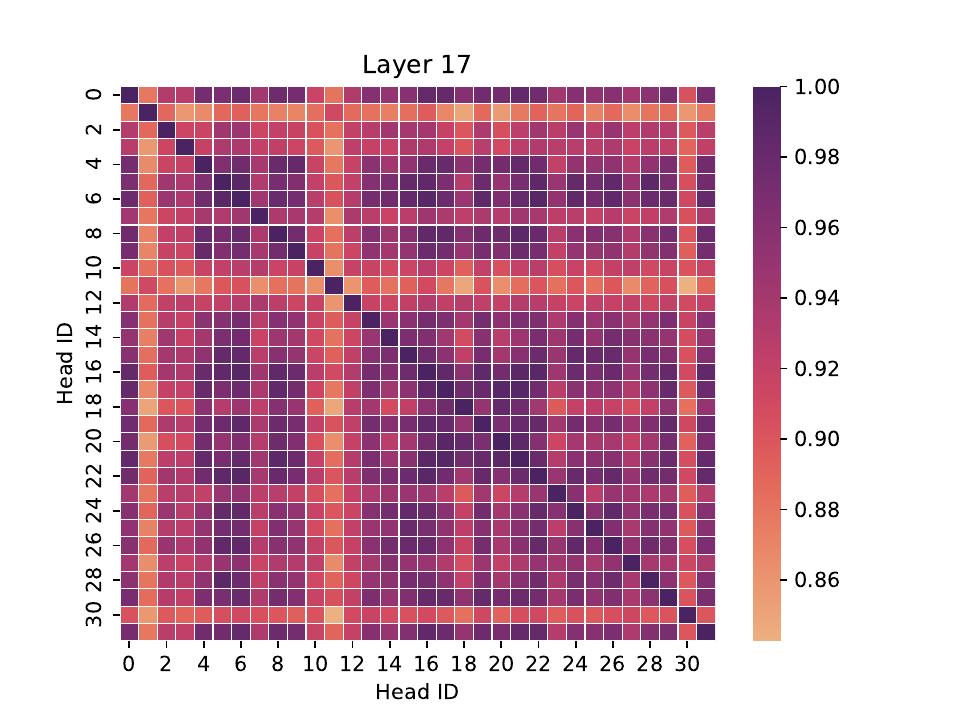}
    \vspace{-0.2in}
    \caption{Layer 17}
    \label{fig:layer_17_avg}
    \end{subfigure}
    \begin{subfigure}[b]{0.24\textwidth}
    \includegraphics[width=\textwidth]{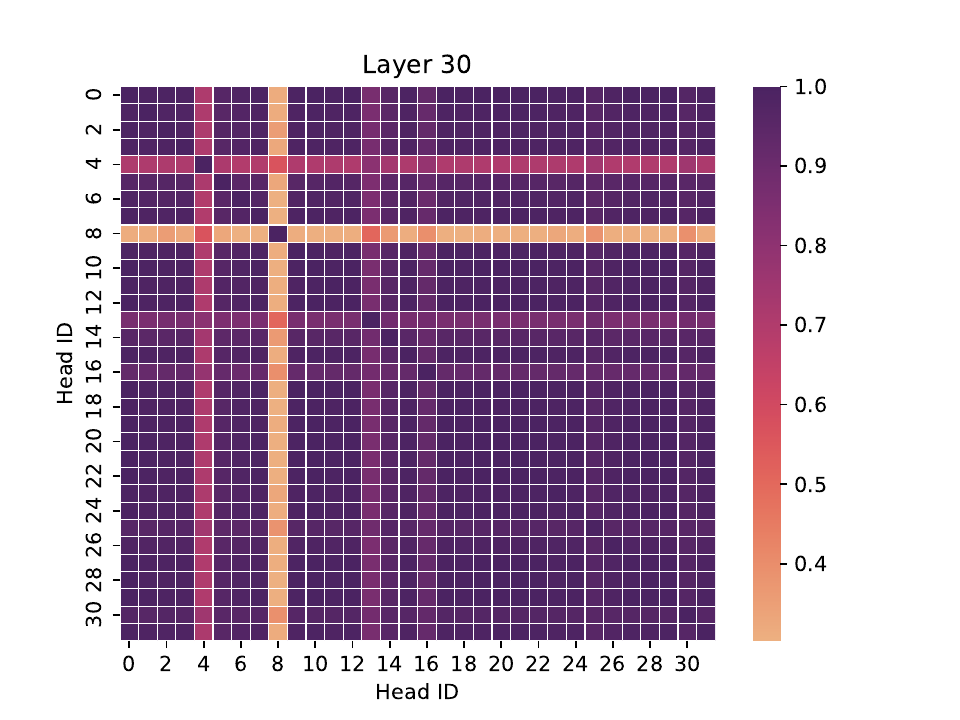}
    \vspace{-0.2in}
    \caption{Layer 30}
    \label{fig:layer_30_avg}
    \end{subfigure}
    \end{center}
    \vspace{-0.2in}
    \caption{\small{\textbf{Average Correlation for 1024 Samples of C4 on \llama-7B:} The above figure shows two interesting observations. First, there exists high amount of correlation across several heads of attention. Second, the correlation is not uniform across layers, with later layers having higher correlation, \ie, first layer has very little correlation but correlation increases in later layers.}}
    \label{fig:avg_correlation}
     \vspace{-10pt}
\end{figure*}
\begin{figure*}[t]
    \begin{center}
    \begin{subfigure}[b]{0.24\textwidth}
        \includegraphics[width=\textwidth]{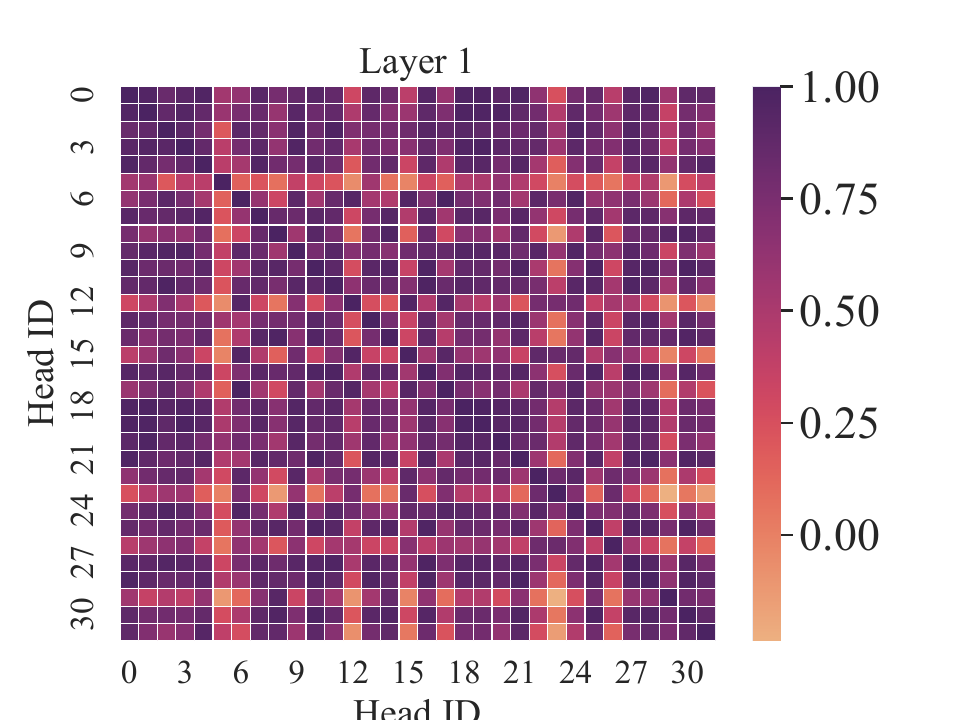}
    \caption{Layer 1}
     \label{fig:layer_1}
    \end{subfigure}
    \begin{subfigure}[b]{0.24\textwidth}
    \includegraphics[width=\textwidth]{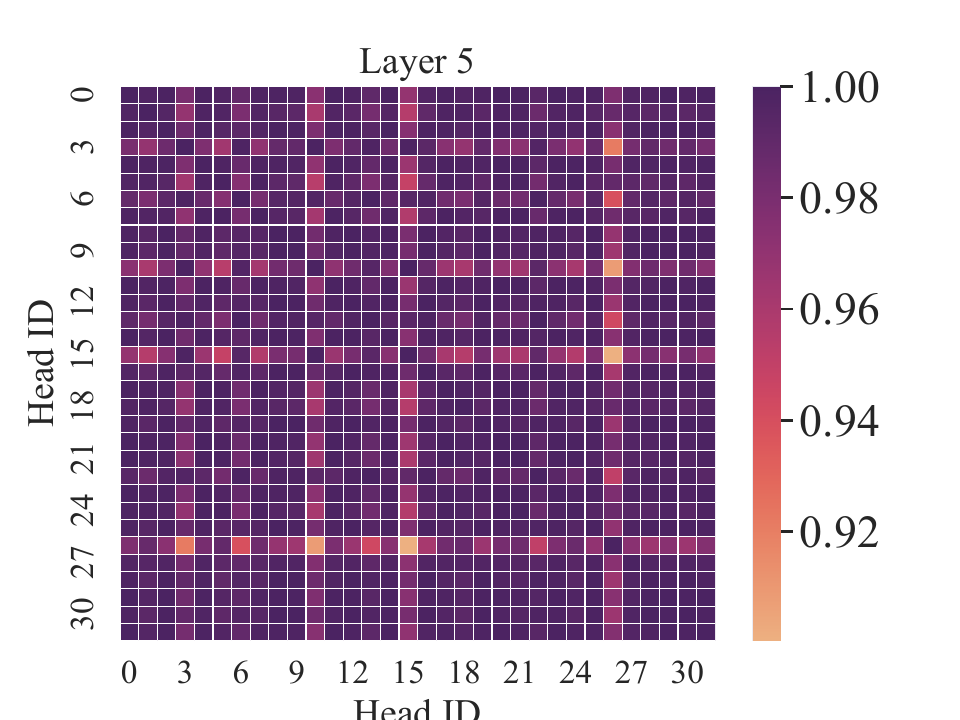}
    \caption{Layer 5}
    \label{fig:layer_5}
    \end{subfigure}
    \begin{subfigure}[b]{0.24\textwidth}
    \includegraphics[width=\textwidth]{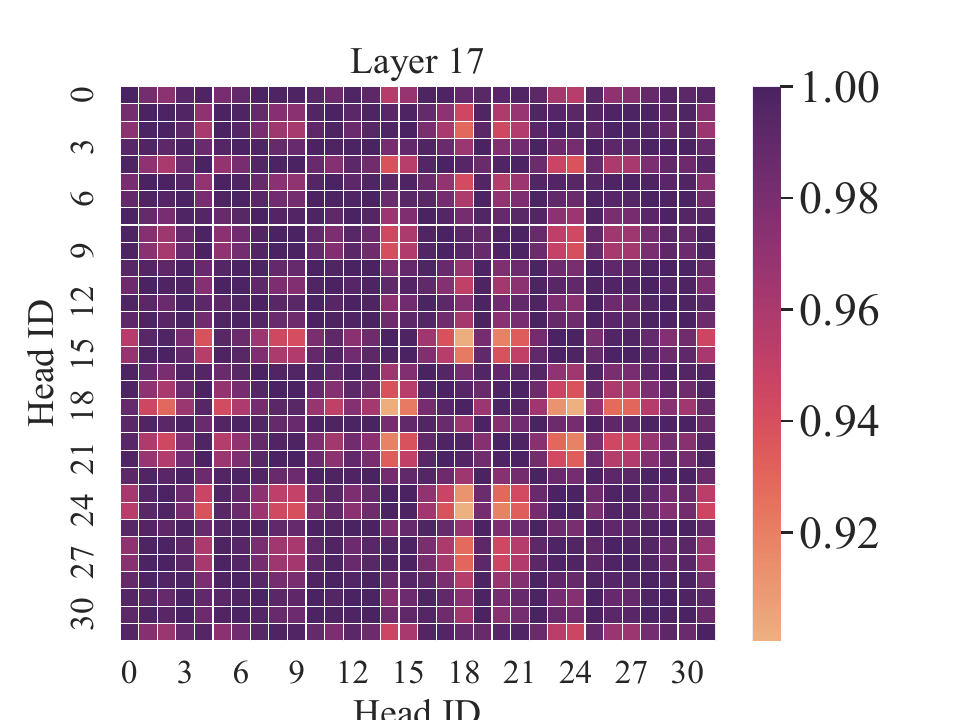}
    \caption{Layer 17}
    \label{fig:layer_17}
    \end{subfigure}
    \begin{subfigure}[b]{0.24\textwidth}
    \includegraphics[width=\textwidth]{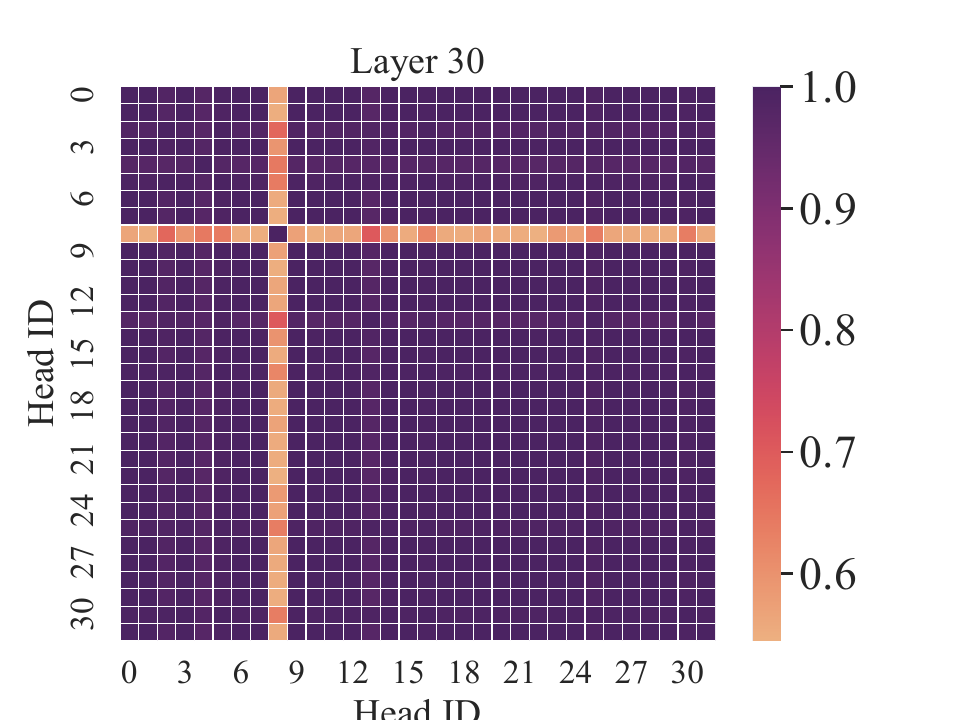}
    \caption{Layer 30}
    \label{fig:layer_30}
    \end{subfigure}
    \end{center}
     \vspace{-10pt}
    \caption{\small{\textbf{Correlation on a randomly selected single sample of \llama-7B.}}}
    \label{fig:single_correlation}
    \vspace{-15pt}
\end{figure*}

\textbf{K,V Cache Compression.} Prior works which have tried to reduce the K,V cache size~\cite{liu2023scissorhands, zhang2023h} by storing the K,V cache values for the most recent important tokens. However, they can not directly improve the latency of generating the next token, as they still perform the full transformer compute before finally deciding which K,V pairs should be stored. 
On the other hand, \system reduces not just the K,V cache size, it is also able to reduce the latency of next word prediction.

\textbf{Speculative Decoding.} Speculative decoding~\cite{leviathan2023fast, yang2023predictive, xia2023speculative} is a popular method where a draft model is used to cheaply generate a sequence of draft tokens which can be efficiently verified by a target LLM. Speculative decoding can significantly reduce the latency of LLM serving, however it further exacerbates the compute and memory requirements as it requires additional resources to run both the draft and target model. \system on the other hand is focused on reducing the resource required for inference.

\section{\system}
Next, we describe \system.
We first describe the key insights which have been used to build \system. Then, we detail \system's runtime pruning algorithm which is inspired by our insights and discuss how we perform inference using \system. Figure~\ref{fig:chai_flow} provides a high level overview of inference using \system, which includes offline and online components.

\subsection{Observations}
\label{sec:observation}
Our primary insight stems from the observation that there is a high amount of correlation across the output of various attention heads in MHA, \ie the output of several attention heads focuses on the same tokens. In Figure~\ref{fig:avg_correlation}, we plot the average correlation across the 32 heads of \llama-7B for 1024 samples of the C4~\cite{c4} dataset for different layers and in Figure~\ref{fig:single_correlation}, we plot correlation for a single sample of the dataset.
These show us two insights - (i) Several heads output similar attention scores for each example and (ii) The amount of correlation increases in later layers, with heads in later layers with having higher correlation.
This indicates that there is an opportunity to cluster attention heads with similar output and only run the self-attention operation for one of the representative attention heads within each cluster, thus reducing the amount of computation as well as the size of K,V cache.

\begin{figure}[t]
    \centering
    \includegraphics[width=0.8\linewidth]{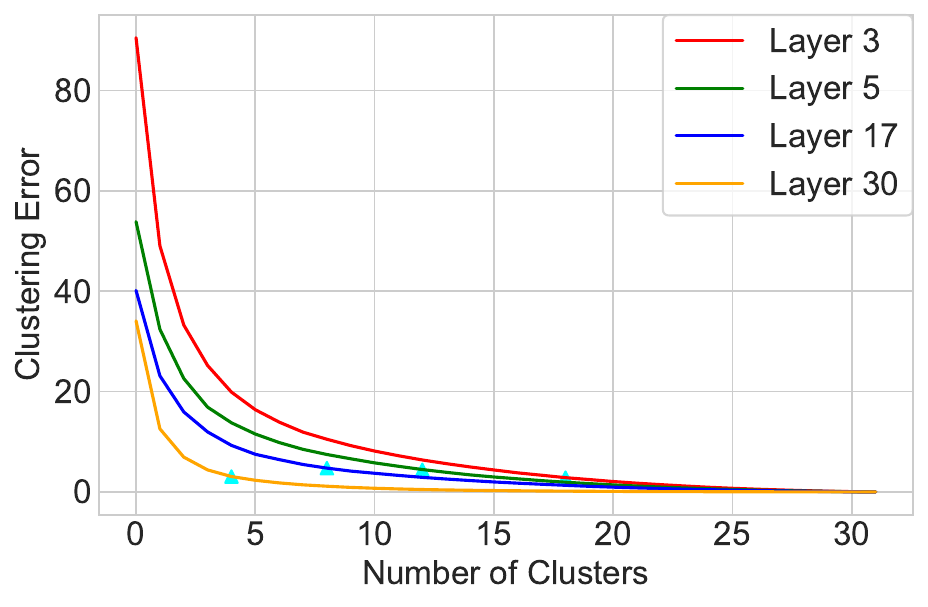}
    \vspace{-15pt}
    \caption{\small{\textbf{ Clustering Error:} We plot the clustering error on 1024 samples of C4-dataset. The markers represent the number of clusters we choose for a layer. }}
    \vspace{-15pt}
    \label{fig:avg_clustering_error}
\end{figure}

\begin{figure}[t]
    \centering
    \includegraphics[width=0.8\linewidth]{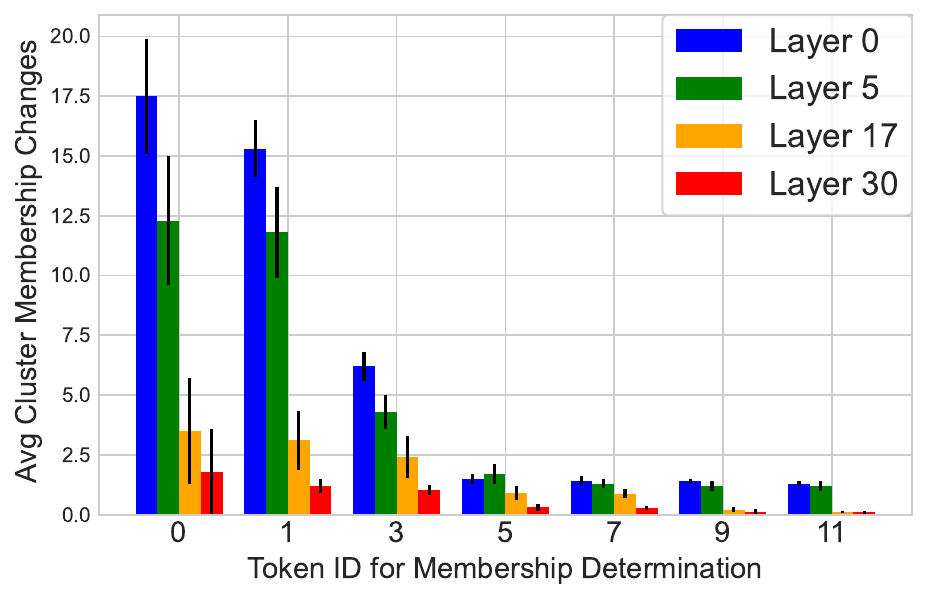}
    \vspace{-15pt}
    \caption{\small{\textbf{Cluster Membership Evaluation:} We evaluate the number of times the cluster membership changes for performing next token prediction. We observed that if clustering is performed beyond the fifth token the number of times cluster membership changes is quite small.}}
    \vspace{-15pt}
    \label{fig:cluster_membership_eval}
\end{figure}

\begin{figure*}[t]
\label{fig:chai_step_schematic}
\begin{center}
    \begin{subfigure}[t]{0.35\textwidth}
    \includegraphics[width=\textwidth]{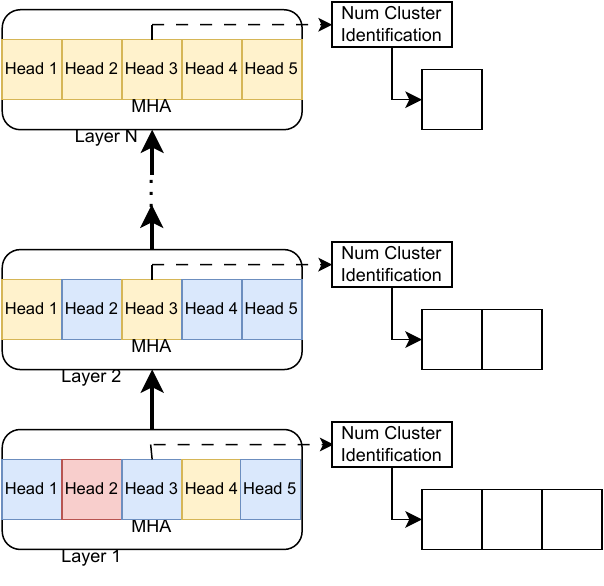}
    \vspace{-0.45cm}
    \caption{\small{\textbf{Offline Cluster Identification:} For each new model we run an offline cluster identification phase. We collect the activations and perform Elbow-plot analysis to decide number of clusters. }}
    \label{fig:cluster_num_identification}
    \end{subfigure}\quad
    \begin{subfigure}[t]{0.35\textwidth}
        \includegraphics[width=\textwidth]{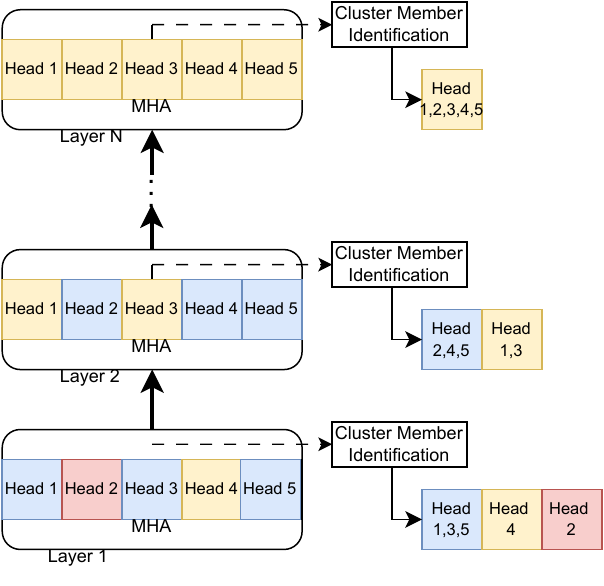}
        \caption{\small{\textbf{Cluster Membership Identification:} For each new request, we initial run with multi-head attention for first five tokens. Using this we determine the number of clusters in each layer.}}
        \label{fig:cluster_mem_identification}
    \end{subfigure}\quad
    \begin{subfigure}[t]{0.18\textwidth}
        \includegraphics[width=\textwidth]{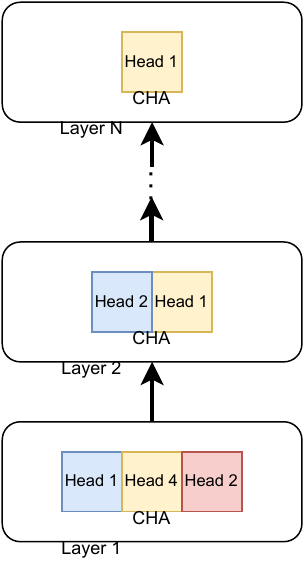}
        \caption{\small{\textbf{\system Inference:} Post cluster membership identification we substitute MHA with Clustered Head Attention. }}
        \label{fig:chai_inference}
    \end{subfigure}
\vspace{-10pt}
\caption{\small{\bf{Schematic of \system} detailing three phases of the system. }}
\label{fig:chai_flow_desc}
\vspace{-10pt}
\end{center}
\end{figure*}

\textbf{Problem Formulation.}
Next, we formally define the problem of finding heads whose attention score is similar. 
Let $H$ be the total number of attention heads, let $S = \{ \opair{K^1,Q^1}, \opair{K^2, Q^2}, \opair{K^3, Q^3}, \cdot \cdot \cdot, \opair{K^H, V^H} \}$ be the set of $Q,K$ pairs associated with each head $h$. 
Our goal is to find $k$ subsets, $S_1 \subset S, S_2 \subset S, S_3 \subset S, \cdot \cdot \cdot S_k \subset S$ such that $<Q,K>$ pairs in each subset $S_i$ produce similar output under function $f$. Where function $f$ is the self attention operation, where $f(Q,K) = \sigma (Q K{^T}) $. Further, we want $\cup_{i=1}^{k}S_{i} = S$. 

Formally, we want to find $S_i$,
$$\forall <K^n, Q^n>, <K^m, Q^m> \in S_i, $$
s.t.$$  f(K^n, Q^n) \approx f(K^m, Q^m) $$
Informally, we want subset of heads, where within each subset the self attention operation gives similar outcome.

In order to solve this problem we need to determine $k$ which represents the number of such subsets,
and the membership of such subset $S_i$.
Our observations empirically demonstrate the existence of such a solution. We can potentially solve this problem using clustering, where determining the number of subsets translates to determining number of clusters and determining cluster membership becomes determination of cluster membership.

To observe memory and compute savings, we need an accurate and efficient method to determine the number of clusters and their membership \emph{without having access to activations}. Solving this forms a core contribution of our work.

\subsection{Determination of Number of Clusters}
\label{sec:num_clusters}
\paragraph{Challenges.} Figure~\ref{fig:avg_correlation} and Figure~\ref{fig:single_correlation} indicate that the number of clusters varies widely per layer in a LLM. Specifically, the last few layers in the LLM exhibit a very low number of clusters (high redundancy), whereas the early layers demonstrate a high degree of variance across the output of heads resulting in large number of clusters. 
This observation suggests that the method used to determine number of clusters needs to make decisions for each layer independently. Additionally, widely used methods such as Elbow plot method~\cite{thorndike1953belongs} for determining number of clusters entail manual effort making cluster number determination impractical at inference time. 

\textbf{Design.} 
To determine the number of clusters, we propose an offline strategy we run once for each model.
In our case, we sample a small number of samples (1024) from the C4~\cite{c4} dataset and perform elbow-plot analysis by plotting clustering error (i.e. sum of squared distance from the closest cluster) as a function of number of clusters. Figure~\ref{fig:avg_clustering_error} shows the clustering error for \llama-7B for the samples selected. Based on the Elbow-plot analysis we choose the number of clusters when the error plateaus.

The offline analysis is performed once for each network by using the C4~\cite{c4} dataset. We do not change the number of clusters determined for a new dataset.
\subsection{Determination of Cluster Membership}
\label{sec:cluster_membership}
\textbf{Challenges.} Having determined number of clusters, we need to determine the membership of these clusters, \ie which heads belong to which cluster in each layer. 
For Figure~\ref{fig:avg_correlation},~\ref{fig:single_correlation} and ~\ref{fig:avg_clustering_error}, we perform clustering based on activations obtained by performing the forward pass. However, for each decoding step, performing clustering on output of self attention post forward pass will not yield any performance benefit as we will still be performing the original compute and using the full K,V cache.
In order to utilize the insights observed in Section~\ref{sec:observation}, we will need to decide the cluster members without having access to the output of the self attention.

\textbf{Design.}
A simple strategy would have been keeping the cluster membership static across the tokens and independent of input context, \eg we use the same cluster membership found during offline analysis with C4 data in the previous section.
For evaluation purposes we call this version of head selection \system-static.

However, we observed that the cluster membership does not remain static and varies based on context. 
When comparing Figure~\ref{fig:single_correlation}, which plots correlation for a single example, 
with Figure~\ref{fig:avg_correlation}, which plots correlation for 1024 samples, we observe that the correlation across heads varies with varying context. Therefore, the correlation across the output of the heads depends on the context (input prompt), \ie \emph{a solution to determine the membership of each cluster has to account for context.}
To understand the effects of accounting for context while clustering heads, we analysed the change in cluster membership changes and clustering with different context. In Figure~\ref{fig:cluster_membership_eval}, we observed an interesting phenomenon, after determining cluster membership by accounting for five tokens, the cluster membership does not change frequently. A direct outcome of this observation is that for each new sequence we can perform clustering based on the output of self-attention after the first five tokens. We observe that \emph{activation from first five tokens of a new sequence are enough to accurately predict the cluster membership.} This dynamic version of head selection further allows us to improve accuracy over \system-static.  
Figure~\ref{fig:cluster_mem_identification} shows an illustration of the membership identification step. Furthermore, evaluation results in Section~\ref{sec:eval} compare \system-static and \system performance. 

\subsection{Clustered Head Attention}
Once we have decided which heads have similar attention output, we can than use Clustered Head Attention to combine key and query vectors for the heads.

\subsection{Inference using \system}
Next we, discuss the inference flow of \system, illustrated in detail in Figure~\ref{fig:chai_flow_desc}.
For each new model we first perform offline cluster identification (Figure~\ref{fig:cluster_num_identification}). Then for each new request, we determine the cluster membership using K-Means clustering once we have processed five tokens, using the observed activations (Figure~\ref{fig:cluster_mem_identification}). After this step, we keep the clustered heads same throughout inference (Figure~\ref{fig:chai_inference}). 

There are two direct outcomes of \system's design. First, we directly reduce the amount of computation by removing redundant heads. Secondly, after a pre-determined token we fix the heads which are going to be pruned, this also allows us to remove the corresponding \emph{Key} tokens associated, which significantly reduces the K,V cache size. 
Therefore, \system allows us to reduce both the inference compute as well as the size of the K,V cache required.

\begin{table}[t]
\caption{Accuracy on \opt-66B}
\label{tab:zeroshotopt66bdelta}
\resizebox{\linewidth}{!}{
\begin{tabular}{@{}llllll@{}}
\toprule
Method      & PIQA  & Hellaswag & Arc-Challenge & Arc-Easy & Boolq \\ \midrule
MHA         & 78.4  & 71.1      & 41.6          & 64.7     & 65.4  \\ \midrule
DejaVu-50\% & -0.25 & -0.7      & -0.6          & -0.2     & -4.0  \\
CHAI-static & -1.35 & -1.7      & -0.7          & -0.7     & -0.7  \\
CHAI        & \bf{-0.15} & \bf{0.1}       & \bf{0.1}           & \bf{-0.1}     & \bf{-0.6}  \\ \bottomrule
\end{tabular}}
\vspace{-15pt}
\end{table}
\vspace{-5pt}
\section{Evaluation}
\label{sec:eval}
We experimentally verify the performance of \system and compare it to \dejavu~\cite{liu2023deja} and \spatten~\cite{wang2021spatten} on three different models of various sizes LLaMa-7B~\cite{touvron2023llama}, LLaMa-33B and OPT-66B~\cite{zhang2022opt}. We evaluate the models on five commonly used NLP tasks: PIQA~\cite{bisk2020piqa}, HellaSwag~\cite{zellers2019hellaswag}, Arc-Challenge and Arc-Easy~\cite{clark2018think} and BoolQA~\cite{clark2019boolq}.
\vspace{-5pt}
\subsection{Experimental Setup}
All our experiments are performed on servers with NVIDIA V100 GPUs. For \opt-66B we used eight GPUs on a single node, for \llama-33B we used four GPUs, and for \llama-7B, we used a single GPU for inference. \system is built on top of Meta's xFormers~\cite{xformers}. 

\vspace{-5pt}
\subsection{Accuracy Evaluation}
In our evaluation, we compare \system with Multi-Head Attention as baseline, static version of \system, as well two other state-of-the-art prior pruning methods; \dejavu and \spatten.
For \dejavu, we try different sparsity ratios, in order to try to match the accuracy number to MHA. We also compare \system to \spatten, a method which removes unimportant tokens and heads.

In Table~\ref{tab:zeroshotopt66bdelta}, we first verify that we are able to reproduce the performance numbers reported by \dejavu. To perform this, we took the \opt-66B and evaluated both \dejavu, \system and \system-static. We used \dejavu with 50\% sparsity as reported by the authors. 
We used the author provided code to train their MLP predictor layers and incorporate their scheme in our setup.
In Table~\ref{tab:zeroshotopt66bdelta}, we observe that we were able to replicate results for OPT-66B. Furthermore, \system is also able to match the accuracy of MHA for \opt-66B.

Next, we compare \system, \system-static and \dejavu with the pre-trained MHA network, using \llama-7B on 5 different datasets. 
For \dejavu we used three configurations, 50\% sparsity, 30\% sparsity and 10\% sparsity. 
 In Table~\ref{tab:zeroshotllama7bdelta}, we observe that when we use \dejavu with more 10\% sparsity we see significant decrease in accuracy (by 18.6\% for \dejavu-30\%). On the other hand, our method based on our close analysis of the behaviour of layers of \llama-7B is able to recover accuracy. We observe a maximum accuracy degradation of 3.7\% for \system. 
Similarly for \llama-33B using sparsity for more than 10\% leads to significant accuracy drop, meanwhile \system closely matches the accuracy of the pre-trained model using MHA with maximum degradation in accuracy by 0.14\%. 
This shows that \system is widely applicable across multiple datasets and models.
We also want to highlight that we do not perform any dataset specific tuning.



\begin{table}[t]
\caption{ Accuracy on \llama-7B}
\label{tab:zeroshotllama7bdelta}
\resizebox{\linewidth}{!}{
\begin{tabular}{@{}llllll@{}}
\toprule
Method      & PIQA  & HellaSwag & Arc-Challenge & Arc-Easy & BoolQ \\ \midrule
MHA         & 79.8  & 76.1      & 47.5          & 72.8     & 76.0  \\ \midrule
DejaVu-10\% & -3.9  & -4.7      & -5.78         & -3.18    & -7.4  \\
DejaVu-30\% & -13.3 & -18.6     & -18.75        & -4.2     & -20.2 \\
DejaVu-50\% & -24.6 & -50.7     & -19.35        & -46.3    & -21.6 \\
SpAtten     & -41.4 & -42.5     & -18.0         & -40.2    & -27.1 \\
CHAI-static & -4.0  & -4.3      & -3.7          & -2.5     & -0.8  \\
CHAI        & \bf{-2.0}  & \bf{-3.2}      & \bf{-0.5}          & \bf{0.3}      & \bf{0.1}   \\ \bottomrule
\end{tabular}}
\vspace{-15pt}
\end{table}


\begin{table}[t]
\caption{Accuracy on \llama-33B}
\label{tab:llama33bdelta}
\resizebox{\linewidth}{!}{
\begin{tabular}{@{}llllll@{}}
\toprule
Method      & PIQA  & HellaSwag & Arc-Challenge & Arc-Easy & BoolQ  \\ \midrule
MHA         & 82.1  & 82.8      & 57.8          & 80.0     & 83.1   \\ \midrule
DejaVu-10\% & -0.7  & 0.1       & \bf{-0.2}         & -0.6     & -0.2   \\
DejaVu-30\% & -9.3  & -24.4     & -17.91        & -12.4    & -12.2  \\
DejaVu-50\% & -27.6 & -43.2     & -24.6         & -37.6    & -21.2  \\
SpAtten     & -31.9 & -44.1     & -26.4         & -40.3    & -34.55 \\
CHAI-static & -0.5  & -0.2      & -1.3          & -3.7     & -1.5   \\
CHAI        & \bf{0}     &  \bf{-0.14}    & -0.21        & \bf{0.9}      & \bf{-0.04} \\ \bottomrule
\end{tabular}}
\vspace{-15pt}
\end{table}
\subsection{Reduction in K,V Cache Memory Requirement}
\begin{figure}[t]
    \centering
    \includegraphics[width=0.8\linewidth]{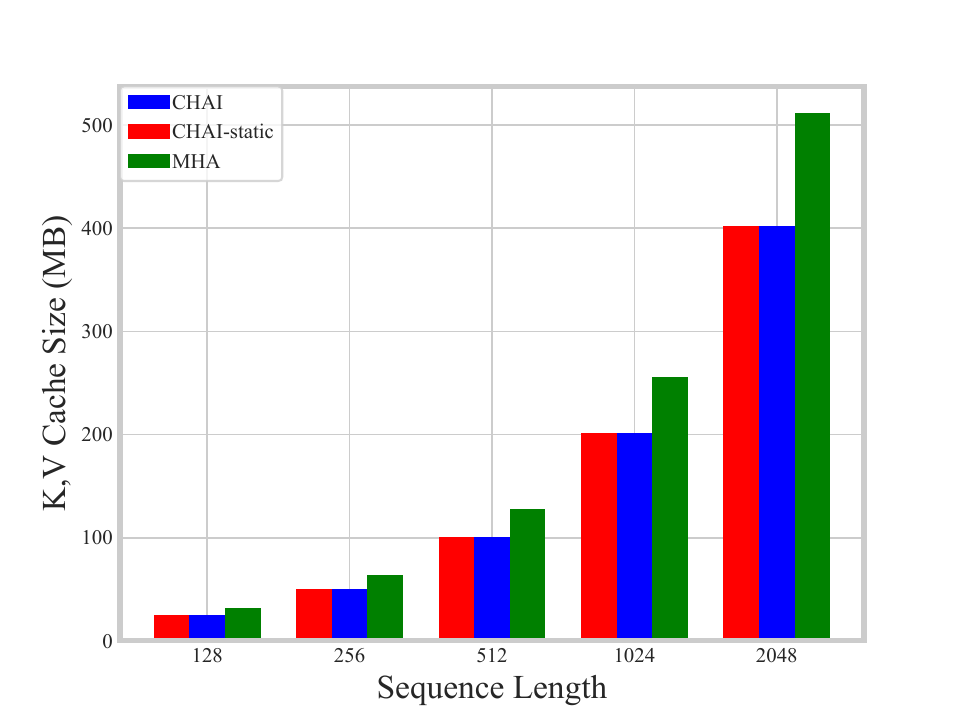}
    \vspace{-10pt}
    \caption{\small{\textbf{Memory Savings}: We observed that for \llama-7B \system provides memory savings of up to 21.4\%.}}
    \label{fig:kv_reduction}
    \vspace{-12pt}
\end{figure}

In this section, we study the memory capacity reduction achieved by use of \system due to reduction in K,V cache size.
In Figure~\ref{fig:kv_reduction}, we show that for \llama-7B \system reduces the size of K,V cache by up to 21.4\% compared to MHA. 
Even for comparatively small models like \llama-7B, the size of the K,V cache for a sequence length of 2048 is around 1.2 GB, while around 12 GB is used for the model weights. A reduction in K,V cache size can enable use of larger context length or serving more requests.
We would also like to note that as shown in Figure~\ref{fig:cha_schematic}, \system only removes the keys associated with redundant heads and keeps all the value vectors.

\begin{figure}[t]
    \centering
    \begin{subfigure}[t]{0.23\textwidth}
    \includegraphics[width=\textwidth]{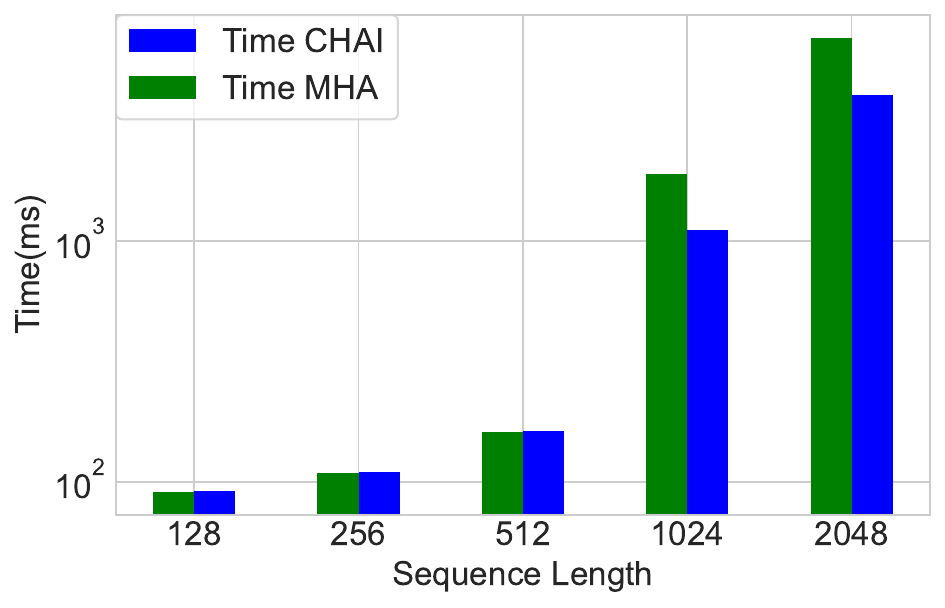}
    \caption{\small{\textbf{Time to first token:} We observe speedups of upto 1.73$\times$ for sequence length of 2048. }}
    \end{subfigure}\quad
    \begin{subfigure}[t]{0.23\textwidth}
    \includegraphics[width=\textwidth]{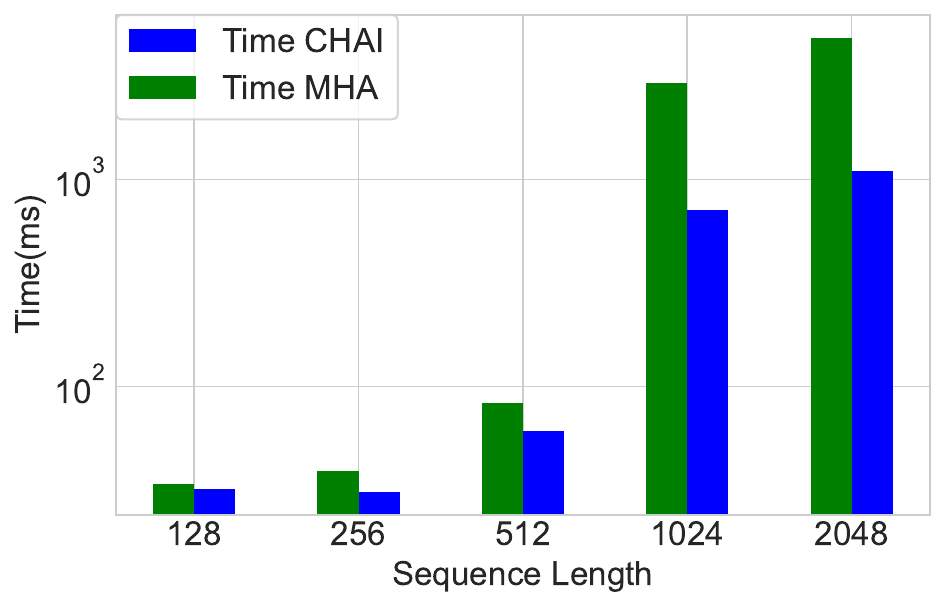}
    \caption{\small{\textbf{Time to next token:} We observe a speedup of upto $5\times$ for sequence length of 2048. }}
    \end{subfigure}
    \vspace{-10pt}
    \caption{\small{\textbf{Latency Analysis:} We observe that the speedups provided by \system increases as the sequence length becomes larger. Even for a comparatively small model like \llama-7B we observe speedups of up to 1.73$\times$ for a large sequence length.}}
    \label{fig:time_to_next_token}
    \vspace{-15pt}
\end{figure}
\vspace{-5pt}
\subsection{End-to-End Latency}
Next, we evaluate time to first token and time to next token comparing it with MHA. These are two standard metrics used for evaluation of an LLM. Time to first token evaluates the time for generating a first token given a new context. Time to first token accounts for generating K,V caches for all the tokens in the context. Whereas time to next token evaluates the time for generating the next token, assuming the K,V caches for all internal tokens is available. 

\textbf{Time to first token.}
Next, in our experiments we compare the speedups provided by \system. In Figure~\ref{fig:time_to_next_token}-(a) for \llama-7B we show that our method provides speedup of up to $1.72\times$ on a sequence length of 2048.
The execution times represented in this figure accounts for the overhead of clustering in \system.


\textbf{Time to next token.}
Another metric for evaluation of LLMs is time to next token. We do not account for the overhead of clustering in the case of time to next token. Our primary wins come from reducing compute and reducing memory bandwidth requirement for performing time to next token. 
Figure~\ref{fig:time_to_next_token}-(b) shows time to predict the next token for different sequence lengths. We observe that \system provides a speedup of over $5\times$ for a sequence length of 2048.

Unfortunately, we are not able to compare times with \dejavu as the authors have not released the specialized kernels used for realizing the speedups on hardware~\cite{githubdejavu},
thus inhibiting a runtime comparison. However, we believe it is unlikely that at less than 10\% sparsity which is needed by \dejavu to get comparable accuracy to MHA, it will yield high speedups~\cite{hooker2021hardware}.
We would like to highlight that because of performing dense computations, unlike \dejavu, \system does not need custom GPU kernels. Further, \system's speedup benefits are independent of the framework used, because irrespective of implementation, \system directly reduces the complexity of MHA.

\begin{figure}[t]
    \centering
    \includegraphics[width=0.6\linewidth]{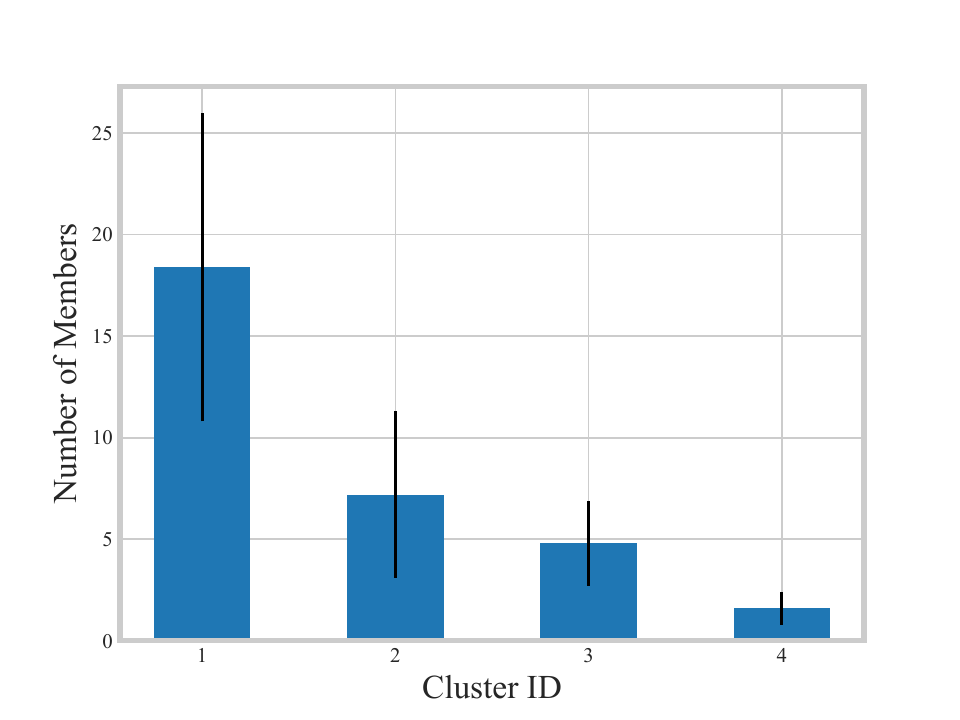}
    \vspace{-10pt}
    \caption{\small{\textbf{Cluster Distribution}: We observe that number of heads within the cluster is quite skewed. We often observe one or two large clusters, while the remaining heads in the cluster. }}
    \label{fig:cluster_distribution}
    \vspace{-10pt}
\end{figure}
\vspace{-5pt}
\subsection{Additional Experiments}
Next we perform additional studies on our algorithm.

\textbf{Pruning K, Q and V.}
In \system, we prune only the Key and Query portion of an attention head leaving the Value vector intact. Next, we study how accuracy changes if we remove the value vector as well. 
To perform this experiment we chose to reuse the value vector generated by the chosen head.
In Table~\ref{tab:qkv_comparison}, we show how reusing the full head (Query, Key and Value vector) lead to additional loss in accuracy. This shows that for smaller networks like \llama it might be hard to remove the whole head in Multi-Head Attention.

\begin{table}[t]
\centering
\caption{Pruning Both Q,K,V}
\resizebox{0.7\linewidth}{!}{
\begin{tabular}{@{}llll@{}}
\toprule
              & \system & \system-QKV & MHA  \\ \midrule
Arc-Challenge & 47.0         & 41.29         & 47.5 \\
PIQA          & 77.8         & 61.93         & 79.8 \\ \bottomrule
\end{tabular}
}
\label{tab:qkv_comparison}
\vspace{-15pt}
\end{table}

\textbf{Cluster Distribution.}
Figure~\ref{fig:cluster_distribution} shows the distribution across clusters for Layer-18 on \llama-7B for different 1024 samples of C4 dataset. We observe that typically for LLMs majority of heads can be grouped into a single head.

\vspace{-5pt}
\section{Conclusion}
In this work, we present \system, an efficient runtime method which identifies attention heads giving similar scores. Using this method we reduce overhead of Multi-Head Attention by clustering the correlated heads and computing attention scores only for heads which lead to disparate attention scores. Our evaluation shows that with minor accuracy loss system can speedup inference by up to $1.73\times$.

\bibliographystyle{icml2023}
\bibliography{ref}
\clearpage
\appendix

\onecolumn
\section{Additional Plots}
\label{sec:additional_plots}

\subsection{Accuracy and Inference Time Trade-off}

\begin{figure}[htb!]
    \centering
    \includegraphics[width=0.4\linewidth]{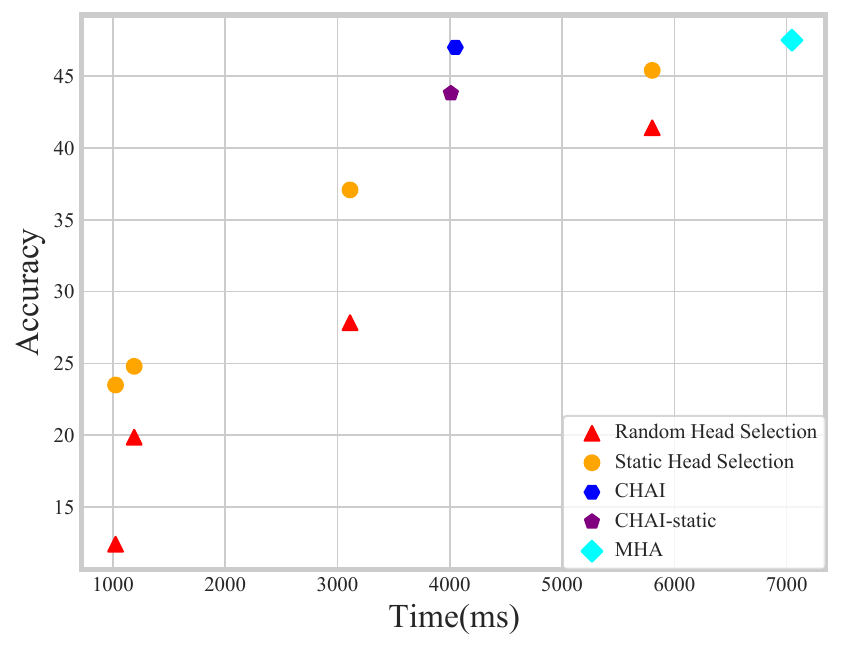}
    \vspace{-10pt}
    \caption{\small{\textbf{Accuracy vs Inference Time for \llama-7B:} We study various methods of clustering attention heads, and plot the runtime for sequence length of 2048. For random head selection we randomly choose heads to combine together in increasing number of 4, 8, 16 and 24. For Static Head selection we choose the heads in increasing order of 4,8,16, and 24 based on activation analysis of activation on C4 dataset~\cite{c4}.}}
    \label{fig:pareto_optimality_time}
    \vspace{-15pt}
\end{figure}

\subsection{OPT-66B Activation Plots}
From Figure~\ref{fig:opt1} to Figure~\ref{fig:optlast} shows the activation plots for all layers of OPT. We consistently observe that for this examples there is high amount of.
\subsection{\llama-7B Activation Plots}
\begin{figure}
\renewcommand\thesubfigure{\roman{subfigure}}
\begin{center}
    \begin{subfigure}[t]{0.33\textwidth}
    \includegraphics[width=\textwidth]{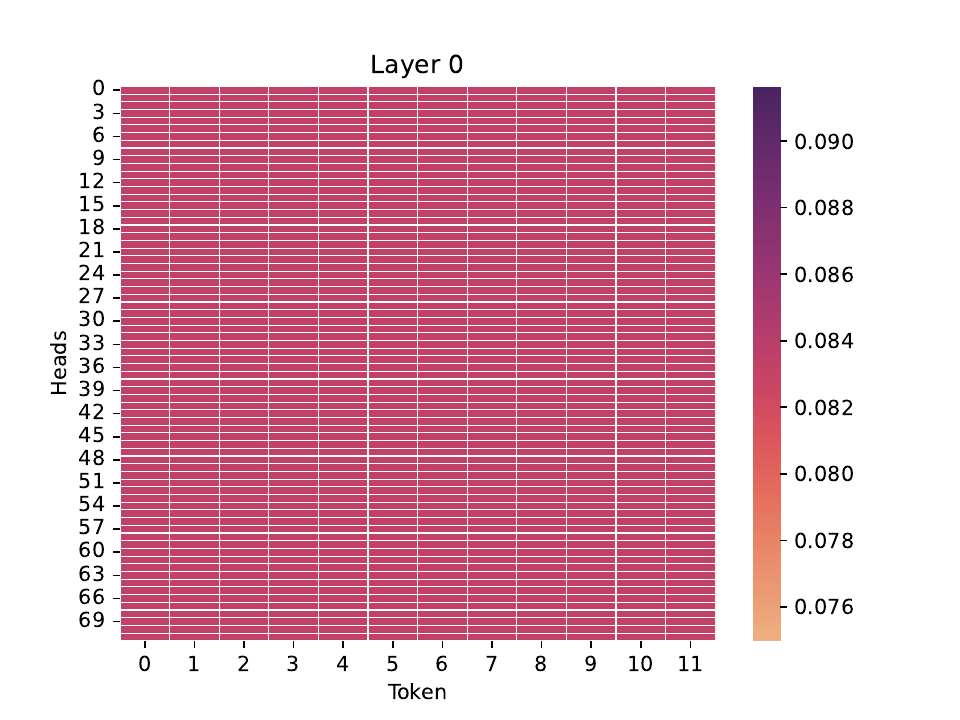}
    \caption{\small{Layer 0}}
    \end{subfigure}\
    \begin{subfigure}[t]{0.33\textwidth}
        \includegraphics[width=\textwidth]{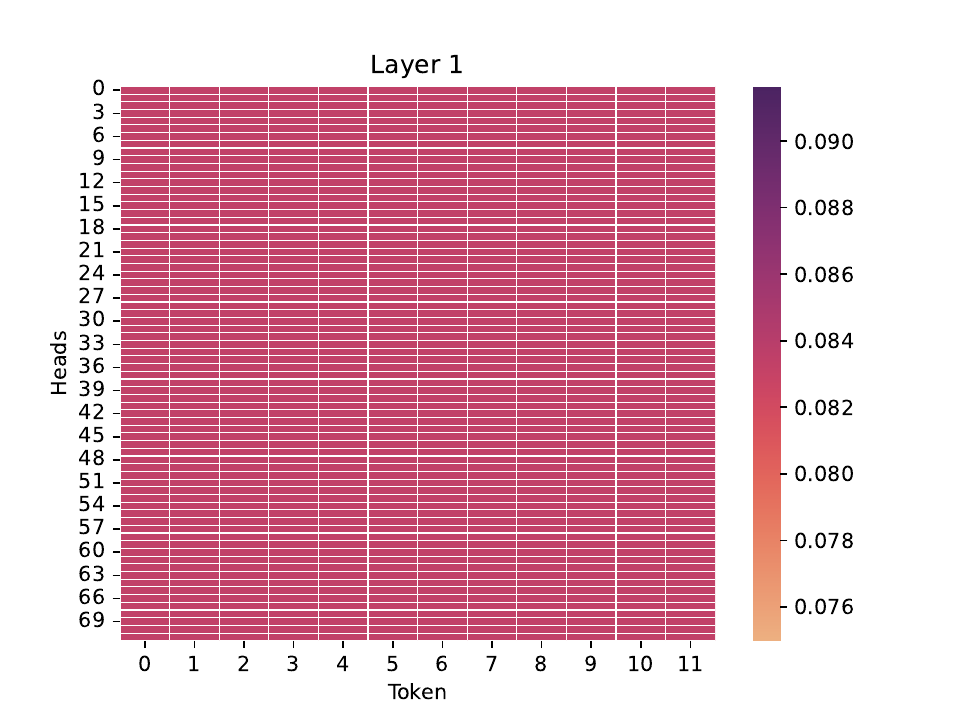}
        \caption{\small{Layer 1}}
    \end{subfigure}
    \begin{subfigure}[t]{0.33\textwidth}
        \includegraphics[width=\textwidth]{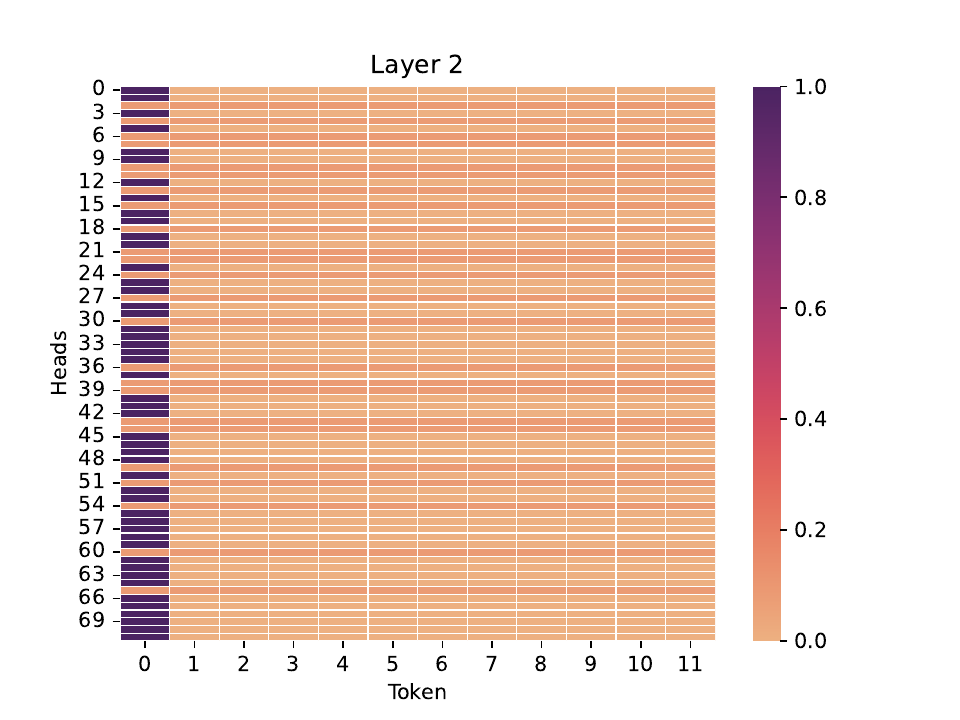}
        \caption{\small{Layer 2}}
    \end{subfigure}
    \begin{subfigure}[t]{0.33\textwidth}
        \includegraphics[width=\textwidth]{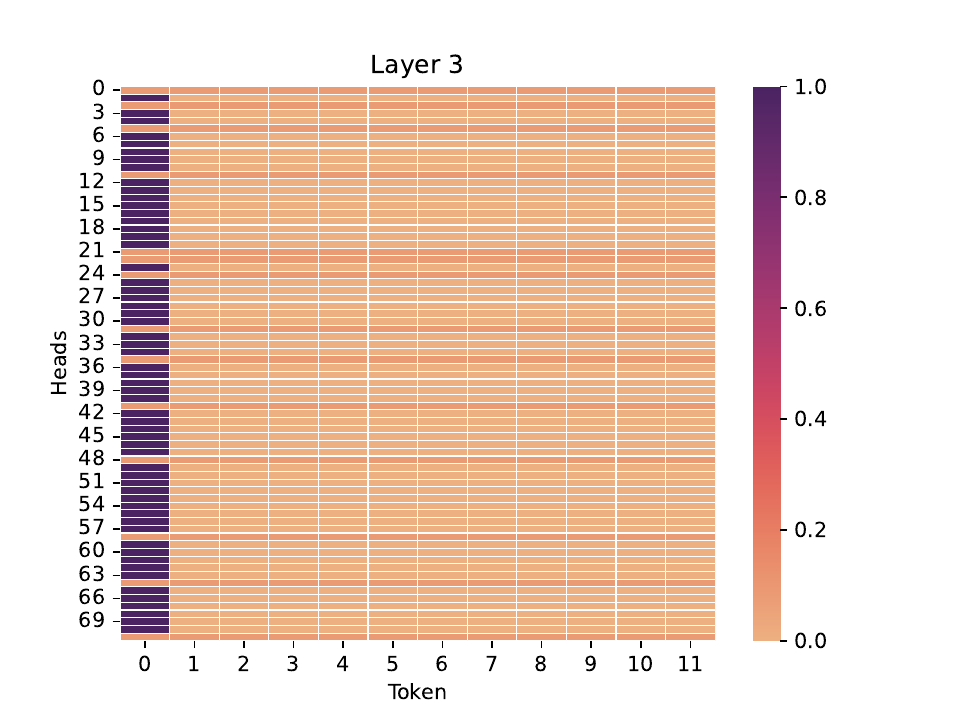}
        \caption{\small{Layer 3}}
    \end{subfigure}
      \begin{subfigure}[t]{0.33\textwidth}
        \includegraphics[width=\textwidth]{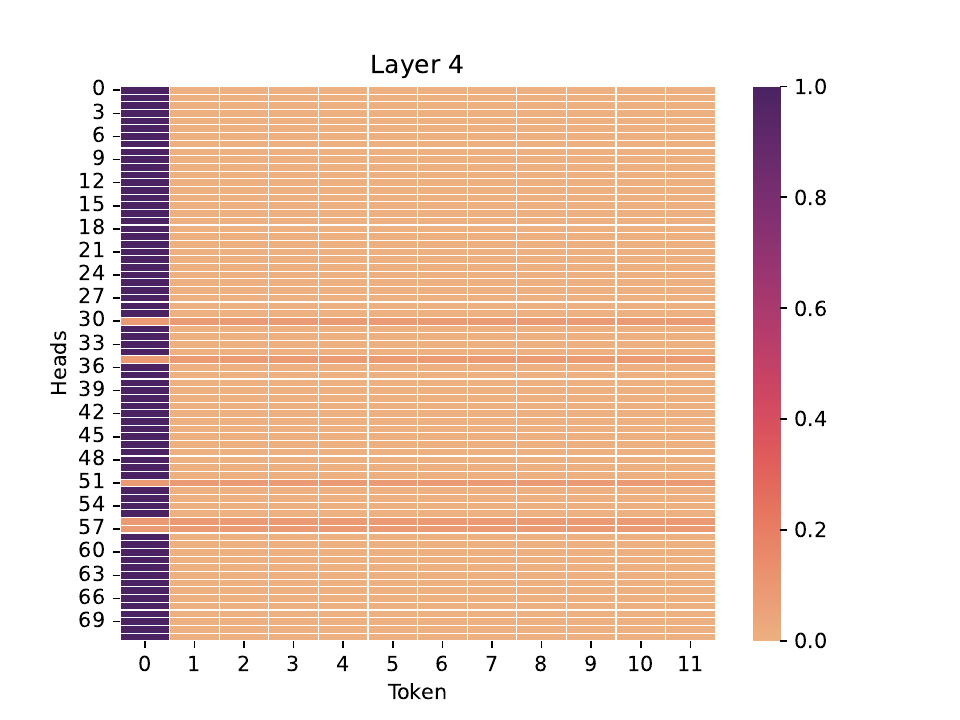}
        \caption{\small{Layer 4}}
    \end{subfigure}
    \begin{subfigure}[t]{0.33\textwidth}
        \includegraphics[width=\textwidth]{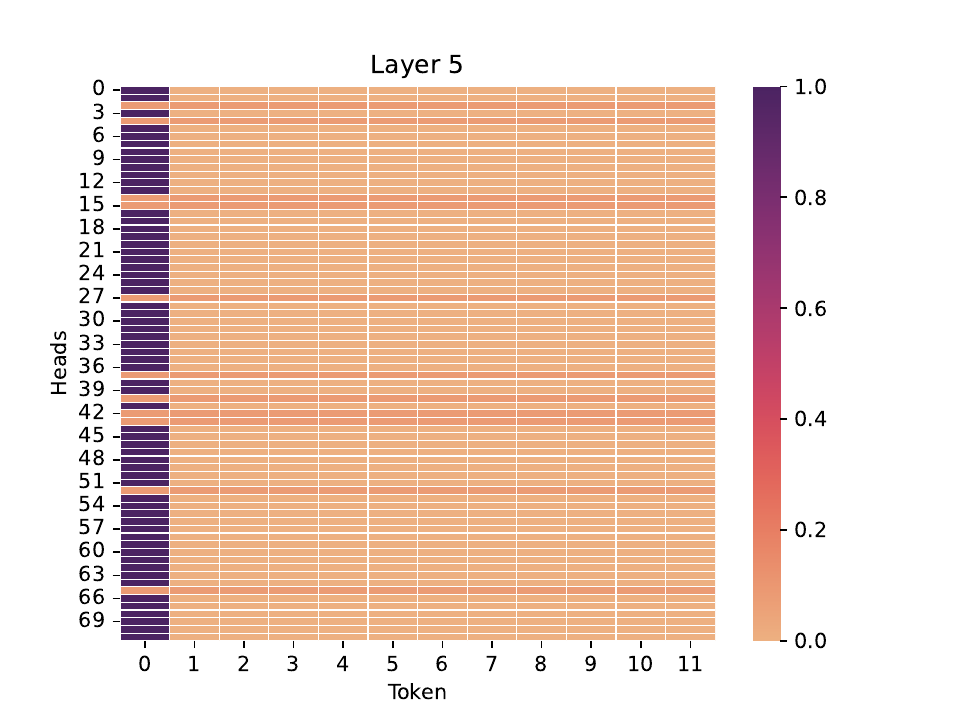}
        \caption{\small{Layer 5}}
    \end{subfigure}
        \end{center}
\caption{Activations for \opt-66B}
\label{fig:opt1}
    \end{figure}
    \begin{figure}\ContinuedFloat
          \renewcommand\thesubfigure{\roman{subfigure}}
    \begin{center}
    \begin{subfigure}[t]{0.33\textwidth}
        \includegraphics[width=\textwidth]{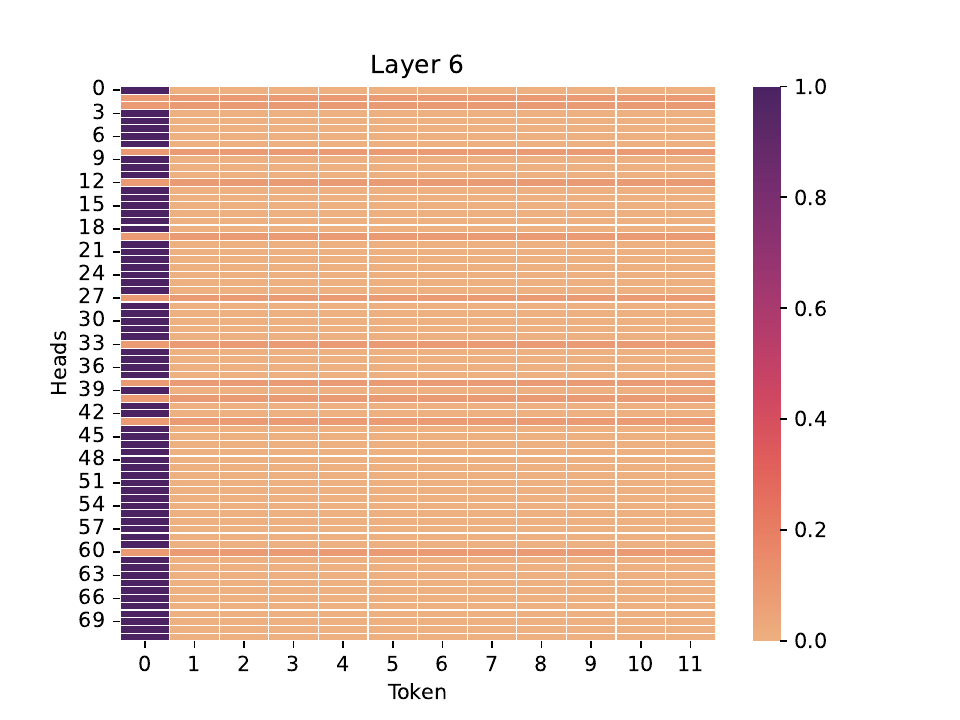}
        \caption{\small{Layer 6}}
    \end{subfigure}
\begin{subfigure}[t]{0.33\textwidth}
        \includegraphics[width=\textwidth]{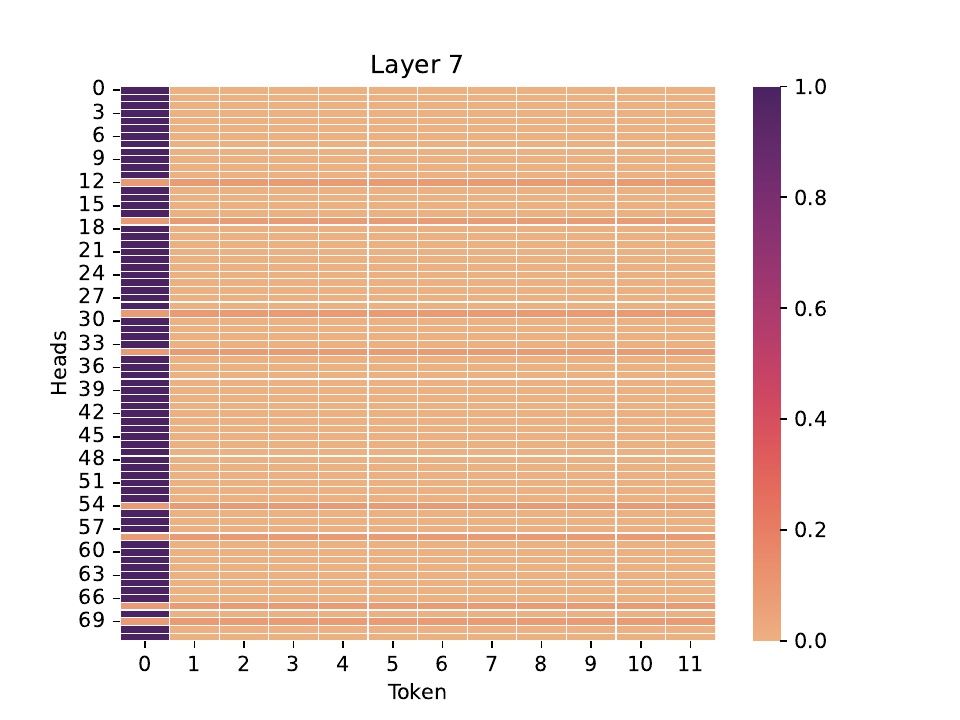}
        \caption{\small{Layer 7}}
    \end{subfigure}
    \begin{subfigure}[t]{0.33\textwidth}
        \includegraphics[width=\textwidth]{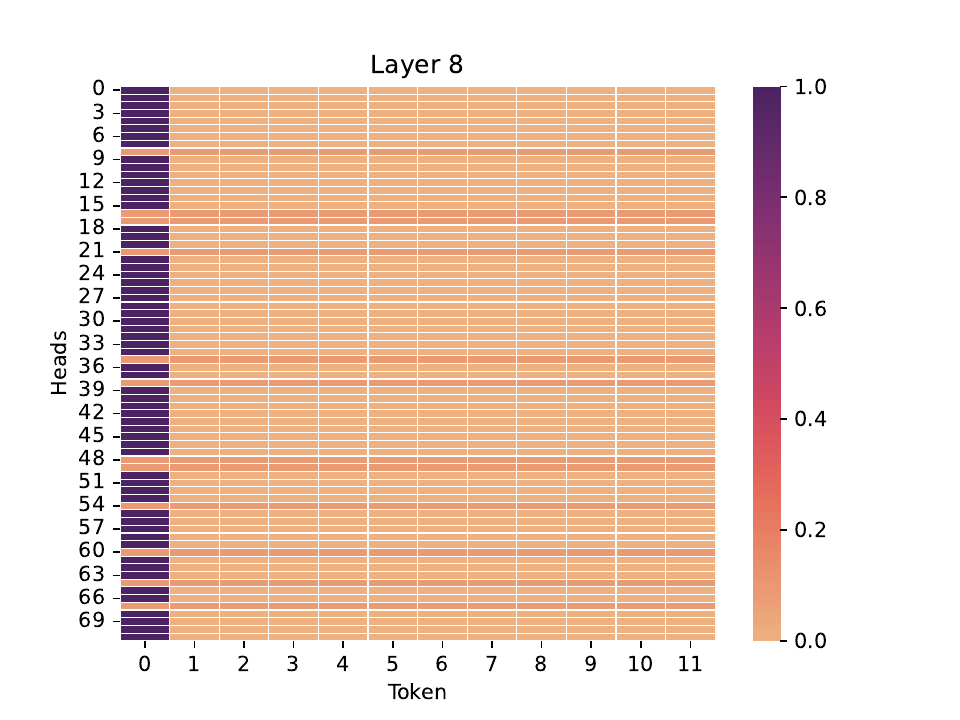}
        \caption{\small{Layer 8}}
    \end{subfigure}
    \begin{subfigure}[t]{0.33\textwidth}
        \includegraphics[width=\textwidth]{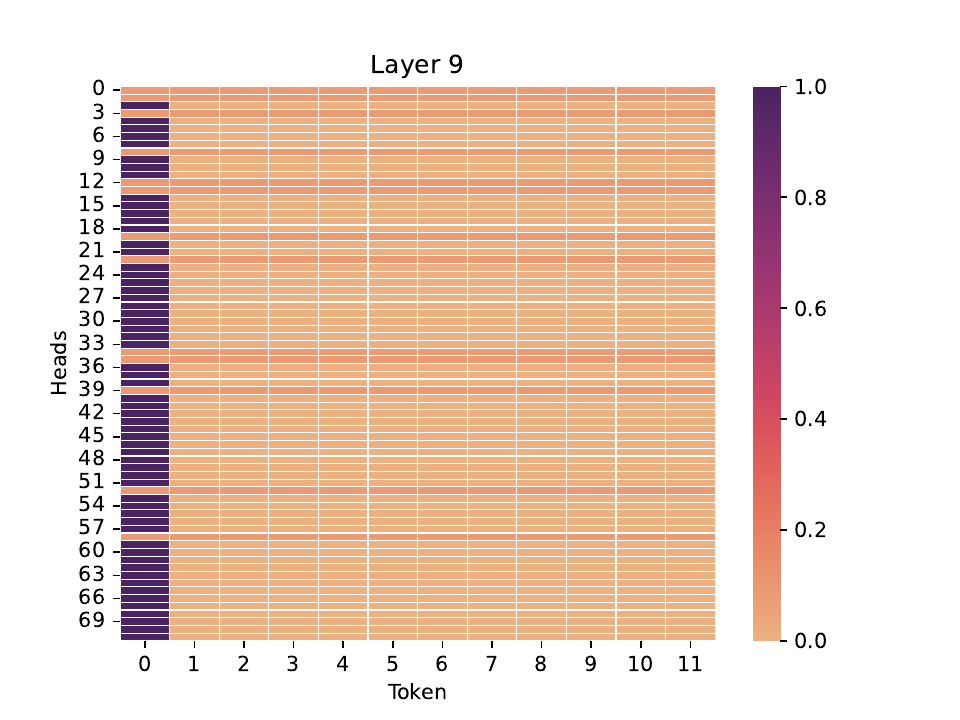}
        \caption{\small{Layer 9}}
    \end{subfigure}
    \begin{subfigure}[t]{0.33\textwidth}
        \includegraphics[width=\textwidth]{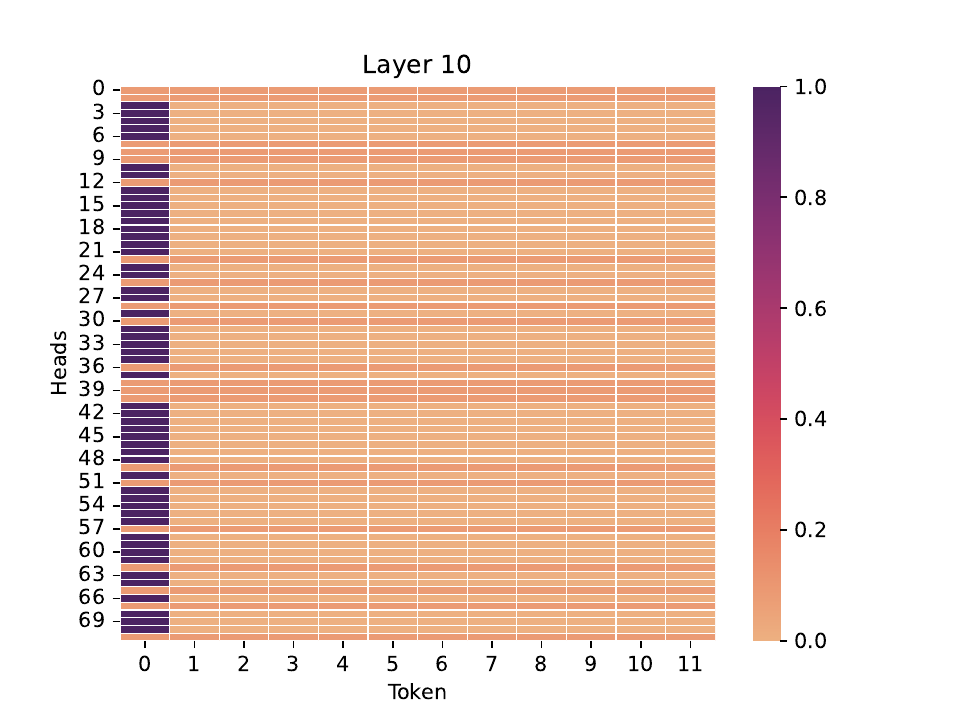}
        \caption{\small{Layer 10}}
    \end{subfigure}
    \begin{subfigure}[t]{0.33\textwidth}
        \includegraphics[width=\textwidth]{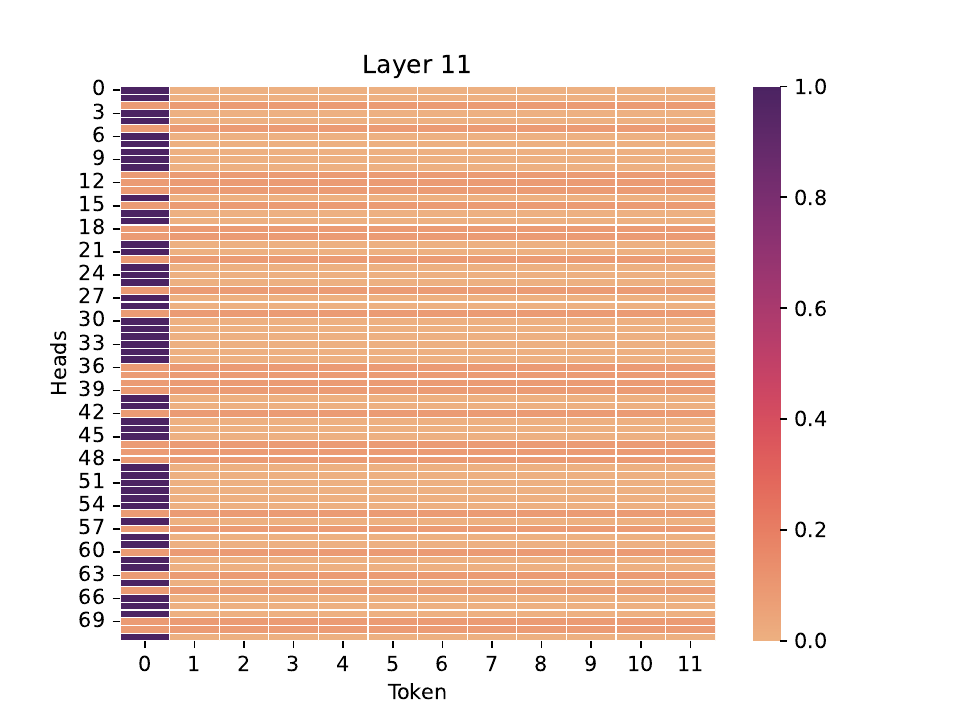}
        \caption{\small{Layer 11}}
    \end{subfigure}
    \end{center}
    \caption{Activations for \opt-66B}
    \end{figure}
    \begin{figure}[t]\ContinuedFloat
    \renewcommand\thesubfigure{\roman{subfigure}}
    \begin{center}
    \begin{subfigure}[t]{0.33\textwidth}
        \includegraphics[width=\textwidth]{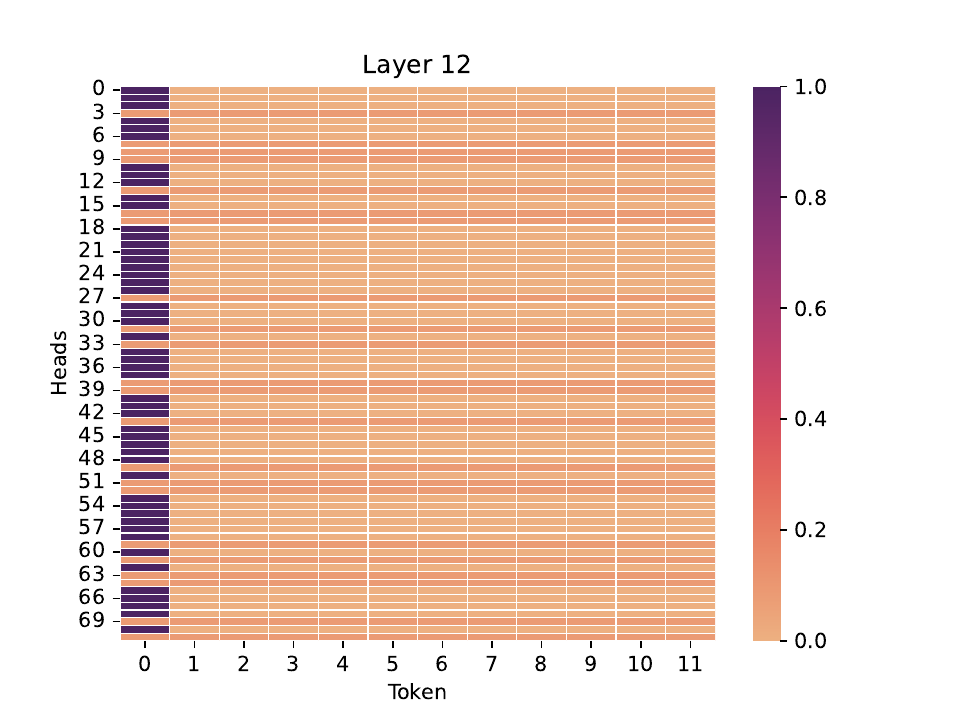}
        \caption{\small{Layer 12}}
    \end{subfigure}
    \begin{subfigure}[t]{0.33\textwidth}
        \includegraphics[width=\textwidth]{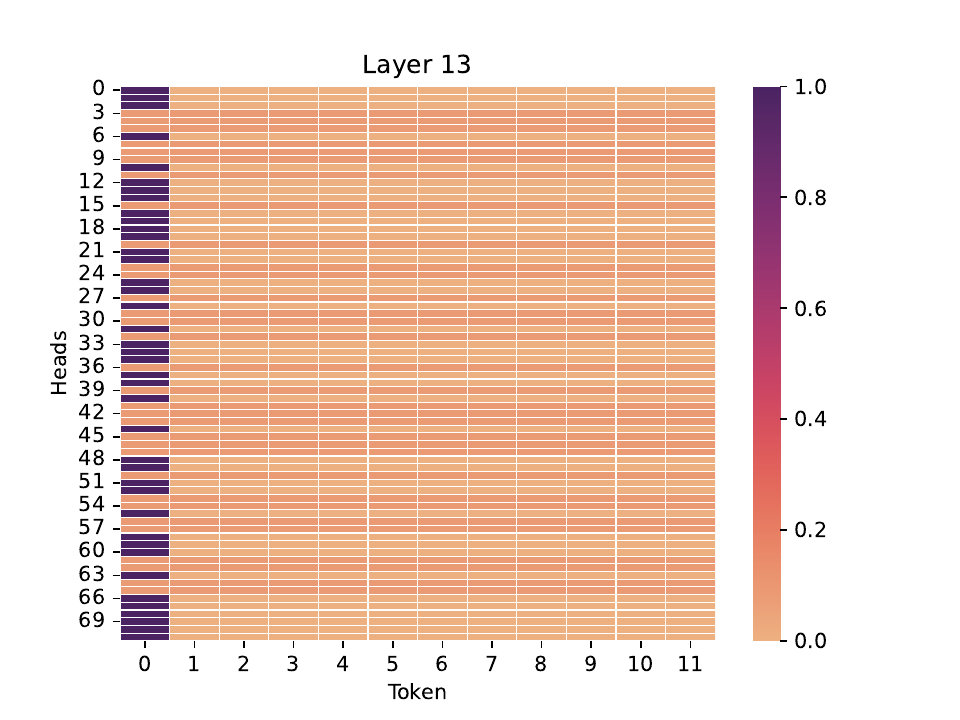}
        \caption{\small{Layer 13}}
    \end{subfigure}
    \begin{subfigure}[t]{0.33\textwidth}
        \includegraphics[width=\textwidth]{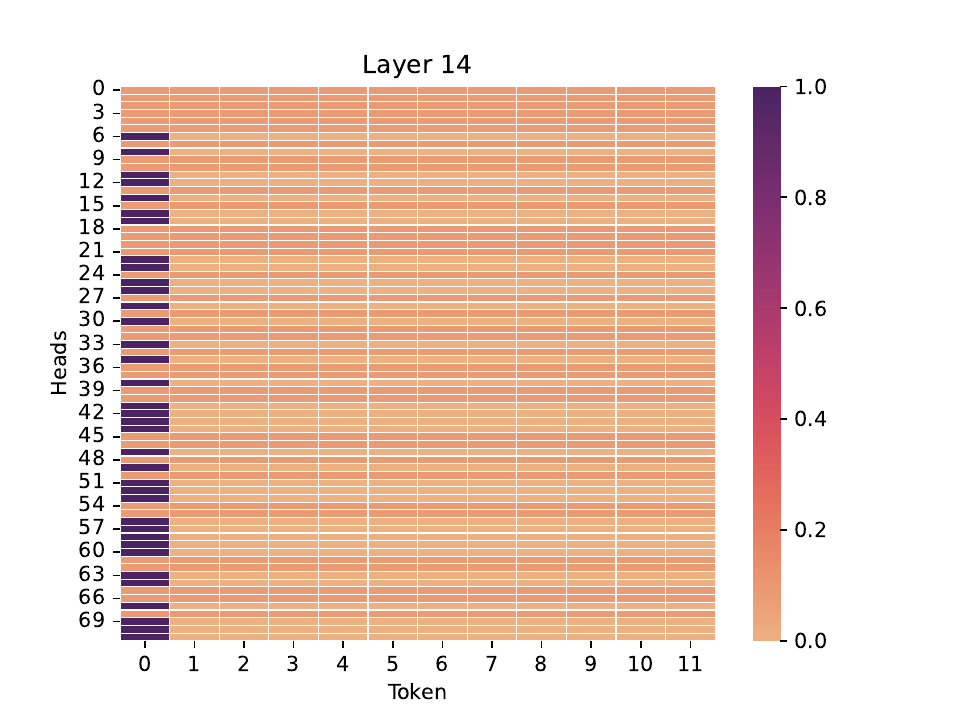}
        \caption{\small{Layer 14}}
    \end{subfigure}
    \begin{subfigure}[t]{0.33\textwidth}
        \includegraphics[width=\textwidth]{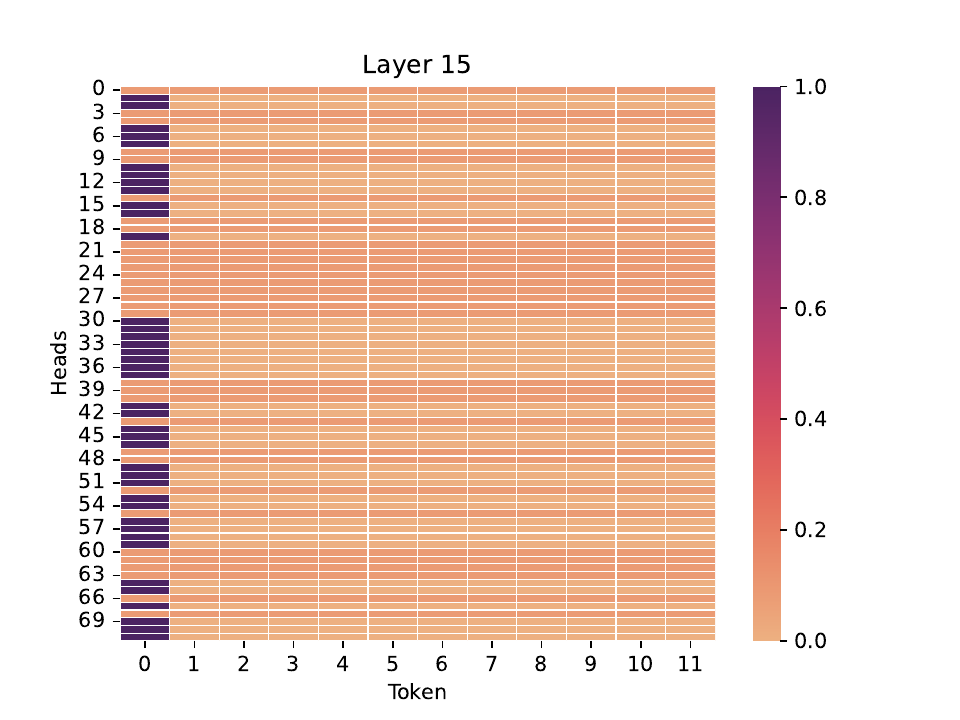}
        \caption{\small{Layer 15}}
    \end{subfigure}
    \begin{subfigure}[t]{0.33\textwidth}
        \includegraphics[width=\textwidth]{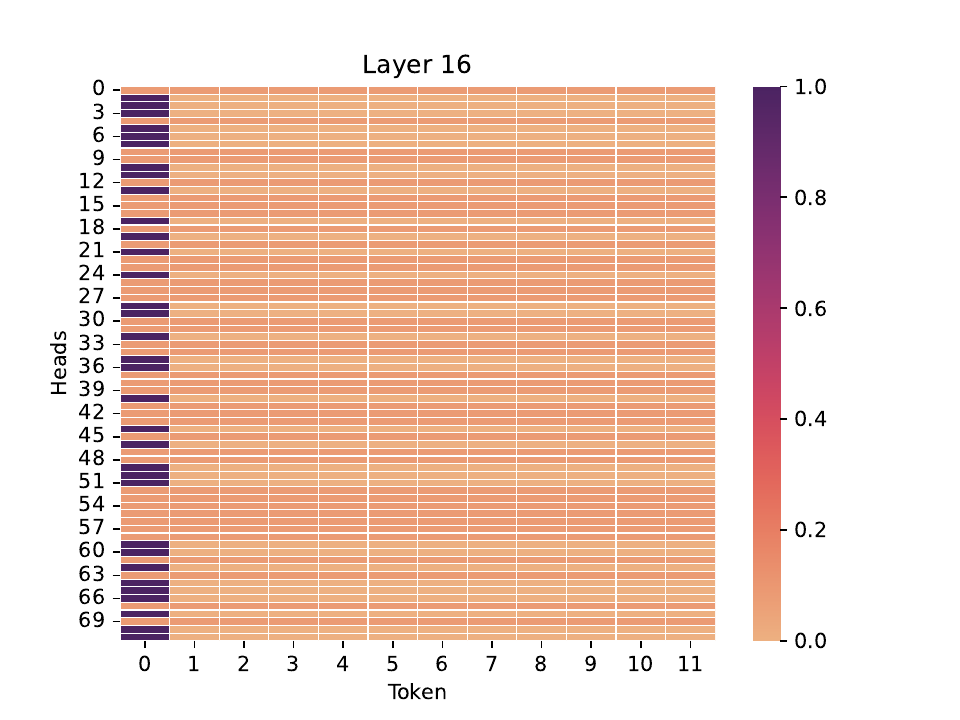}
        \caption{\small{Layer 16}}
    \end{subfigure}
\begin{subfigure}[t]{0.33\textwidth}
        \includegraphics[width=\textwidth]{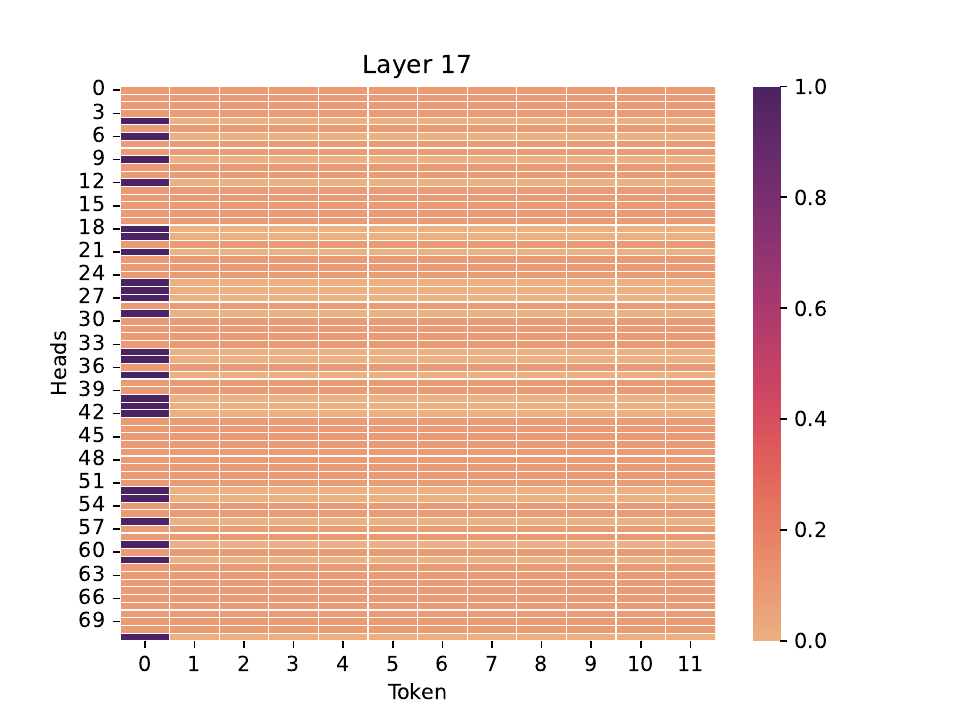}
        \caption{\small{Layer 17}}
    \end{subfigure}
    \begin{subfigure}[t]{0.33\textwidth}
        \includegraphics[width=\textwidth]{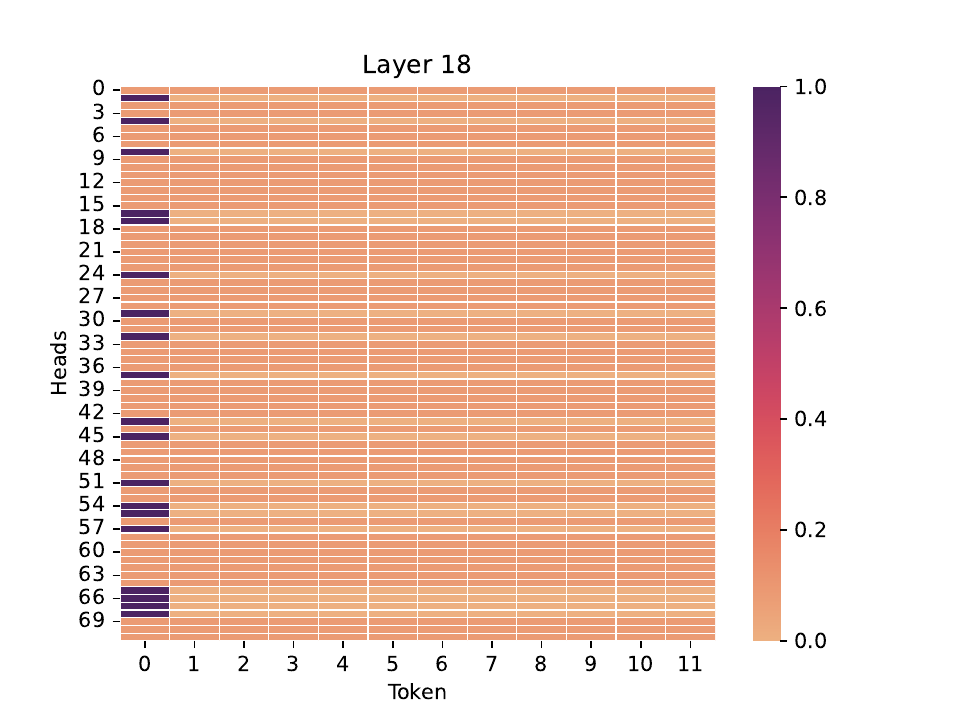}
        \caption{\small{Layer 18}}
    \end{subfigure}
    \begin{subfigure}[t]{0.33\textwidth}
        \includegraphics[width=\textwidth]{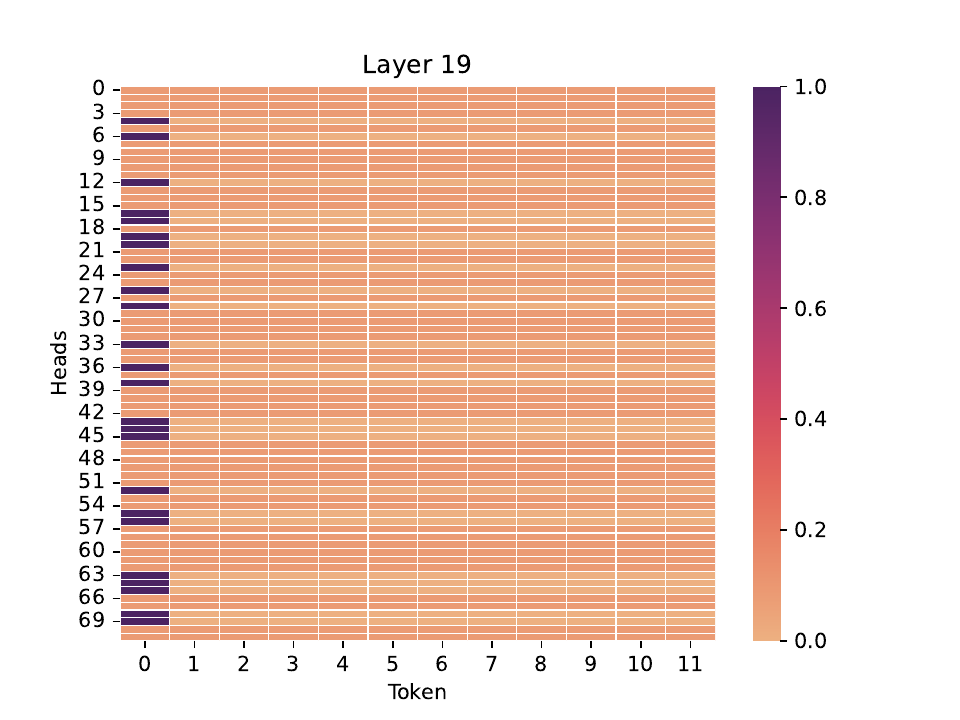}
        \caption{\small{Layer 19}}
    \end{subfigure}
    \begin{subfigure}[t]{0.33\textwidth}
        \includegraphics[width=\textwidth]{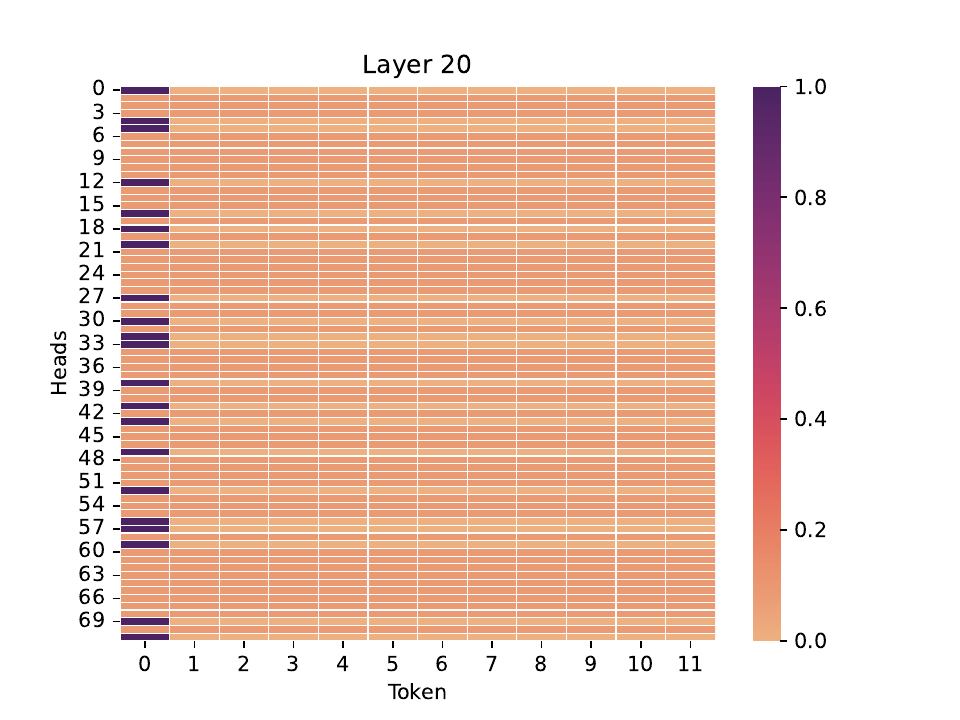}
        \caption{\small{Layer 20}}
    \end{subfigure}    
     \begin{subfigure}[t]{0.33\textwidth}
        \includegraphics[width=\textwidth]{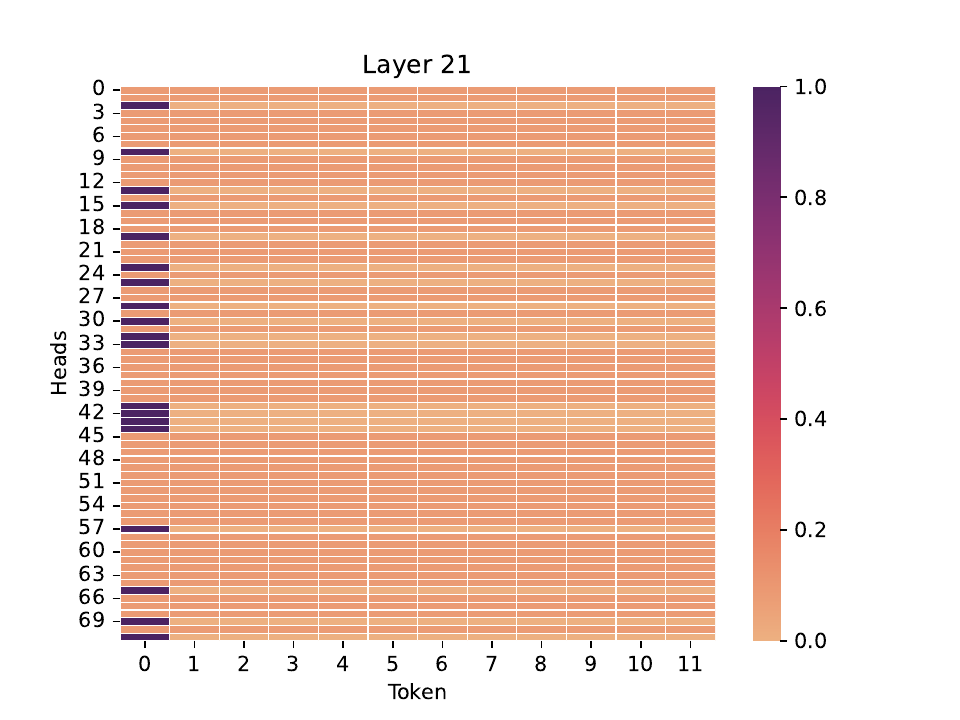}
        \caption{\small{Layer 21}}
    \end{subfigure}
 \begin{subfigure}[t]{0.33\textwidth}
        \includegraphics[width=\textwidth]{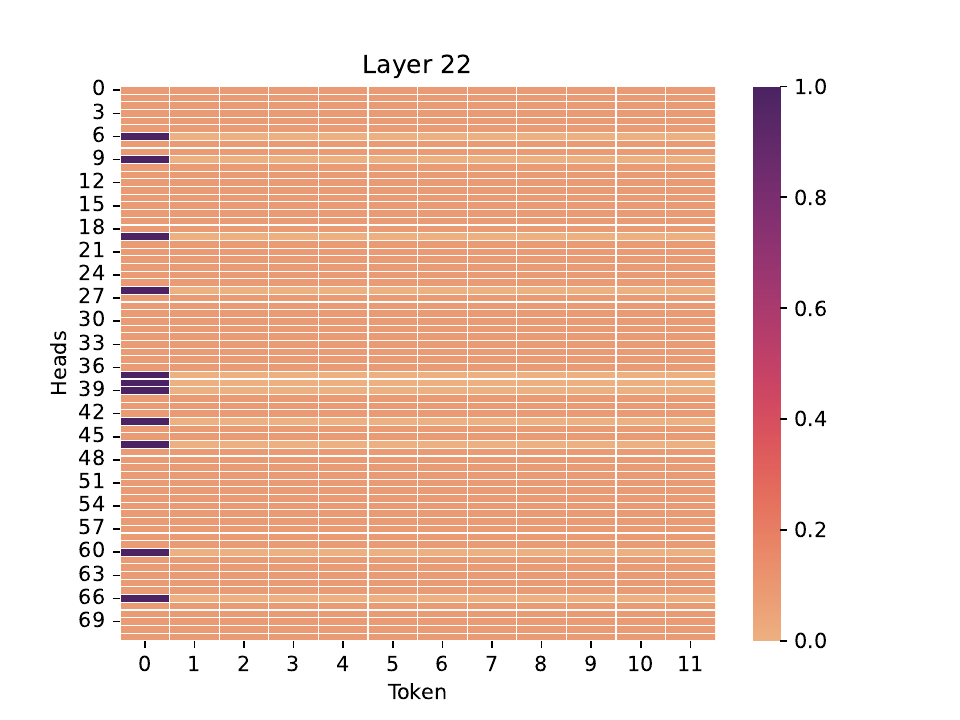}
        \caption{\small{Layer 22}}
    \end{subfigure}
     \begin{subfigure}[t]{0.33\textwidth}
        \includegraphics[width=\textwidth]{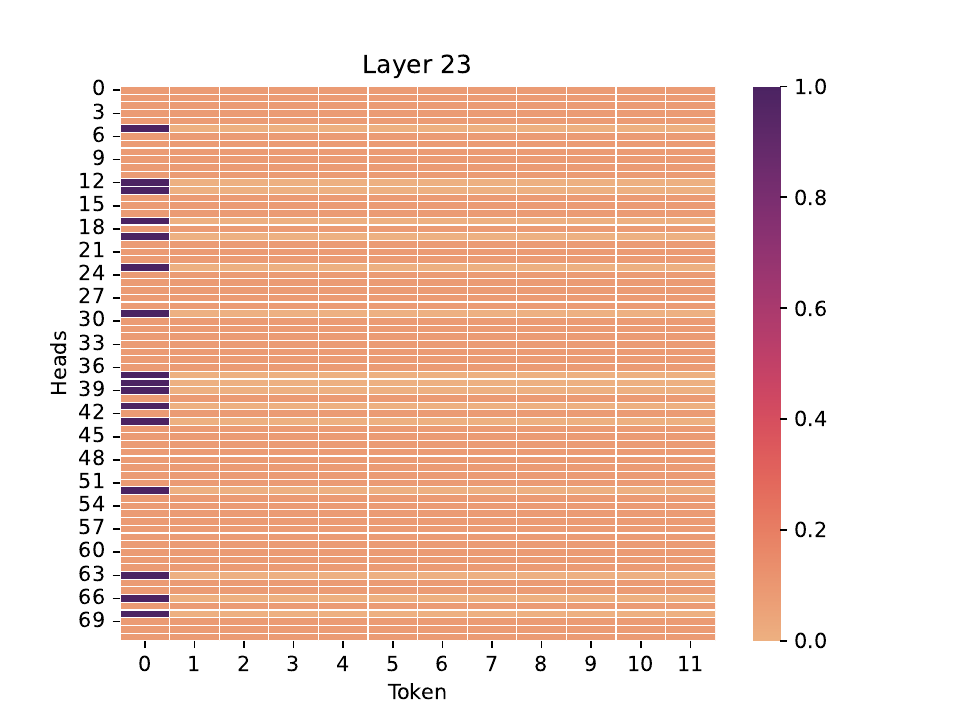}
        \caption{\small{Layer 20}}
        
    \end{subfigure}
             \end{center}
             \caption{Activations of \opt-66B}
    \end{figure}
    \begin{figure}[t]\ContinuedFloat
      \renewcommand\thesubfigure{\roman{subfigure}}
    \begin{center}   
 \begin{subfigure}[t]{0.33\textwidth}
        \includegraphics[width=\textwidth]{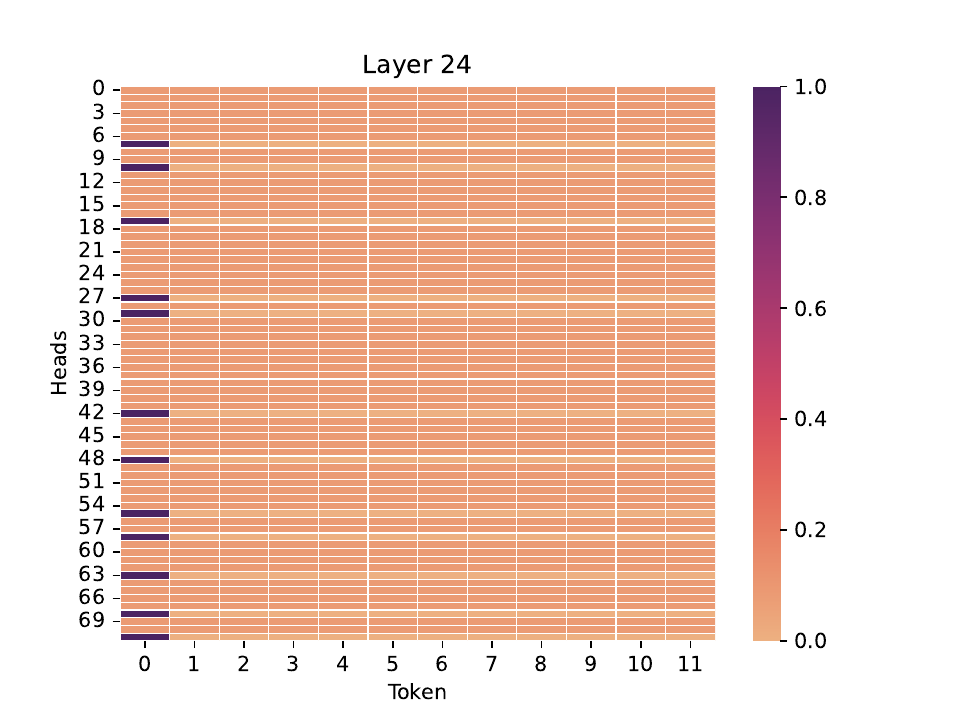}
        \caption{\small{Layer 24}}
    \end{subfigure}
 \begin{subfigure}[t]{0.33\textwidth}
        \includegraphics[width=\textwidth]{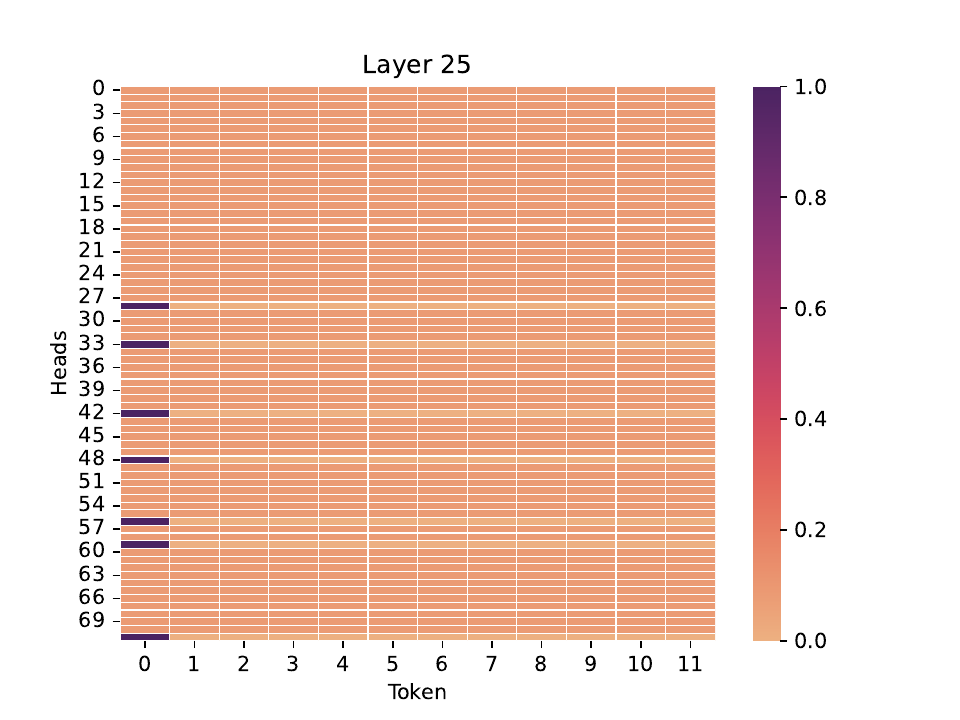}
        \caption{\small{Layer 25}}
    \end{subfigure}
     \begin{subfigure}[t]{0.33\textwidth}
        \includegraphics[width=\textwidth]{figure/opt_token_softmax_25_opt66.pdf}
        \caption{\small{Layer 25}}
    \end{subfigure}
    \begin{subfigure}[t]{0.33\textwidth}
        \includegraphics[width=\textwidth]{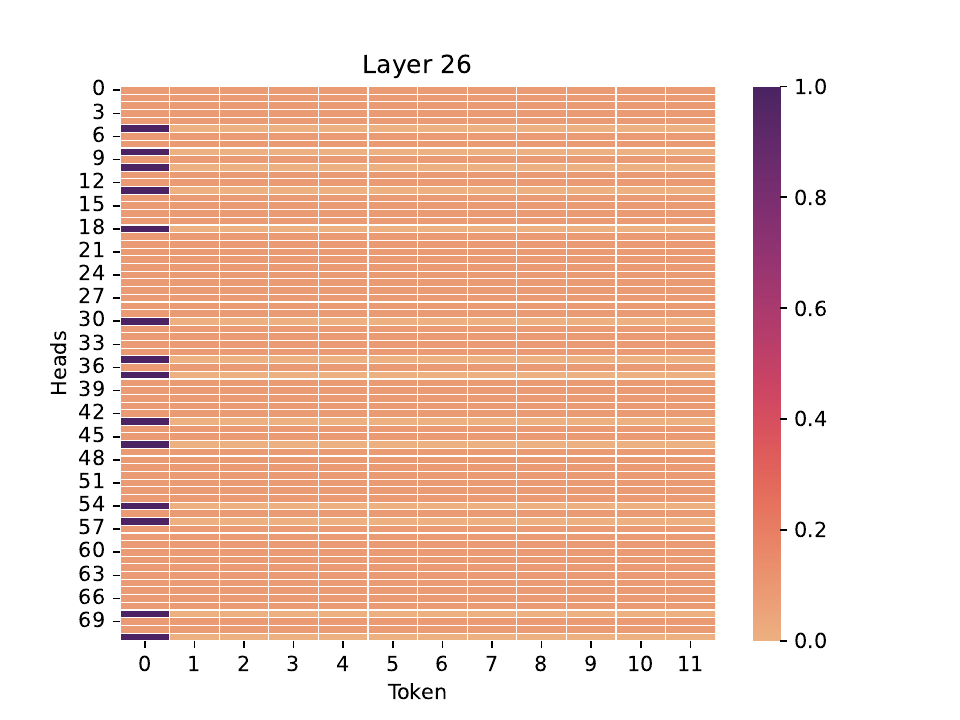}
        \caption{\small{Layer 26}}
    \end{subfigure}
        \begin{subfigure}[t]{0.33\textwidth}
        \includegraphics[width=\textwidth]{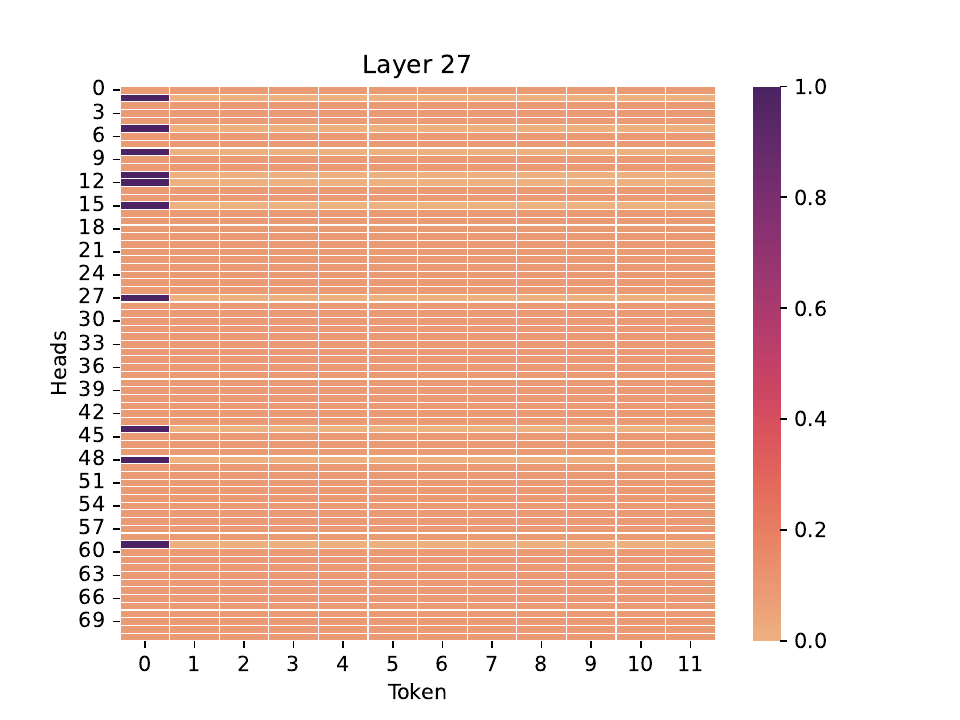}
        \caption{\small{Layer 27}}
    \end{subfigure}
        \begin{subfigure}[t]{0.33\textwidth}
        \includegraphics[width=\textwidth]{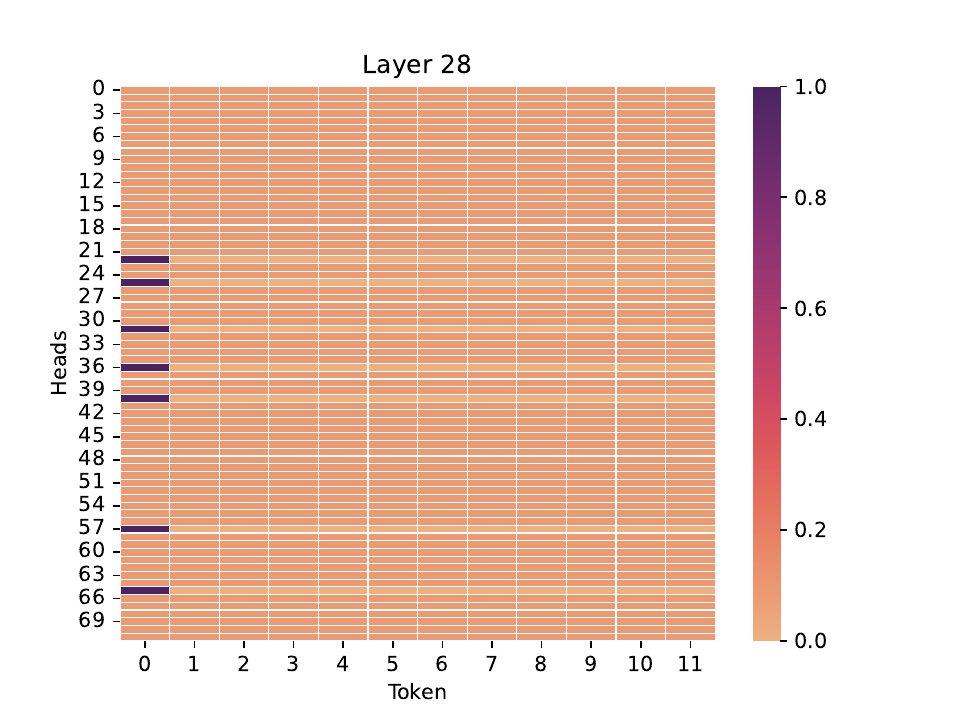}
        \caption{\small{Layer 28}}
    \end{subfigure}
        \begin{subfigure}[t]{0.33\textwidth}
        \includegraphics[width=\textwidth]{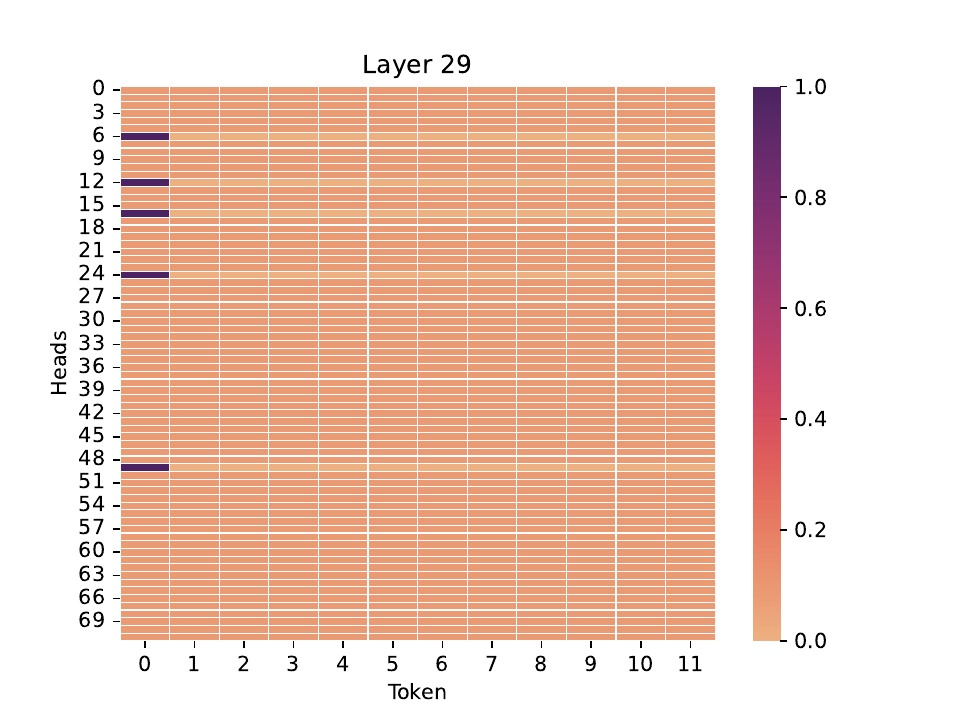}
        \caption{\small{Layer 29}}
    \end{subfigure}
        \begin{subfigure}[t]{0.33\textwidth}
        \includegraphics[width=\textwidth]{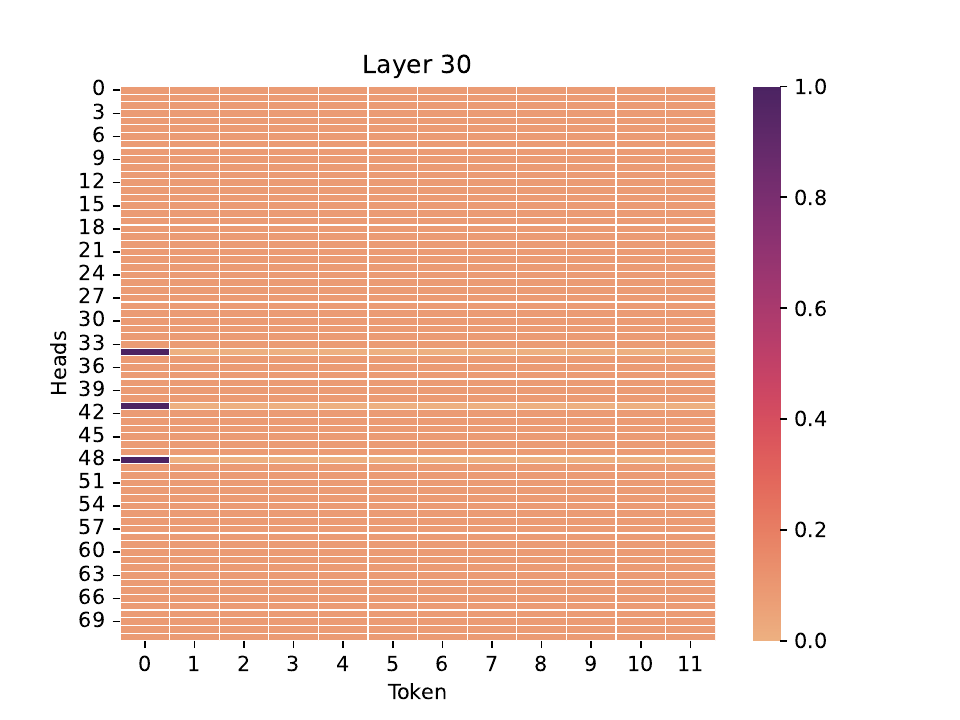}
        \caption{\small{Layer 30}}
    \end{subfigure}
        \begin{subfigure}[t]{0.33\textwidth}
        \includegraphics[width=\textwidth]{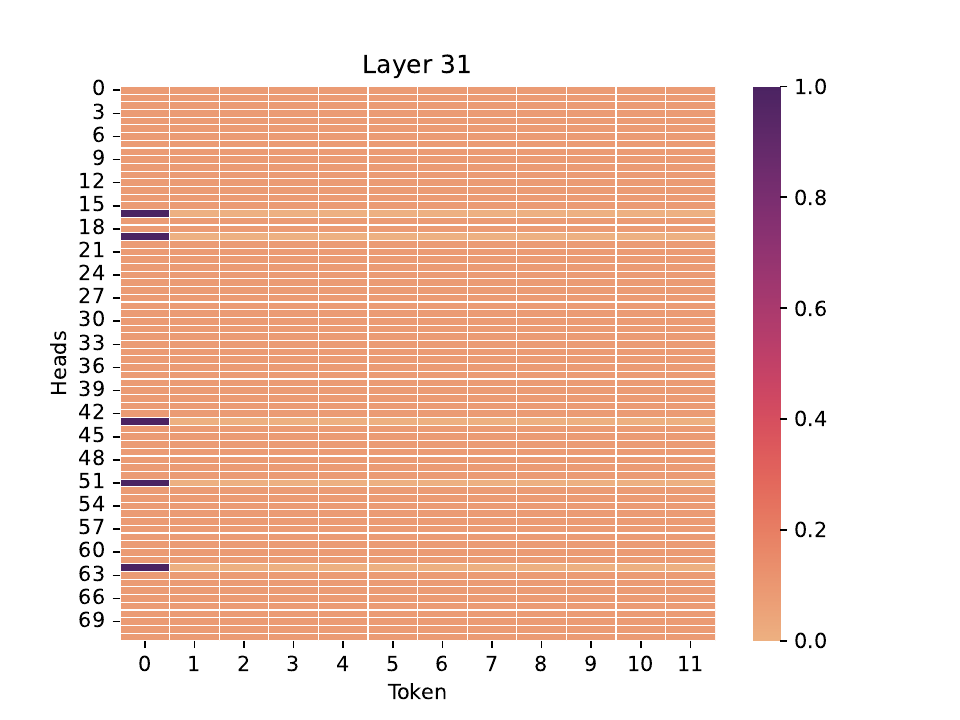}
        \caption{\small{Layer 31}}
    \end{subfigure}
    \begin{subfigure}[t]{0.33\textwidth}
        \includegraphics[width=\textwidth]{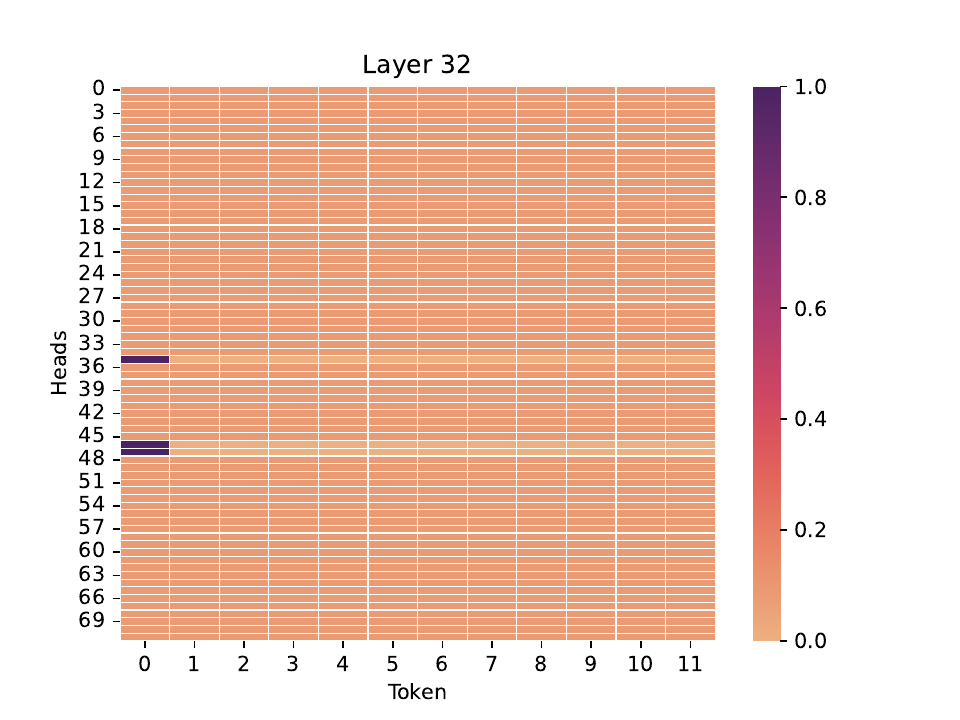}
        \caption{\small{Layer 32}}
    \end{subfigure}
    \begin{subfigure}[t]{0.33\textwidth}
        \includegraphics[width=\textwidth]{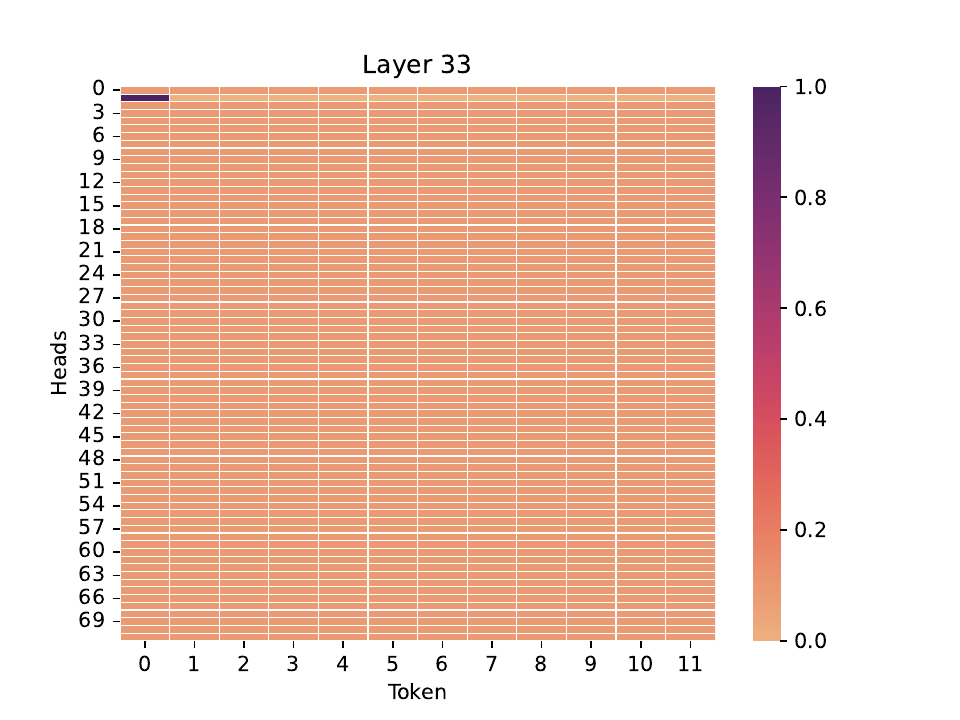}
        \caption{\small{Layer 33}}
    \end{subfigure}
    \begin{subfigure}[t]{0.33\textwidth}
        \includegraphics[width=\textwidth]{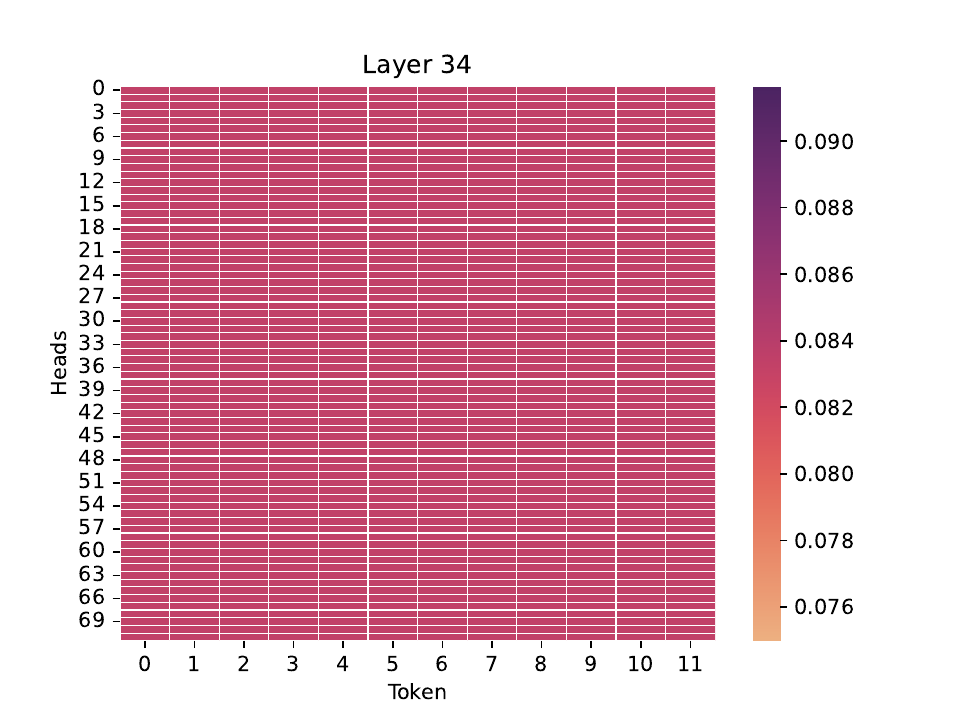}
        \caption{\small{Layer 34}}
    \end{subfigure}
    \begin{subfigure}[t]{0.33\textwidth}
        \includegraphics[width=\textwidth]{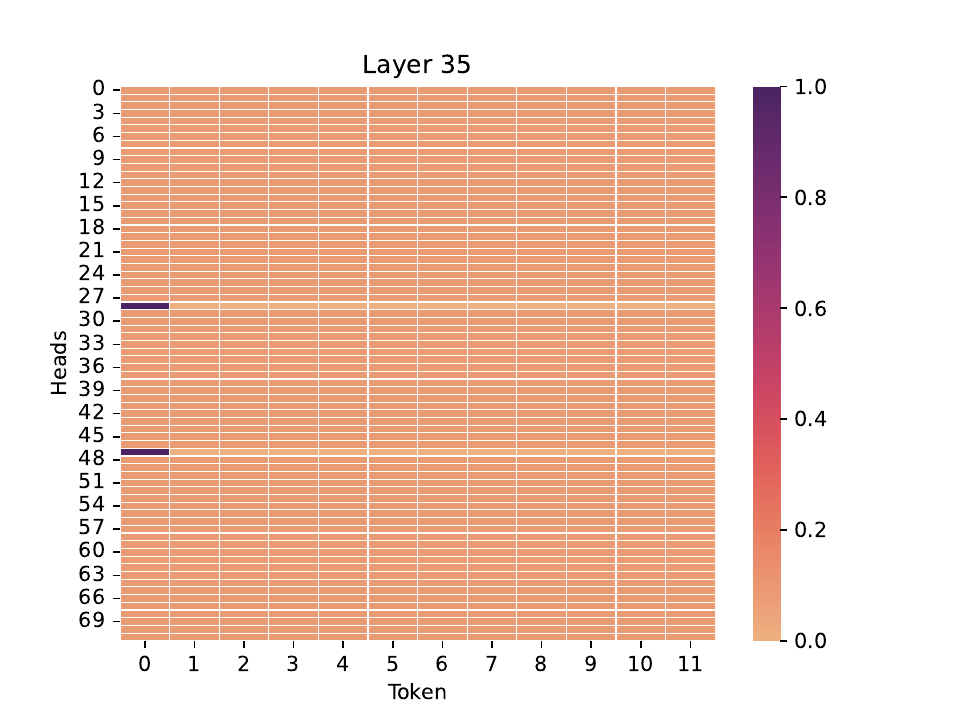}
        \caption{\small{Layer 35}}
    \end{subfigure}
    \begin{subfigure}[t]{0.33\textwidth}
        \includegraphics[width=\textwidth]{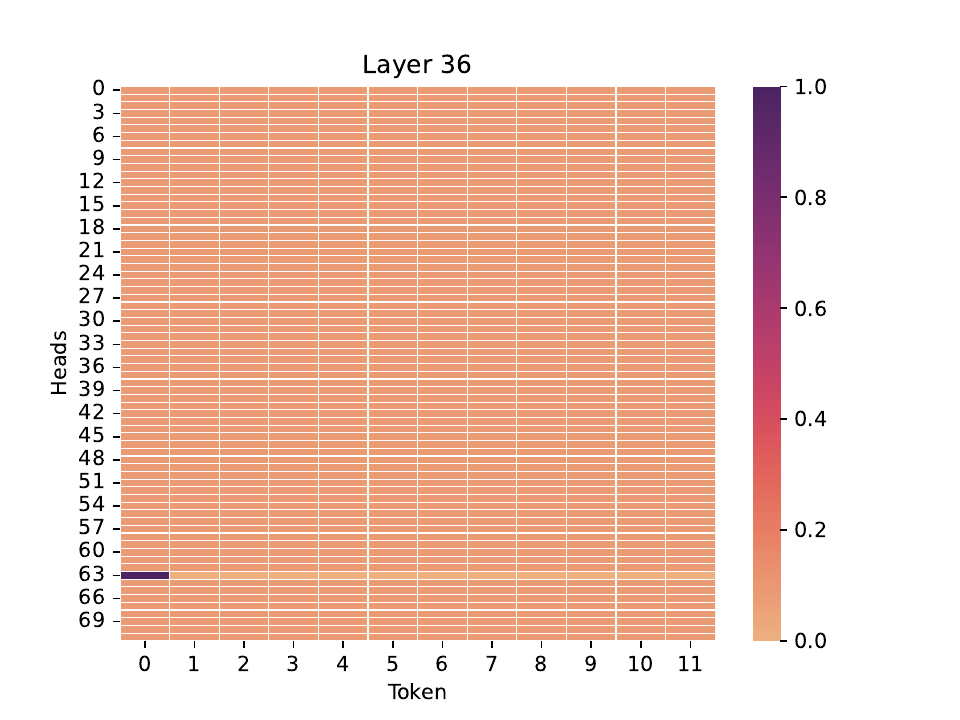}
        \caption{\small{Layer 36}}
    \end{subfigure}
    \begin{subfigure}[t]{0.33\textwidth}
        \includegraphics[width=\textwidth]{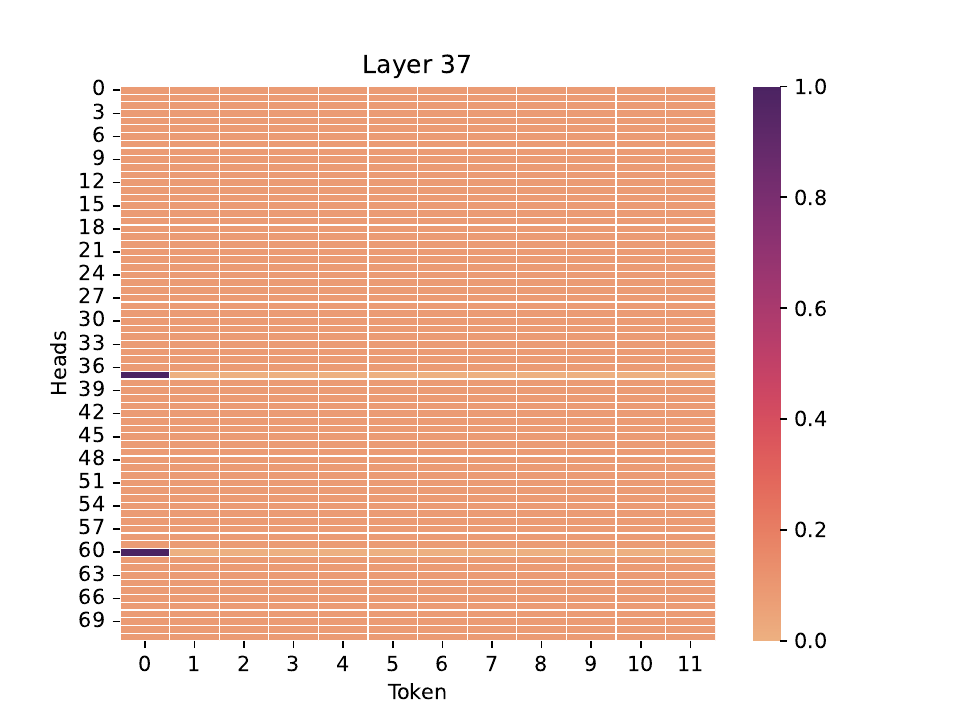}
        \caption{\small{Layer 37}}
    \end{subfigure}
 
 \end{center}
 \caption{Activations of \opt-66B}
    \end{figure}
    \begin{figure}[t]\ContinuedFloat
          \renewcommand\thesubfigure{\roman{subfigure}}
    \begin{center}
       \begin{subfigure}[t]{0.33\textwidth}
        \includegraphics[width=\textwidth]{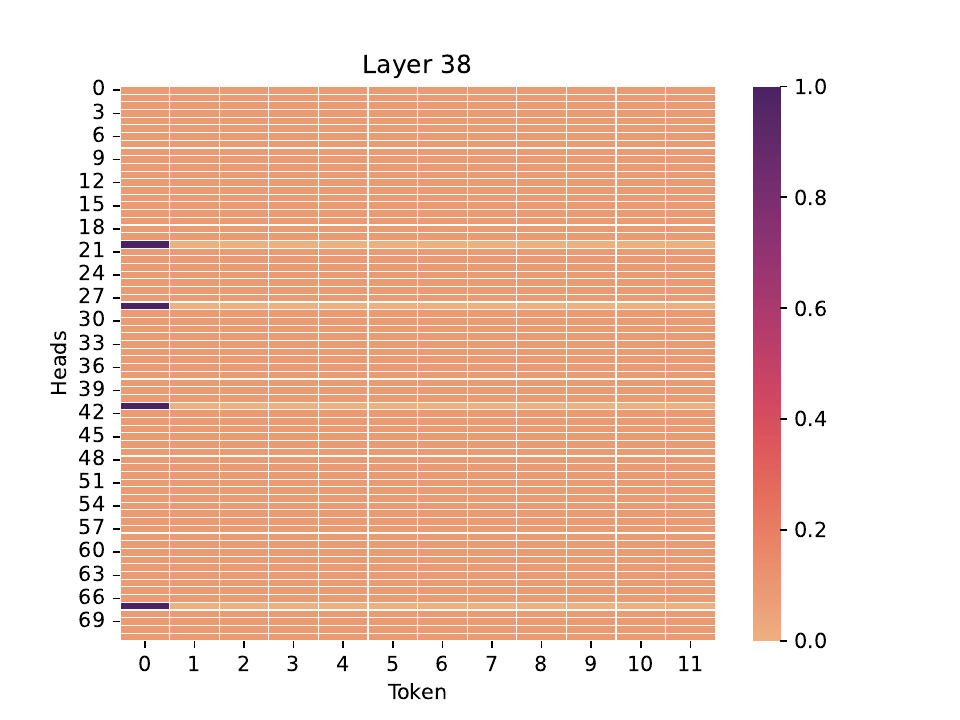}
        \caption{\small{Layer 38}}
    \end{subfigure}
\begin{subfigure}[t]{0.33\textwidth}
        \includegraphics[width=\textwidth]{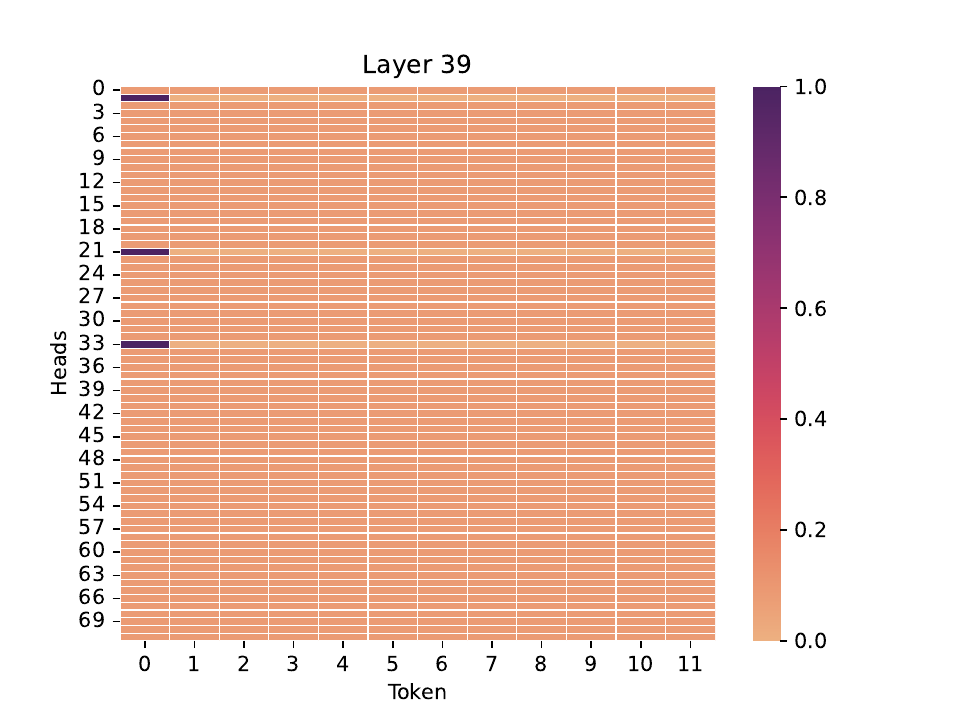}
        \caption{\small{Layer 39}}
    \end{subfigure}
    \begin{subfigure}[t]{0.33\textwidth}
        \includegraphics[width=\textwidth]{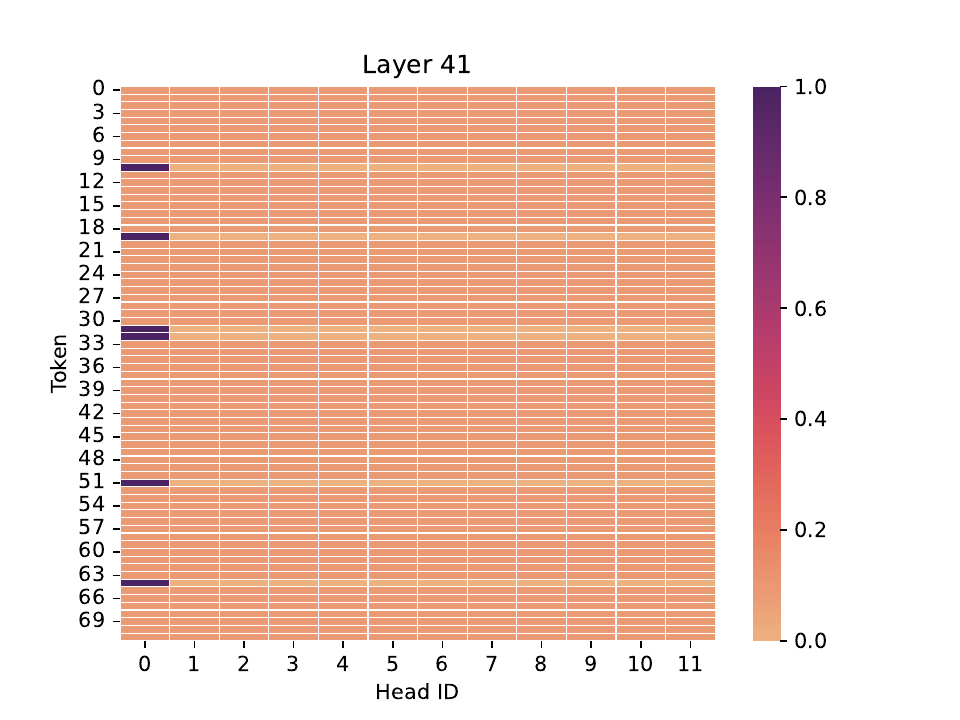}
        \caption{\small{Layer 41}}
    \end{subfigure}
    \begin{subfigure}[t]{0.33\textwidth}
        \includegraphics[width=\textwidth]{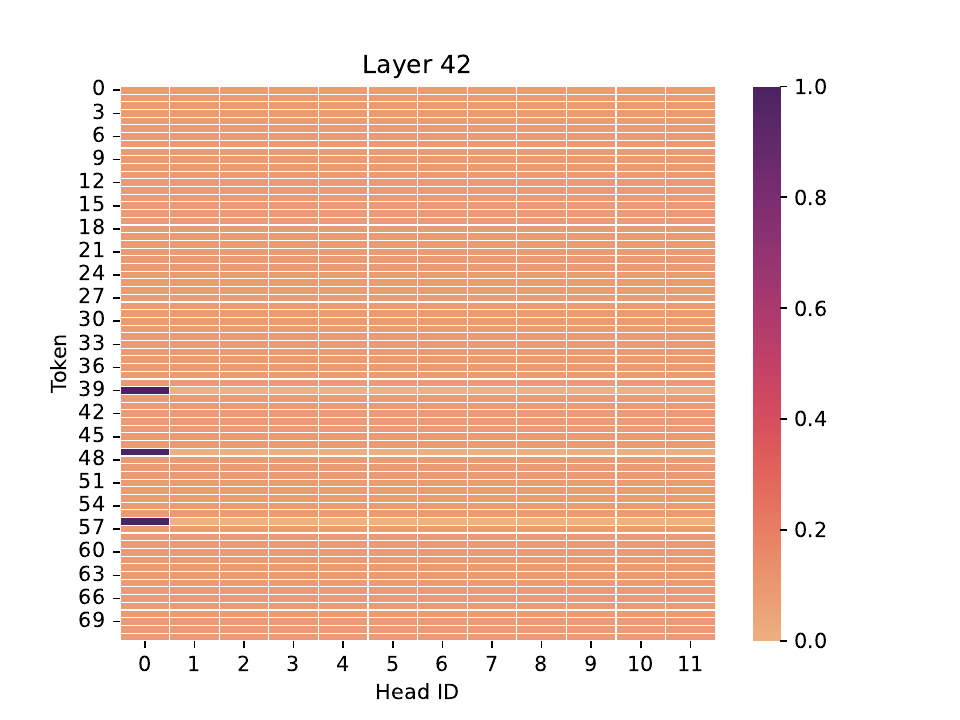}
        \caption{\small{Layer 42}}
    \end{subfigure}
    \begin{subfigure}[t]{0.33\textwidth}
        \includegraphics[width=\textwidth]{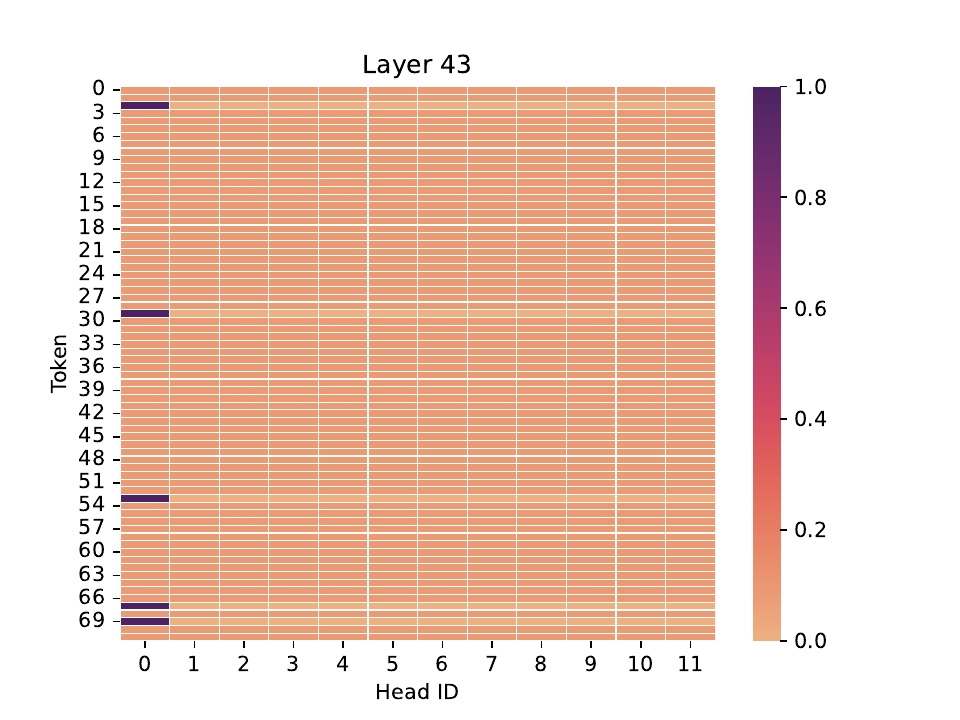}
        \caption{\small{Layer 43}}
    \end{subfigure}
    \begin{subfigure}[t]{0.33\textwidth}
        \includegraphics[width=\textwidth]{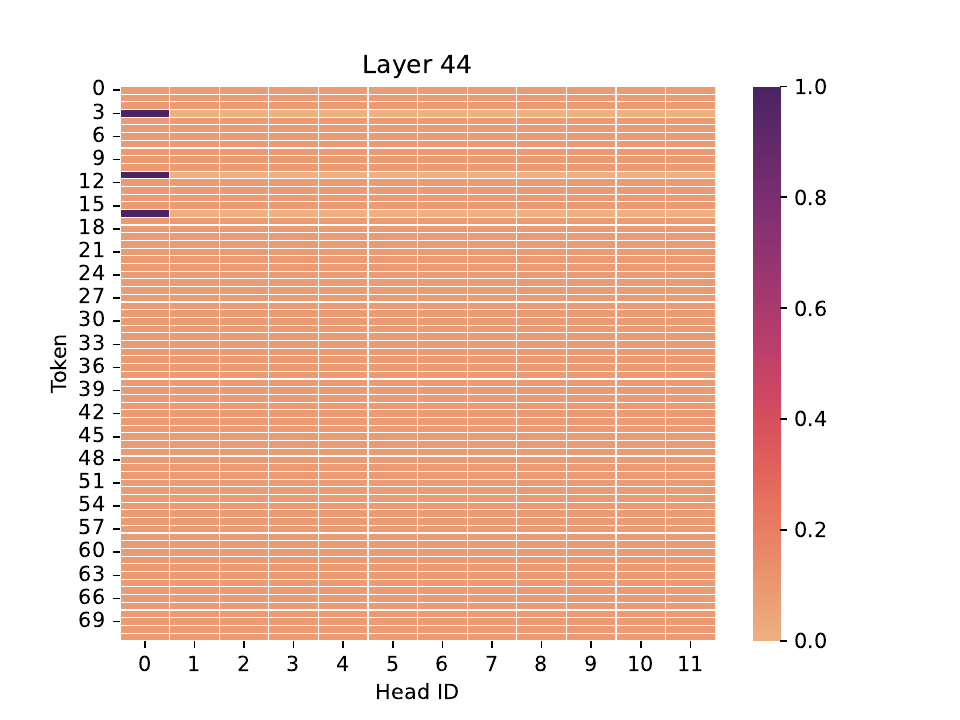}
        \caption{\small{Layer 44}}
    \end{subfigure}
    \begin{subfigure}[t]{0.33\textwidth}
        \includegraphics[width=\textwidth]{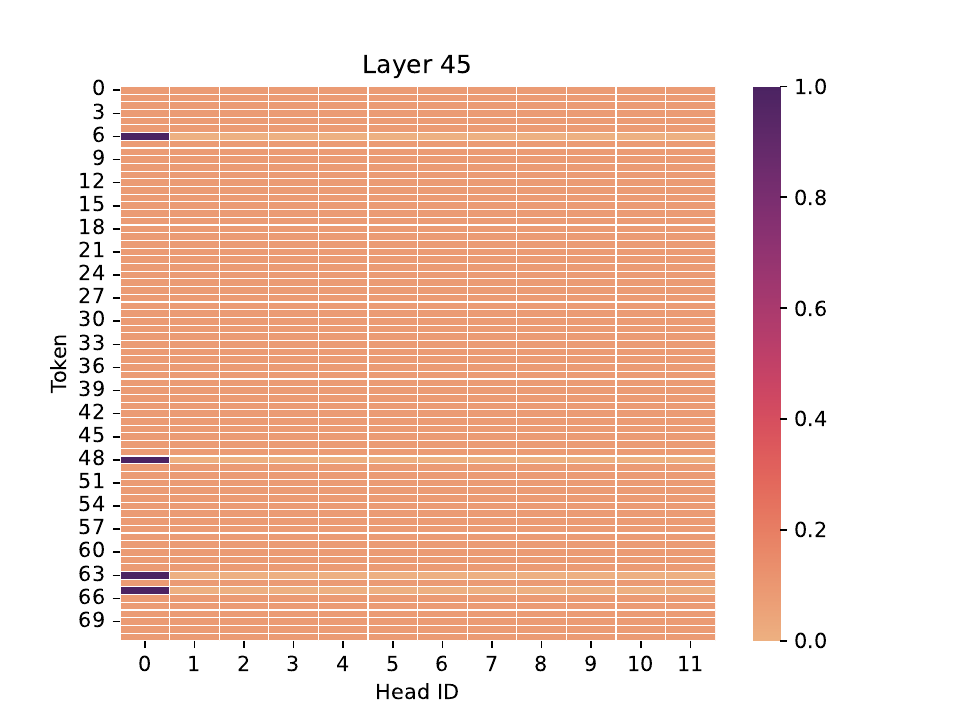}
        \caption{\small{Layer 45}}
    \end{subfigure}
    \begin{subfigure}[t]{0.33\textwidth}
        \includegraphics[width=\textwidth]{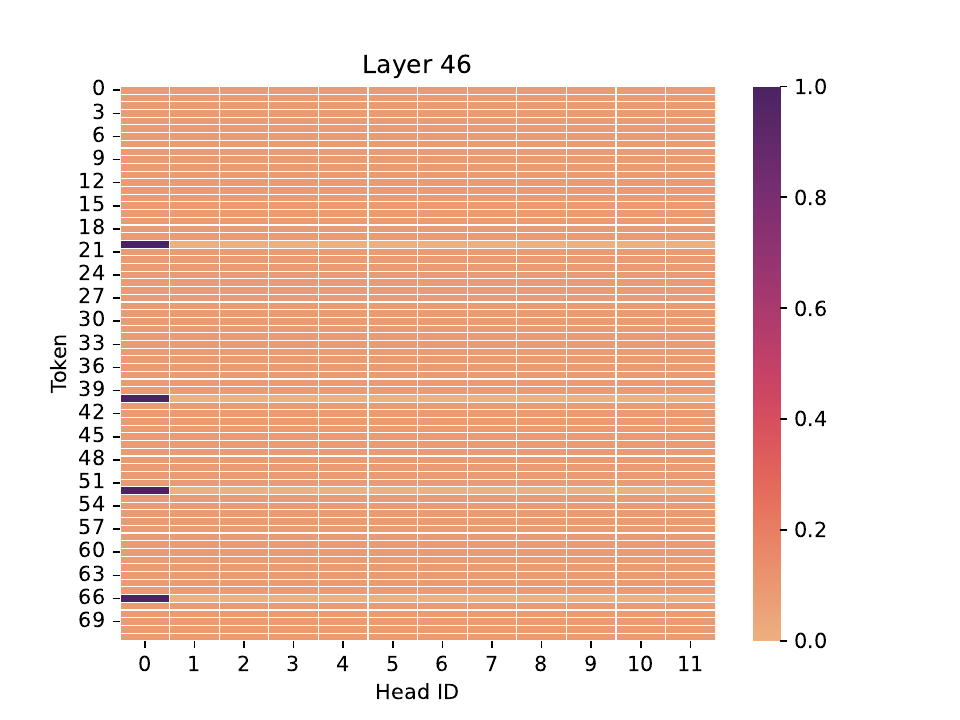}
        \caption{\small{Layer 46}}
    \end{subfigure}
     \begin{subfigure}[t]{0.33\textwidth}
        \includegraphics[width=\textwidth]{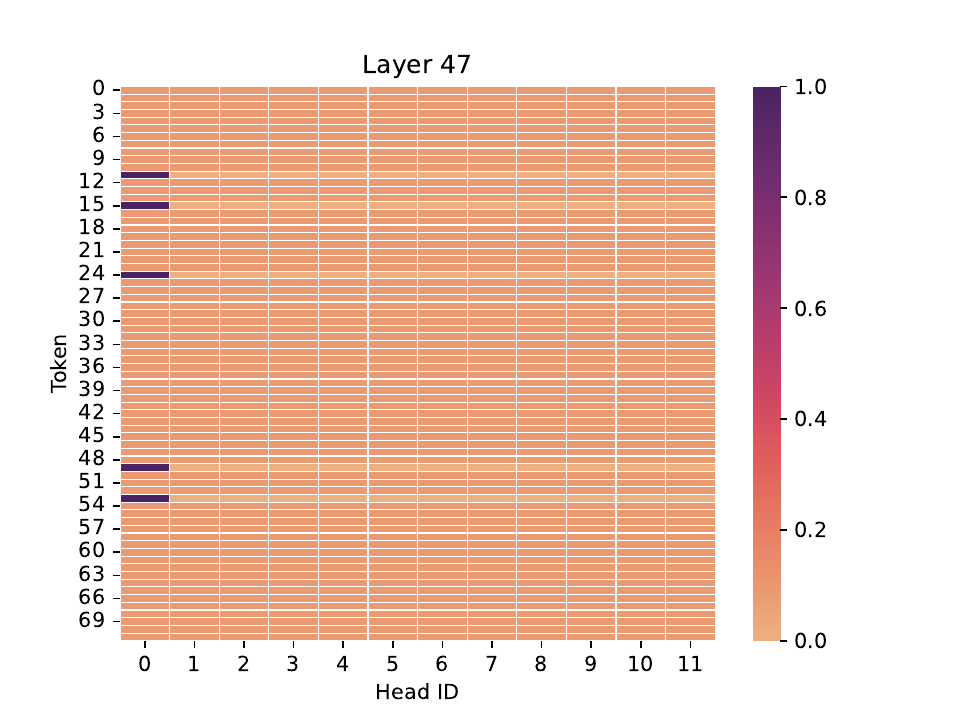}
        \caption{\small{Layer 47}}
    \end{subfigure}
     \begin{subfigure}[t]{0.33\textwidth}
        \includegraphics[width=\textwidth]{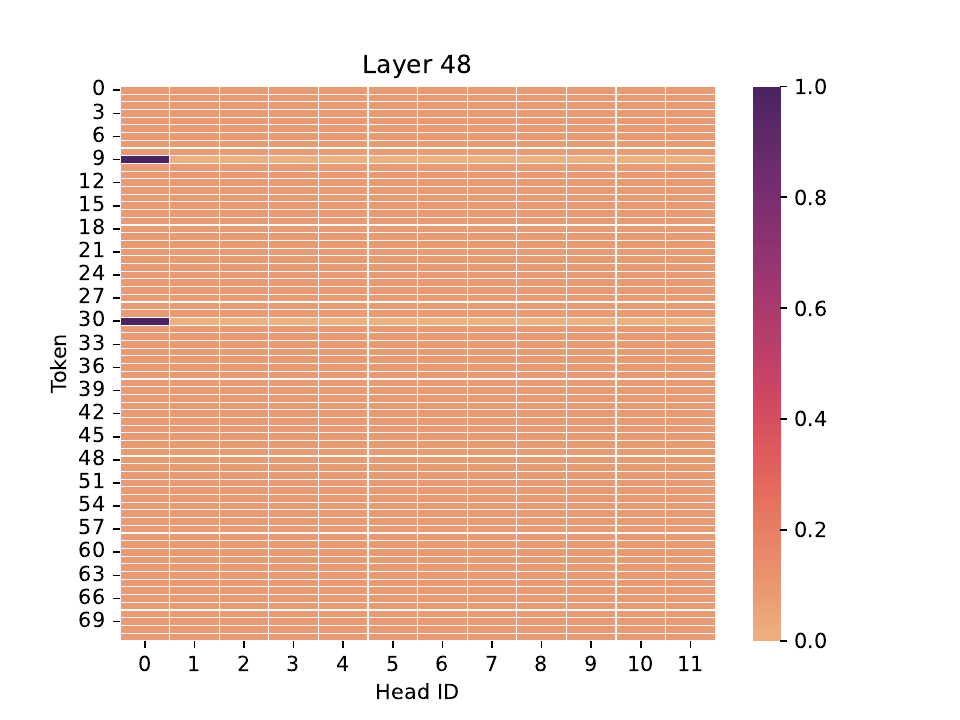}
        \caption{\small{Layer 48}}
    \end{subfigure}
     \begin{subfigure}[t]{0.33\textwidth}
        \includegraphics[width=\textwidth]{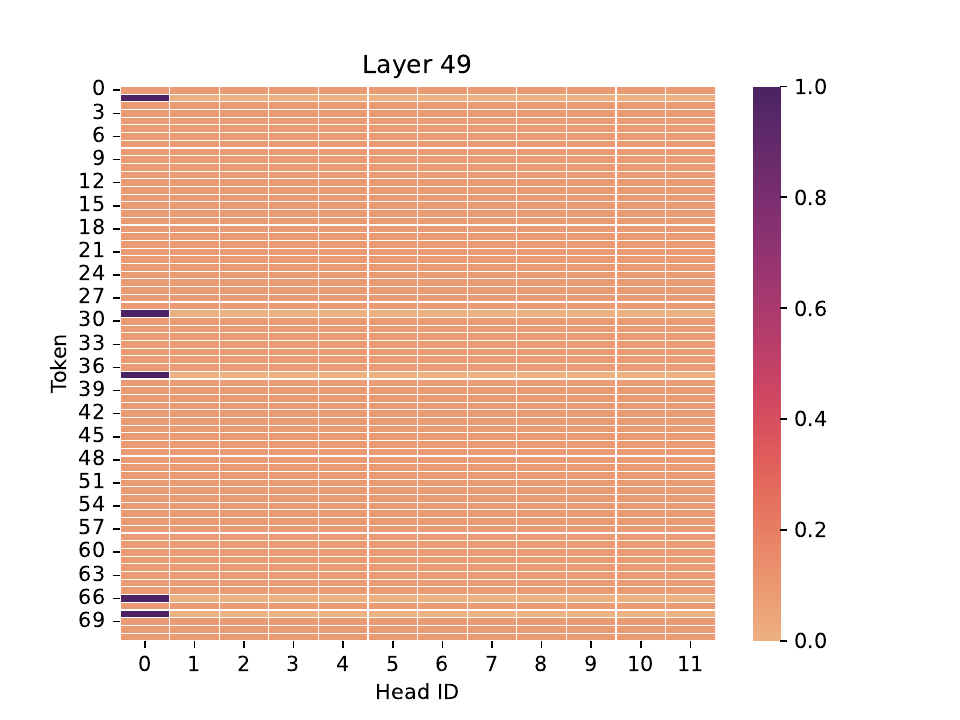}
        \caption{\small{Layer 49}}
    \end{subfigure}
         \begin{subfigure}[t]{0.33\textwidth}
        \includegraphics[width=\textwidth]{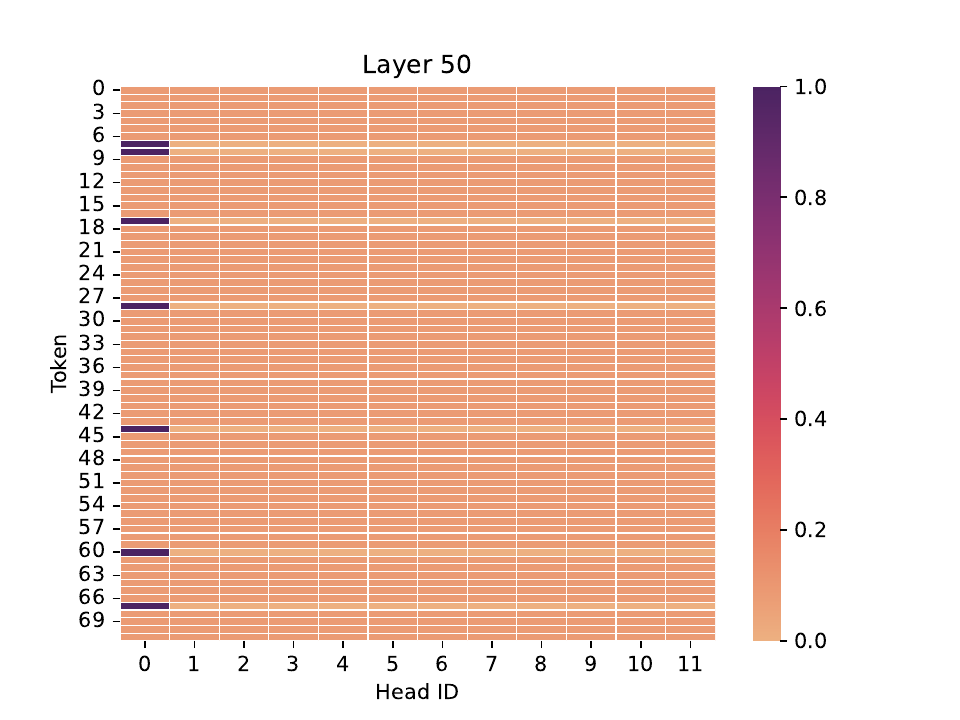}
        \caption{\small{Layer 50}}
        \end{subfigure}
 \begin{subfigure}[t]{0.33\textwidth}
        \includegraphics[width=\textwidth]{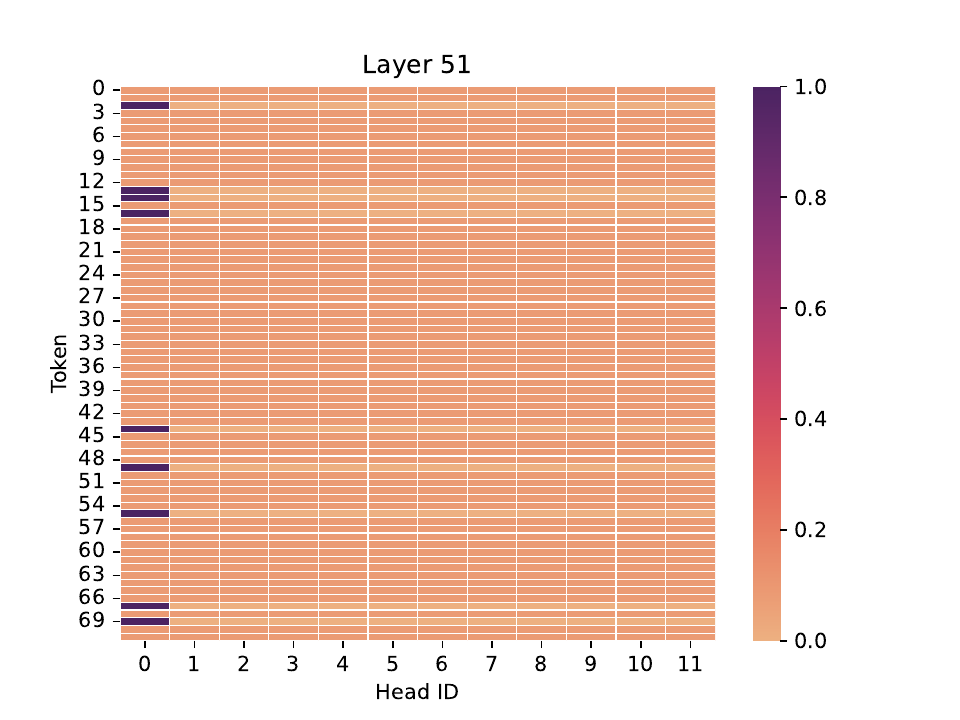}
        \caption{\small{Layer 51}}
    \end{subfigure}
\begin{subfigure}[t]{0.33\textwidth}
        \includegraphics[width=\textwidth]{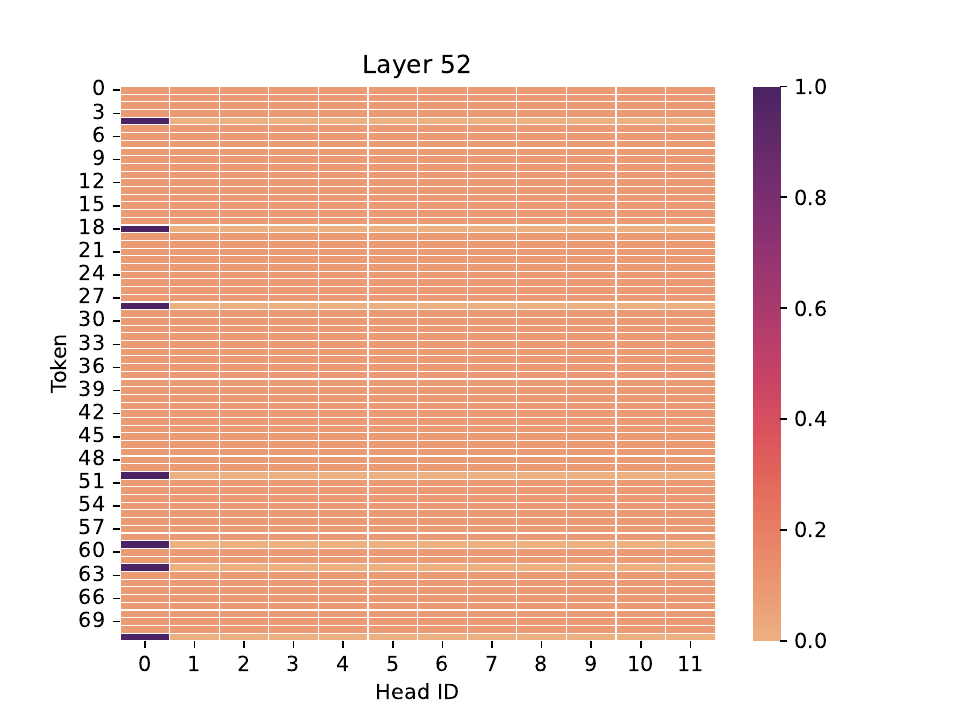}
        \caption{\small{Layer 52}}
    \end{subfigure}
    \begin{subfigure}[t]{0.33\textwidth}
        \includegraphics[width=\textwidth]{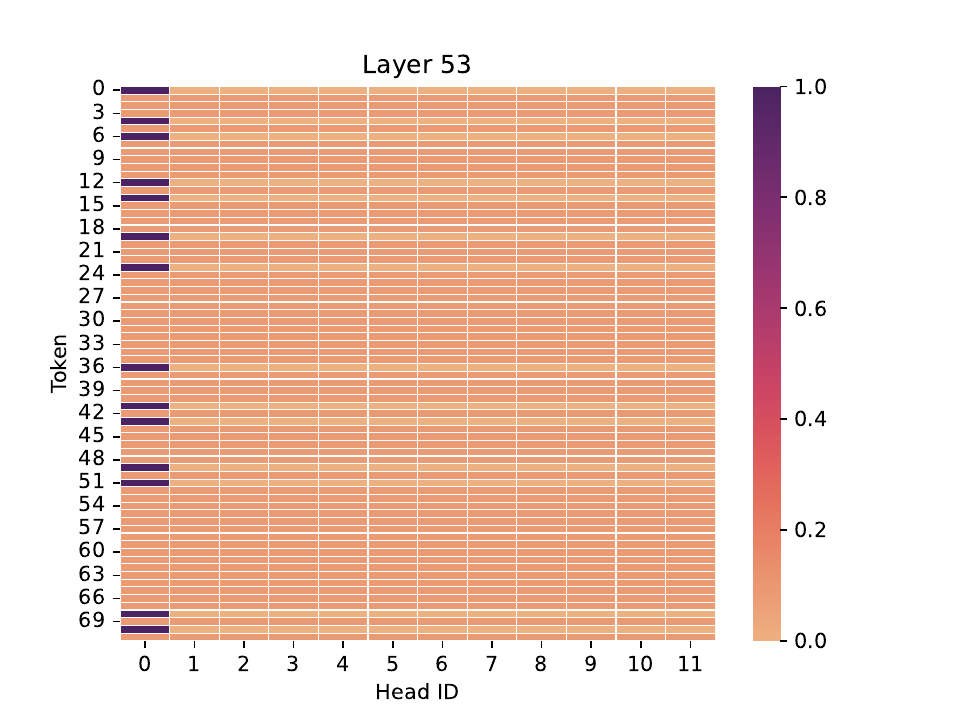}
        \caption{\small{Layer 53}}
    \end{subfigure}
    \end{center}
    \caption{Activations of \opt-66B}
    \end{figure}
    \begin{figure}[t]\ContinuedFloat
          \renewcommand\thesubfigure{\roman{subfigure}}
    \begin{center}
    \begin{subfigure}[t]{0.33\textwidth}
        \includegraphics[width=\textwidth]{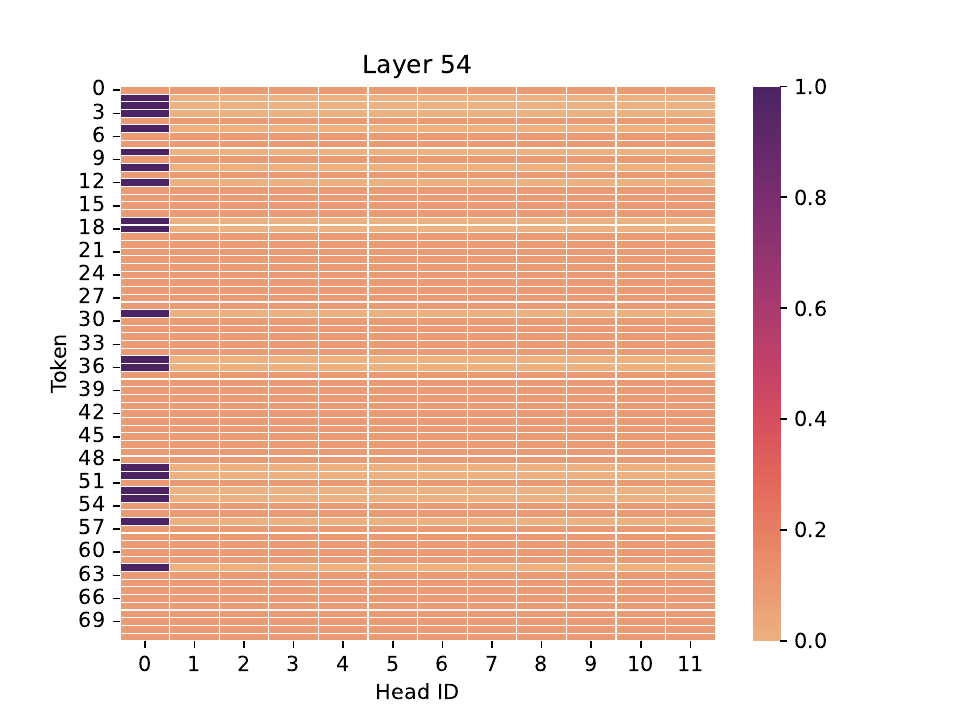}
        \caption{\small{Layer 54}}
    \end{subfigure}
    \begin{subfigure}[t]{0.33\textwidth}
        \includegraphics[width=\textwidth]{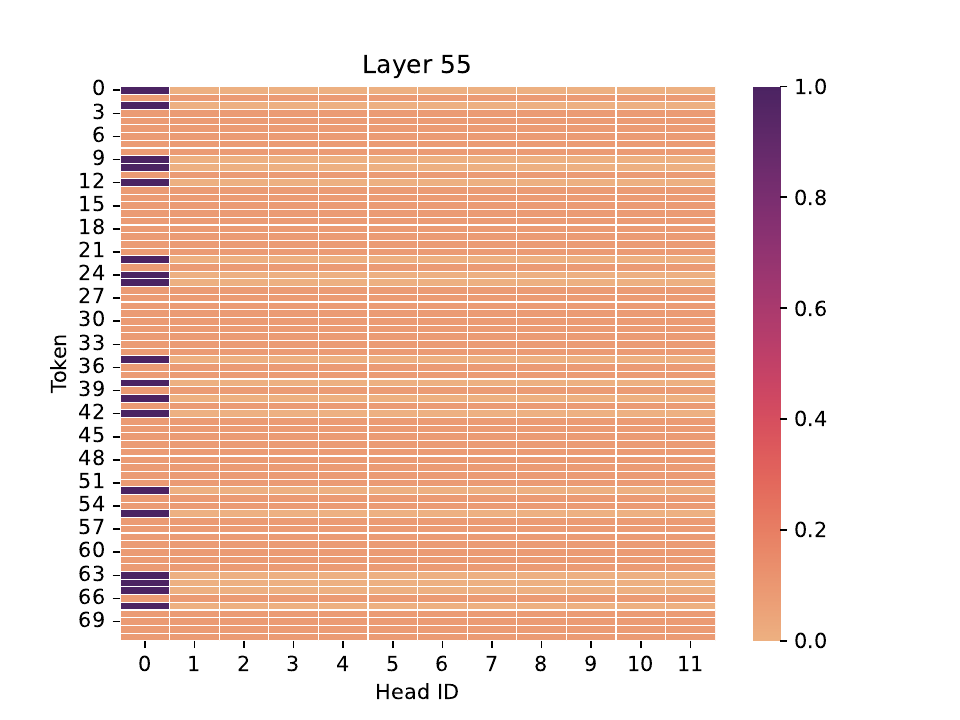}
        \caption{\small{Layer 55}}
    \end{subfigure}
    \begin{subfigure}[t]{0.33\textwidth}
        \includegraphics[width=\textwidth]{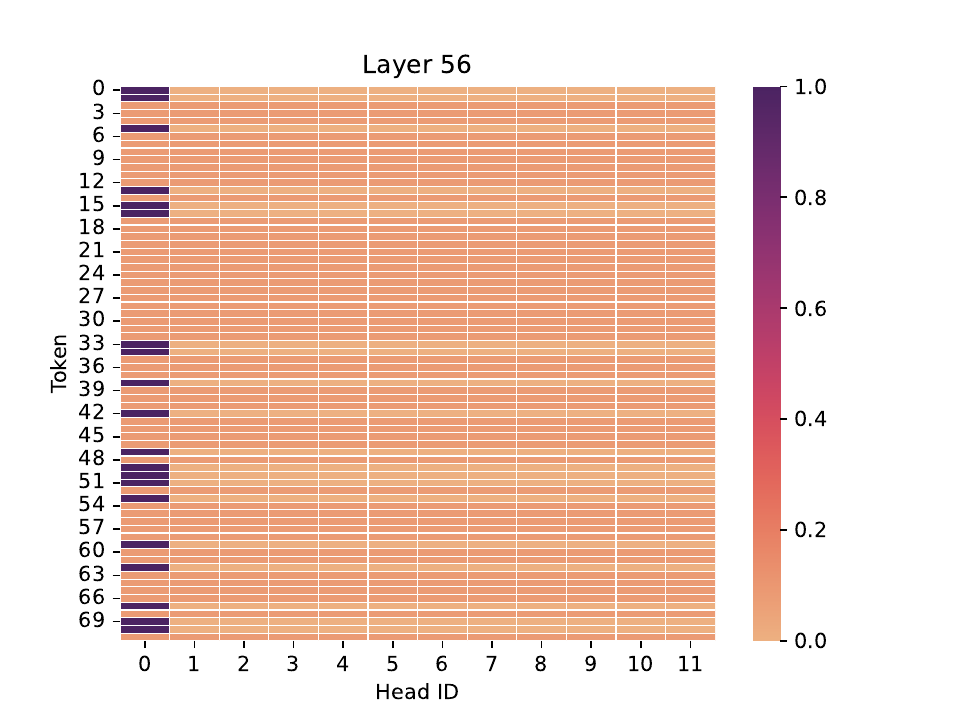}
        \caption{\small{Layer 56}}
    \end{subfigure}
    \begin{subfigure}[t]{0.33\textwidth}
        \includegraphics[width=\textwidth]{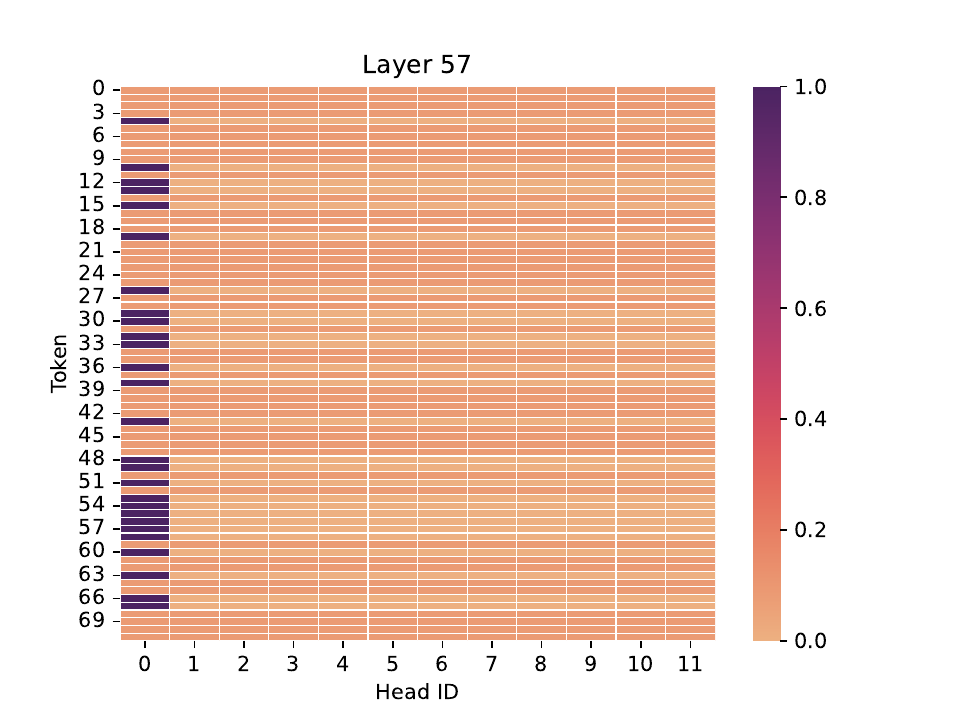}
        \caption{\small{Layer 57}}
    \end{subfigure}
\begin{subfigure}[t]{0.33\textwidth}
        \includegraphics[width=\textwidth]{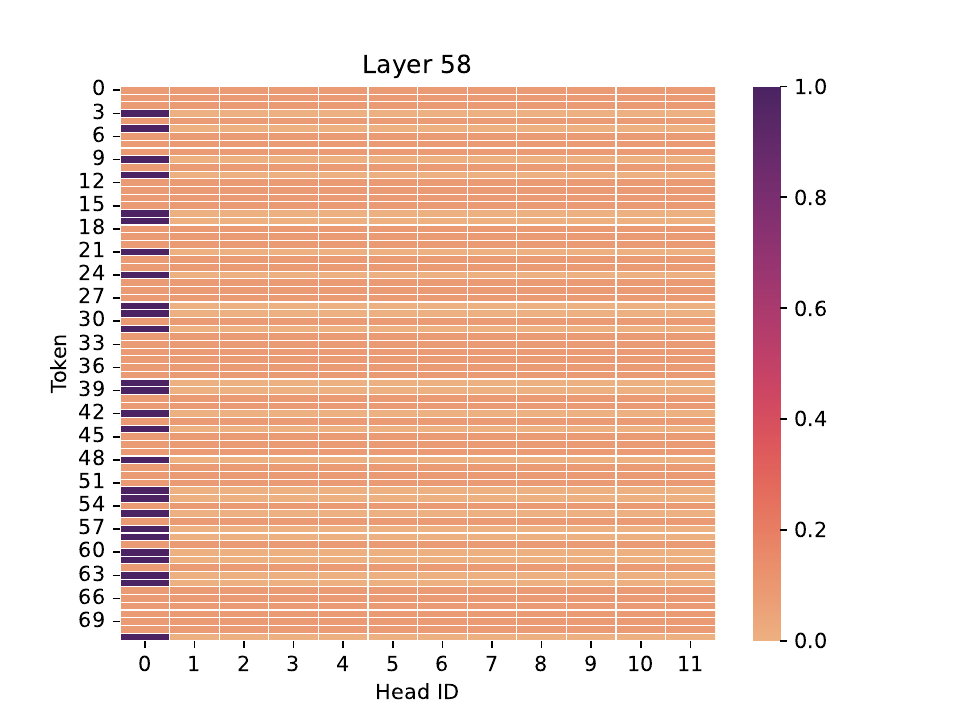}
        \caption{\small{Layer 58}}
    \end{subfigure}
    \begin{subfigure}[t]{0.33\textwidth}
        \includegraphics[width=\textwidth]{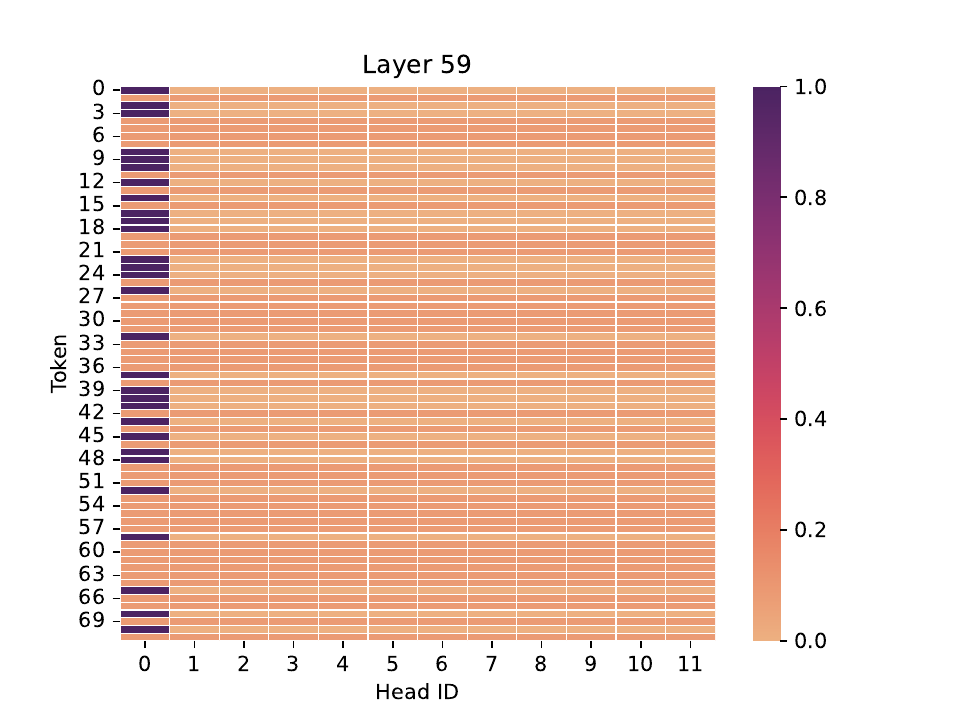}
        \caption{\small{Layer 59}}
    \end{subfigure}
    \begin{subfigure}[t]{0.33\textwidth}
        \includegraphics[width=\textwidth]{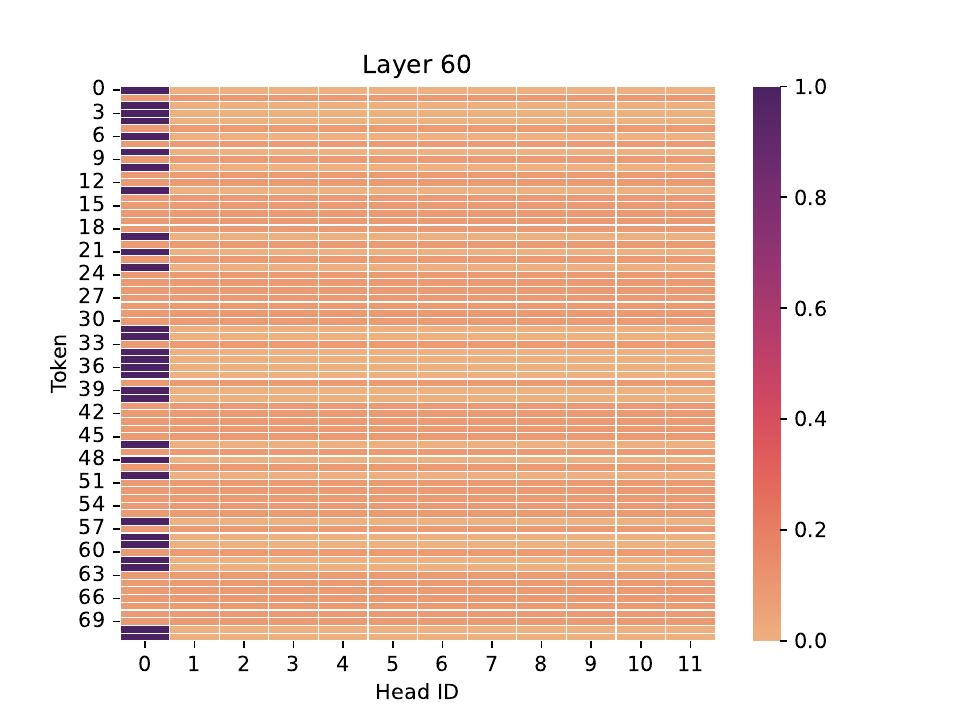}
        \caption{\small{Layer 60}}
    \end{subfigure}
        \begin{subfigure}[t]{0.33\textwidth}
        \includegraphics[width=\textwidth]{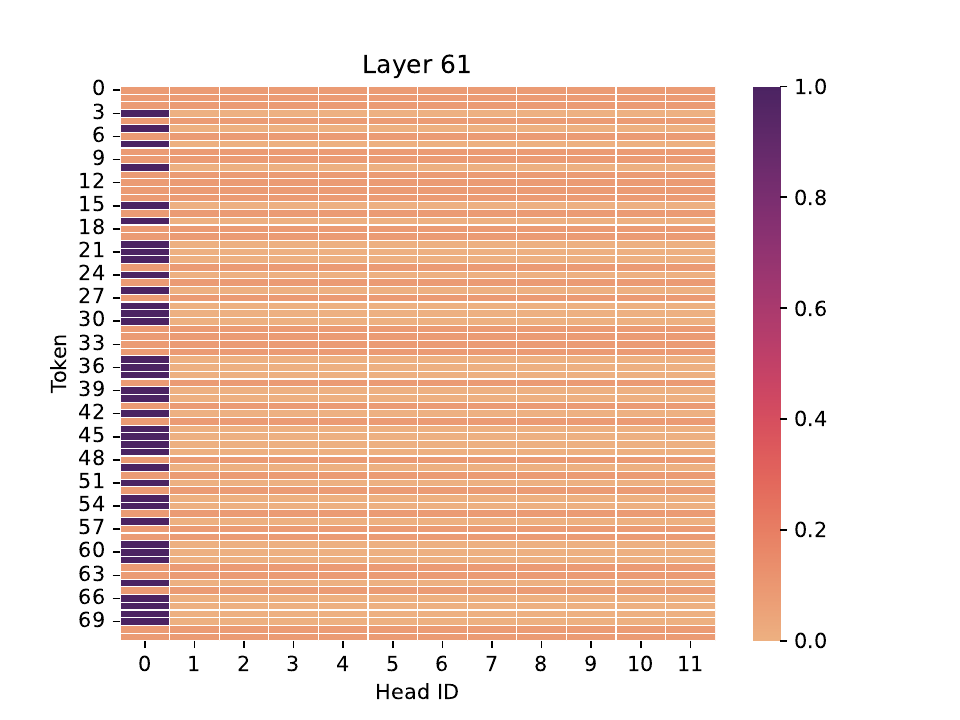}
        \caption{\small{Layer 61}}
    \end{subfigure}
        \begin{subfigure}[t]{0.33\textwidth}
        \includegraphics[width=\textwidth]{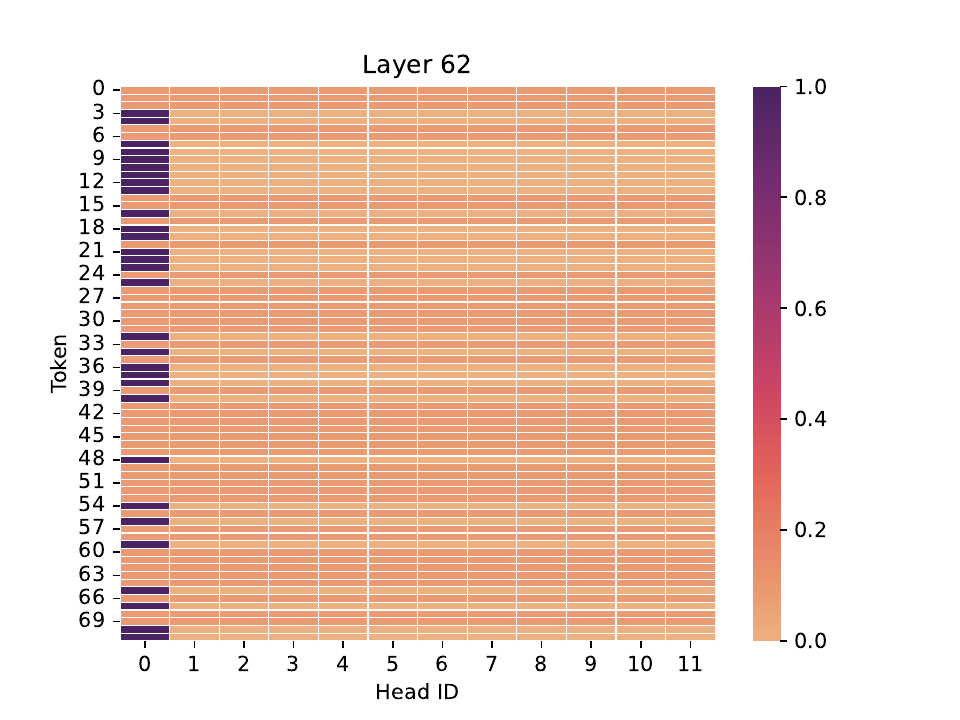}
        \caption{\small{Layer 62}}
    \end{subfigure}
        \begin{subfigure}[t]{0.33\textwidth}
        \includegraphics[width=\textwidth]{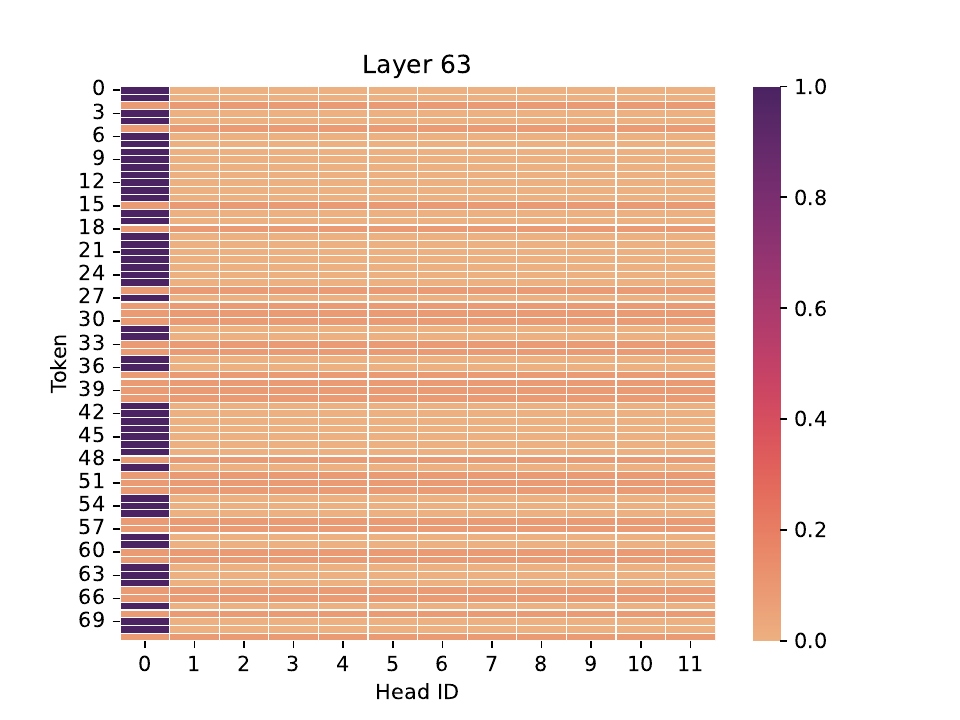}
        \caption{\small{Layer 63}}
    \end{subfigure}
\caption{Layer Map \opt-66B}
\label{fig:optlast}
\end{center}
\end{figure}

\begin{figure}[t]
\begin{center}
    \begin{subfigure}[t]{0.33\textwidth}
    \includegraphics[width=\textwidth]{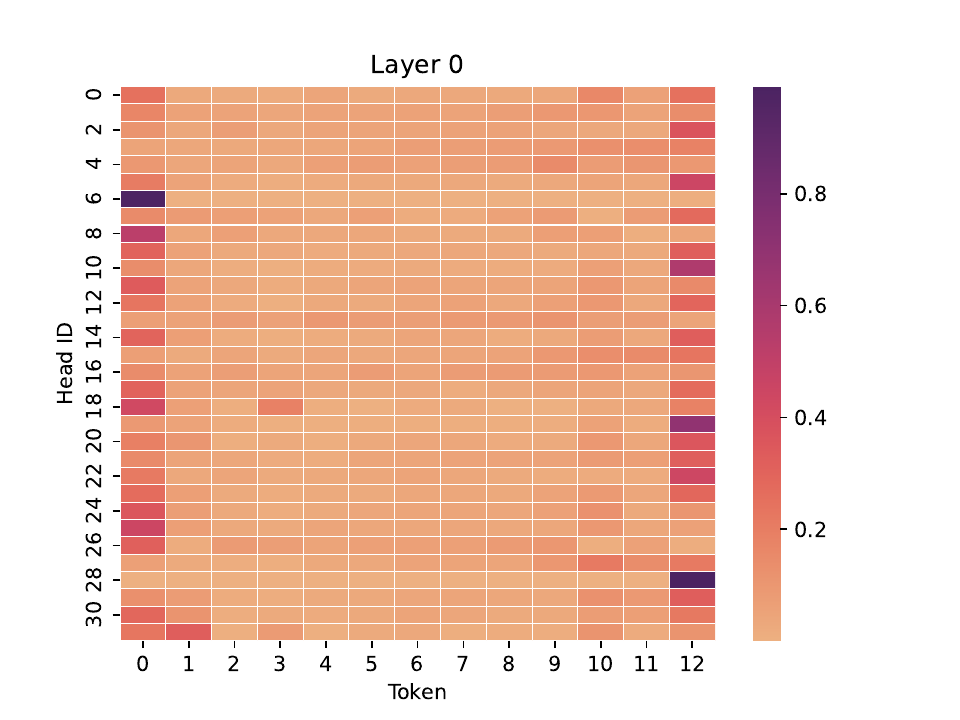}
    \caption{\small{Layer 0}}
    \end{subfigure}\
    \begin{subfigure}[t]{0.33\textwidth}
        \includegraphics[width=\textwidth]{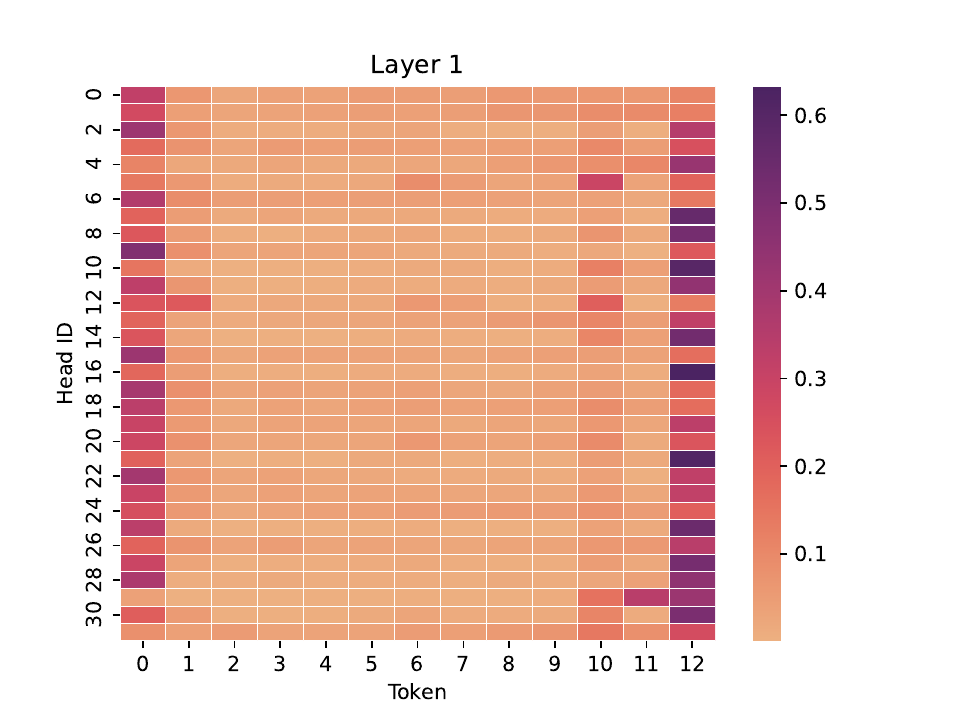}
        \caption{\small{Layer 1}}
    \end{subfigure}
    \begin{subfigure}[t]{0.33\textwidth}
        \includegraphics[width=\textwidth]{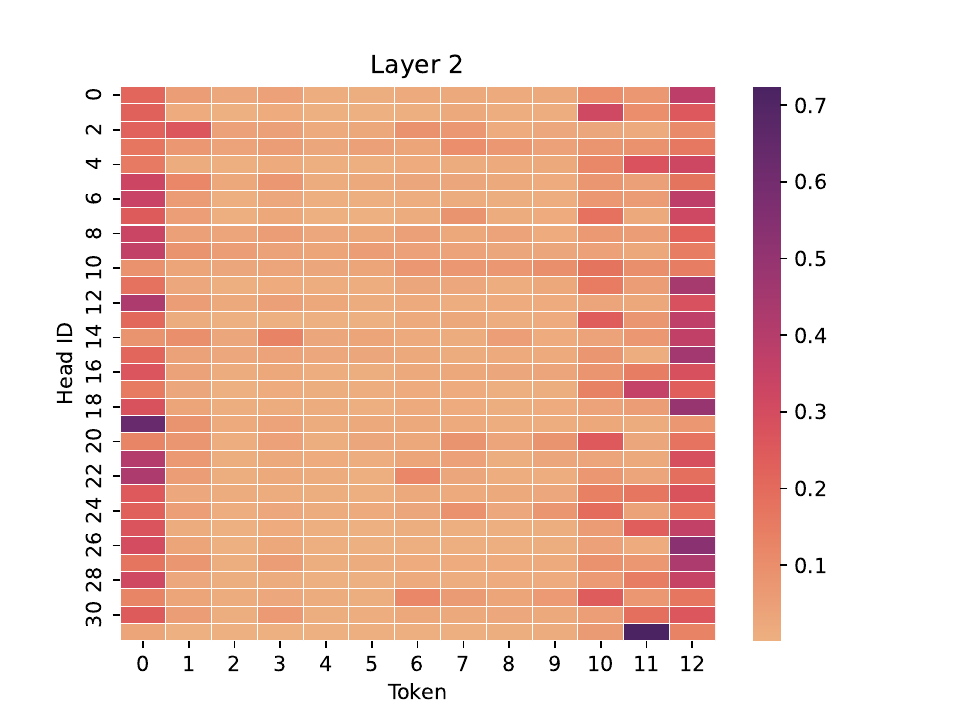}
        \caption{\small{Layer 2}}
    \end{subfigure}
    \begin{subfigure}[t]{0.33\textwidth}
        \includegraphics[width=\textwidth]{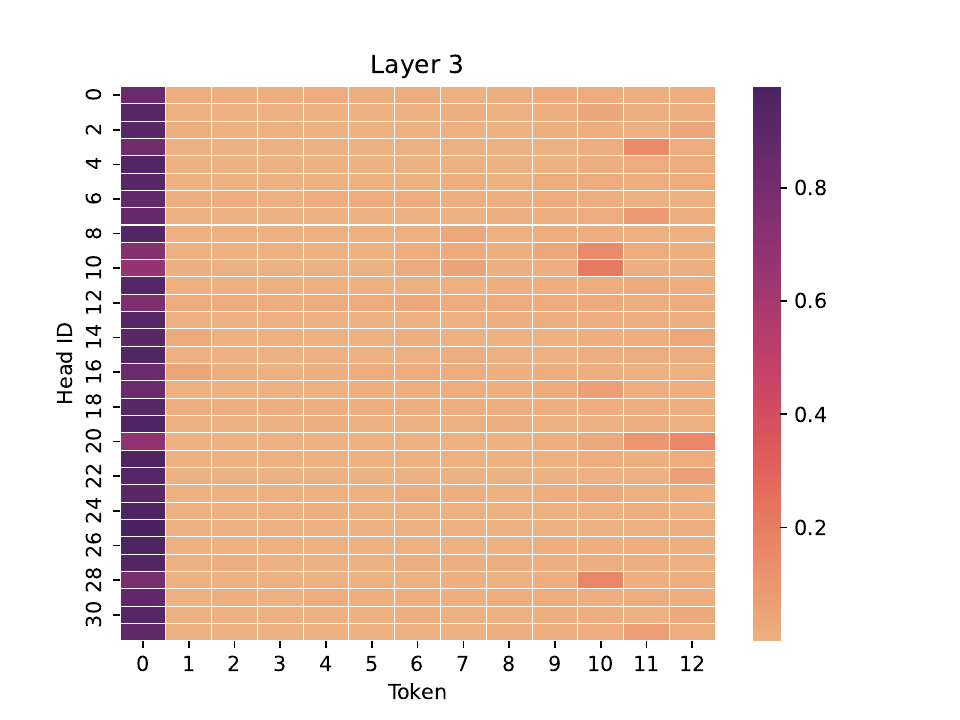}
        \caption{\small{Layer 3}}
    \end{subfigure}
      \begin{subfigure}[t]{0.33\textwidth}
        \includegraphics[width=\textwidth]{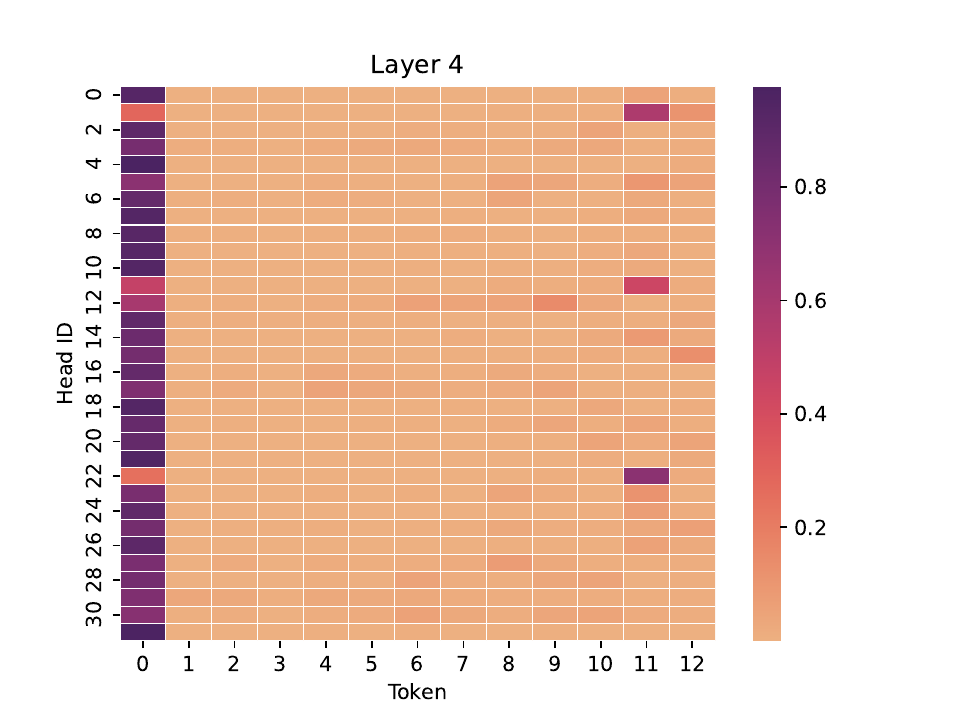}
        \caption{\small{Layer 4}}
    \end{subfigure}
    \begin{subfigure}[t]{0.33\textwidth}
        \includegraphics[width=\textwidth]{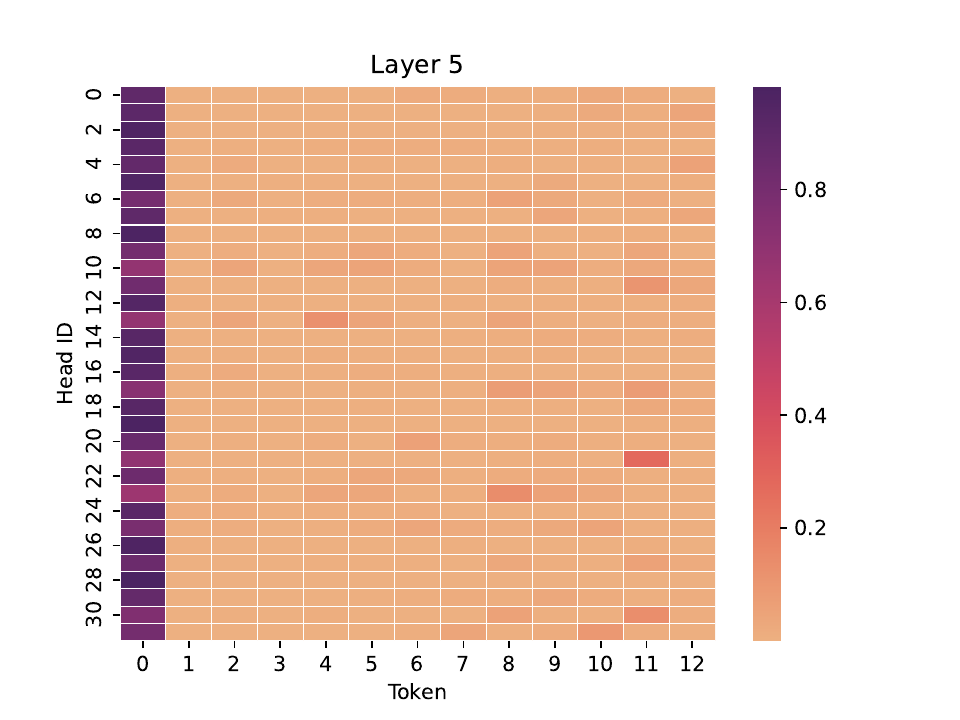}
        \caption{\small{Layer 5}}
    \end{subfigure}
    \begin{subfigure}[t]{0.33\textwidth}
        \includegraphics[width=\textwidth]{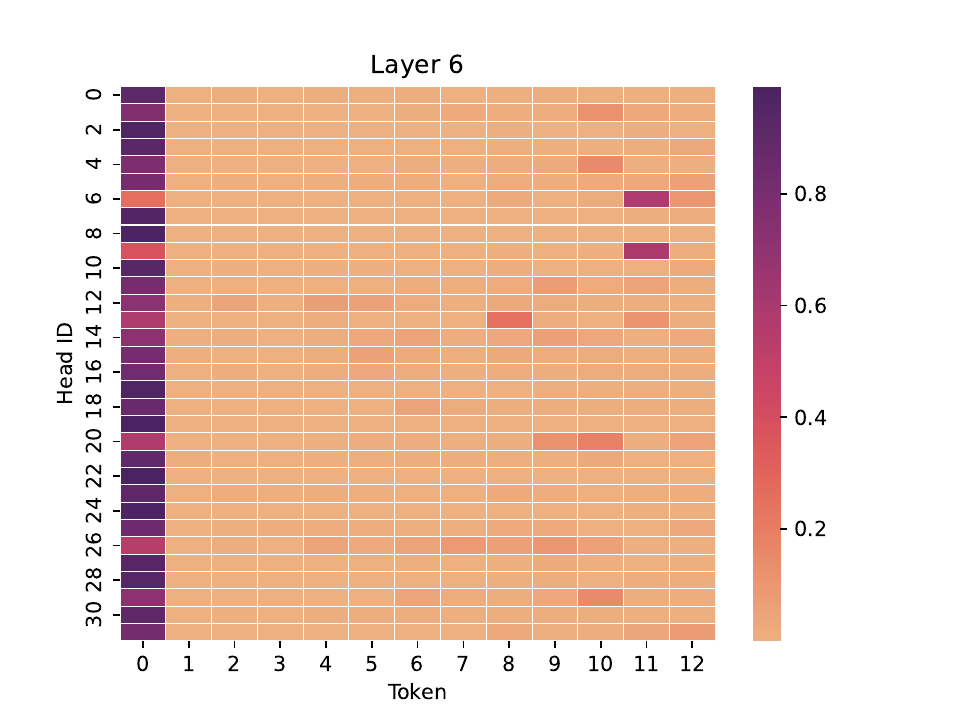}
        \caption{\small{Layer 6}}
    \end{subfigure}
\begin{subfigure}[t]{0.33\textwidth}
        \includegraphics[width=\textwidth]{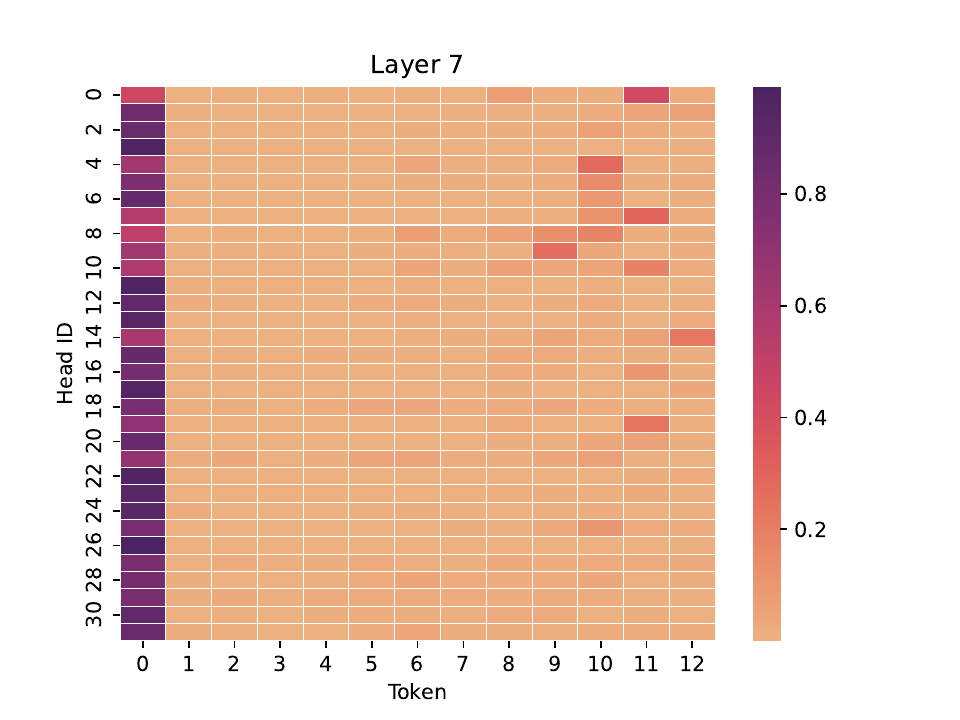}
        \caption{\small{Layer 7}}
    \end{subfigure}
    \begin{subfigure}[t]{0.33\textwidth}
        \includegraphics[width=\textwidth]{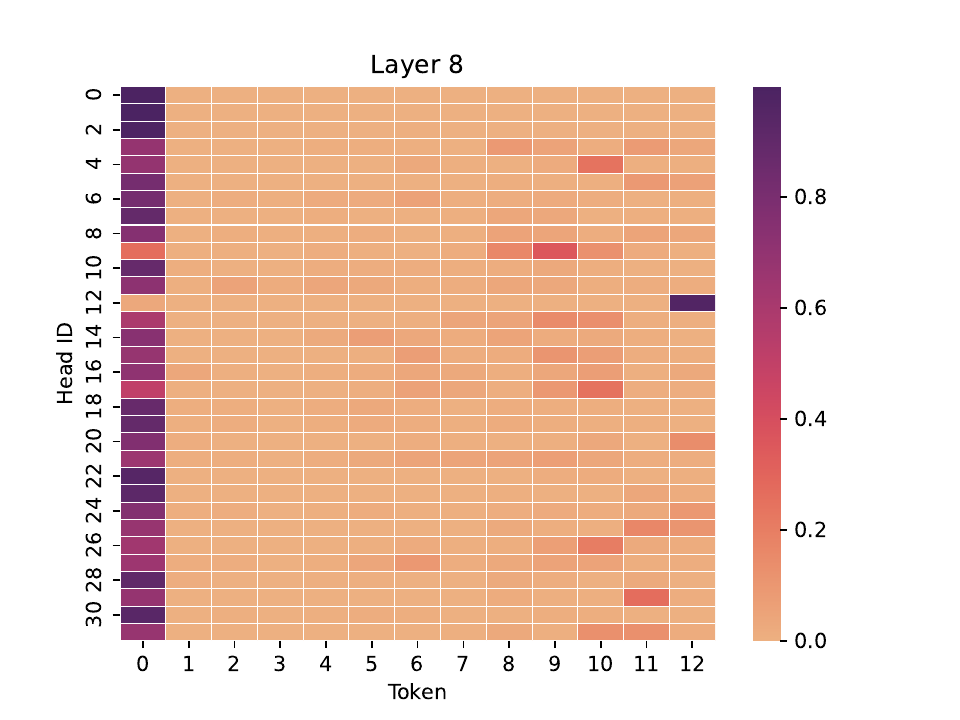}
        \caption{\small{Layer 8}}
    \end{subfigure}
    \begin{subfigure}[t]{0.33\textwidth}
        \includegraphics[width=\textwidth]{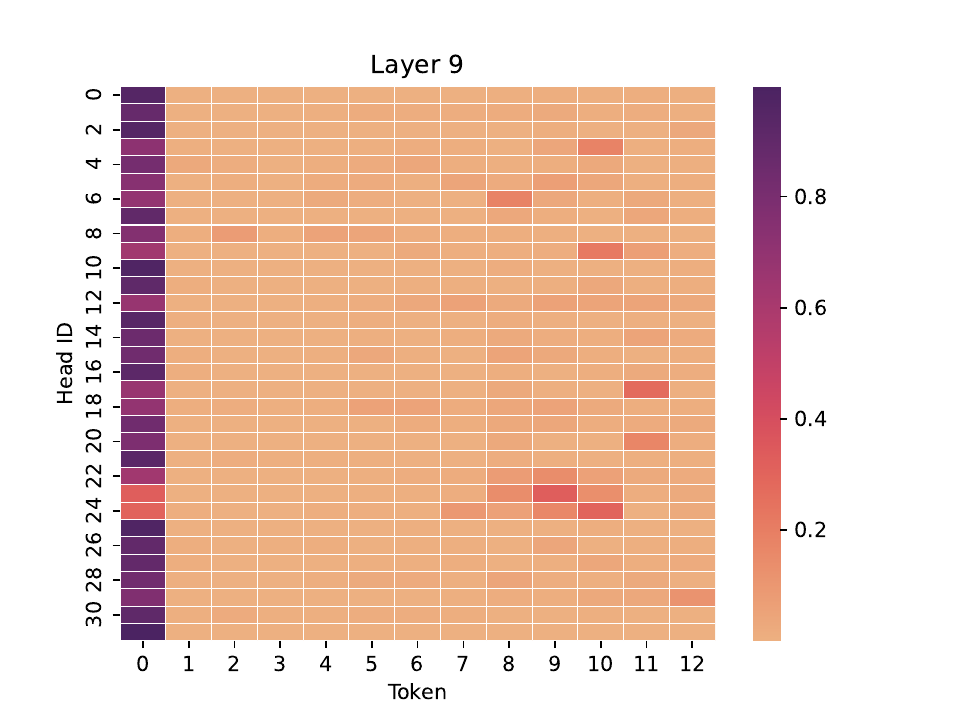}
        \caption{\small{Layer 9}}
    \end{subfigure}
    \begin{subfigure}[t]{0.33\textwidth}
        \includegraphics[width=\textwidth]{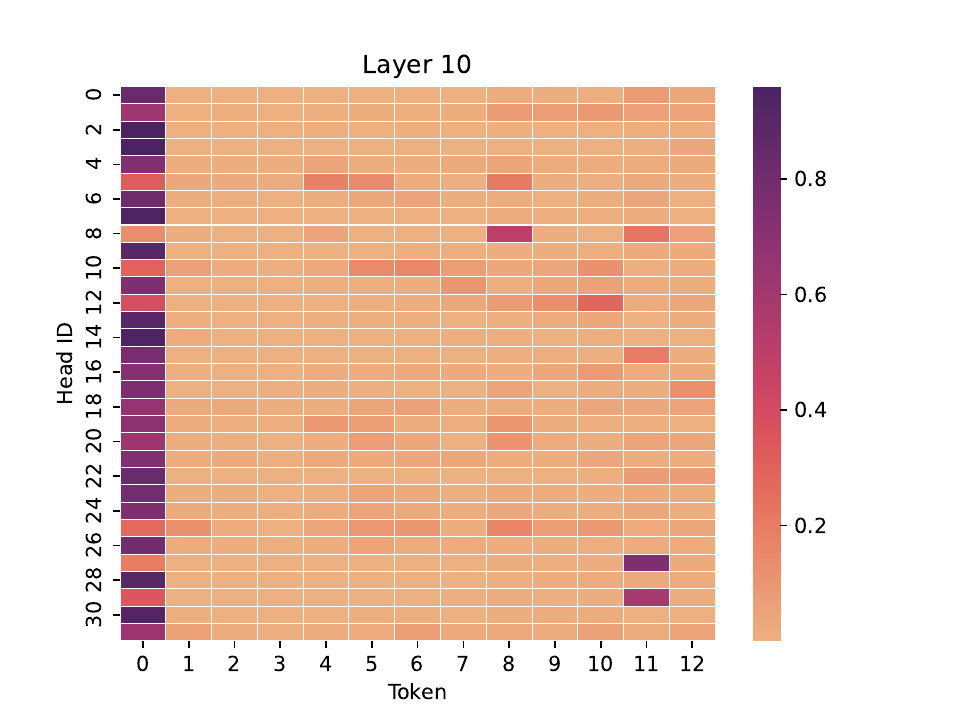}
        \caption{\small{Layer 10}}
    \end{subfigure}
    \begin{subfigure}[t]{0.33\textwidth}
        \includegraphics[width=\textwidth]{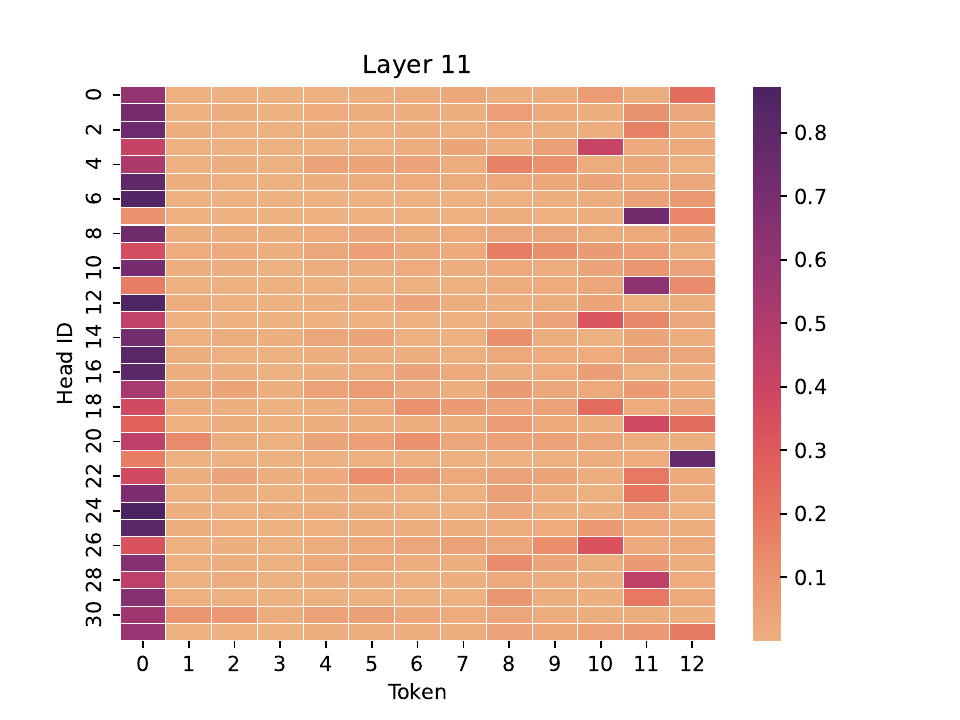}
        \caption{\small{Layer 11}}
    \end{subfigure}
    \begin{subfigure}[t]{0.33\textwidth}
        \includegraphics[width=\textwidth]{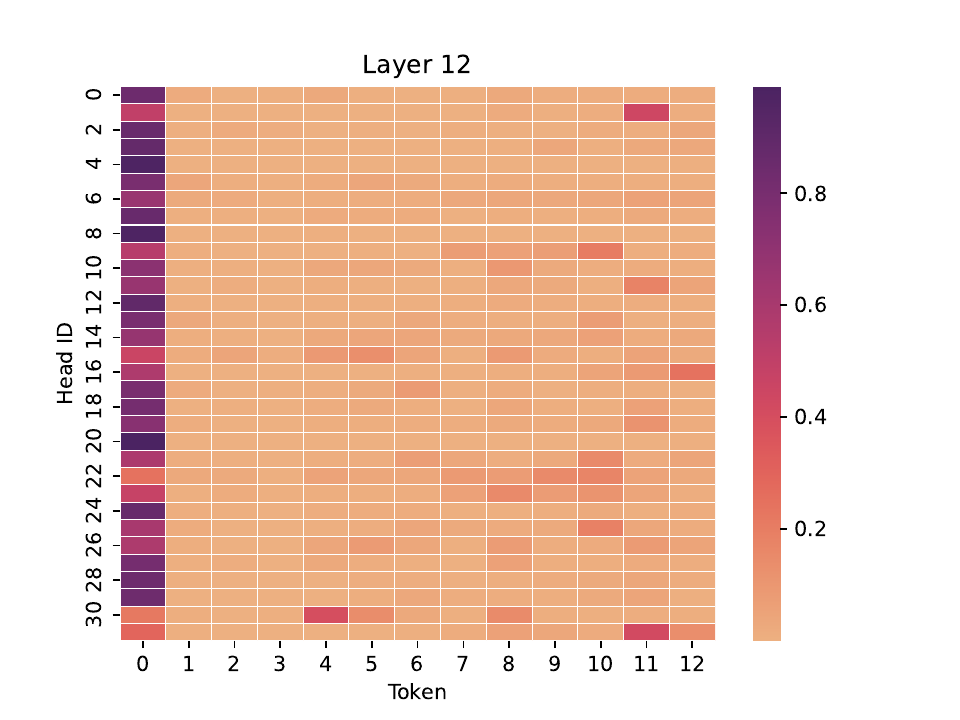}
        \caption{\small{Layer 12}}
    \end{subfigure}
    \begin{subfigure}[t]{0.33\textwidth}
        \includegraphics[width=\textwidth]{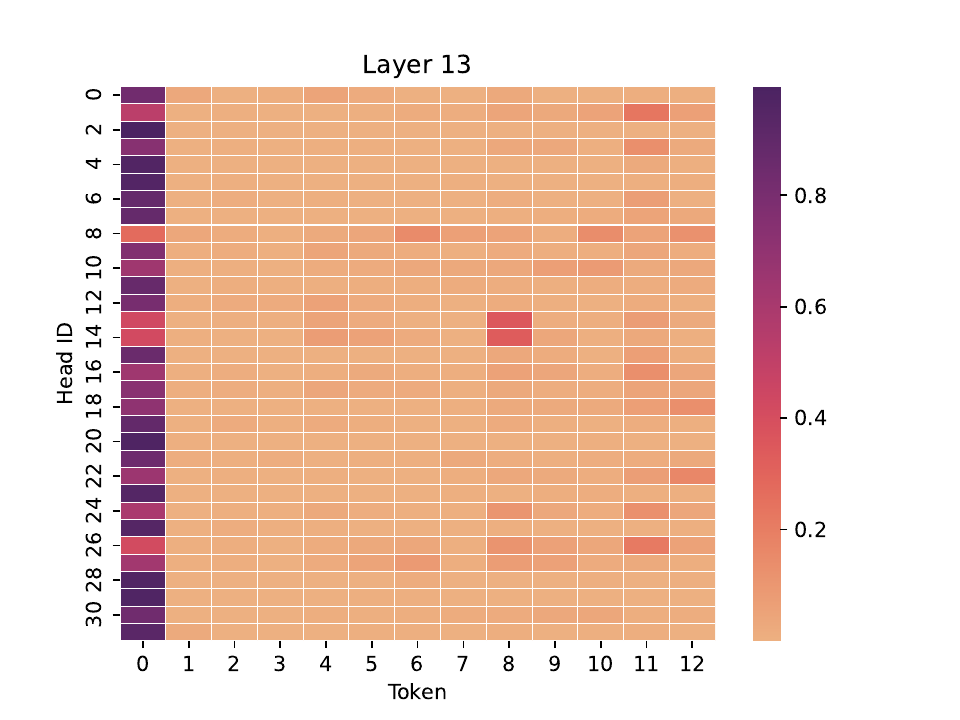}
        \caption{\small{Layer 13}}
    \end{subfigure}
    \begin{subfigure}[t]{0.33\textwidth}
        \includegraphics[width=\textwidth]{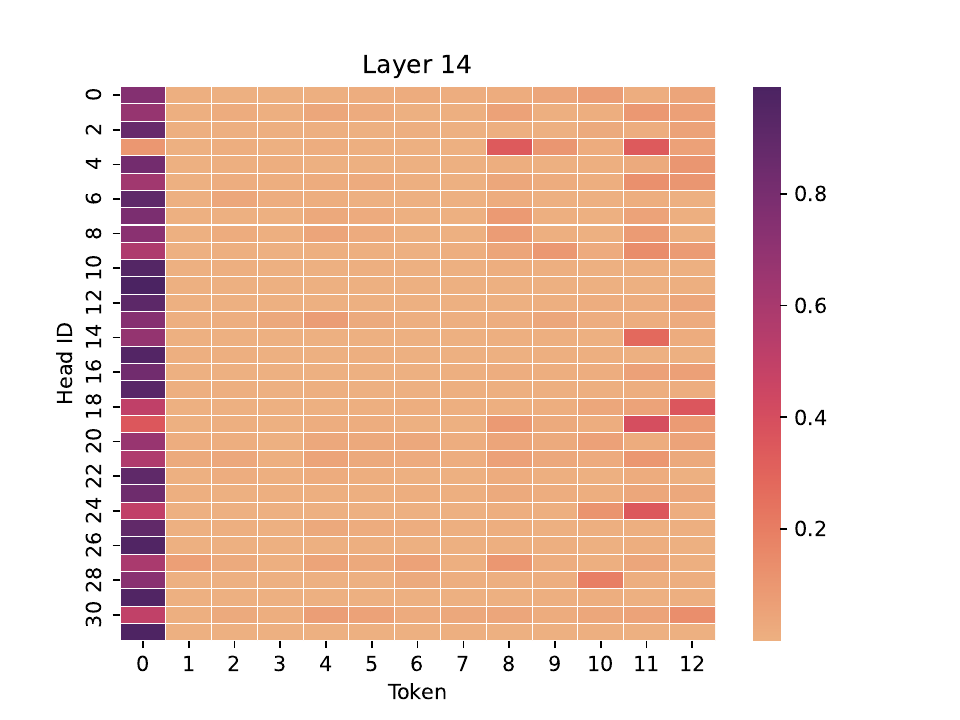}
        \caption{\small{Layer 14}}
    \end{subfigure}
    \end{center}
 \caption{Activations of \llama-7B}
    \end{figure}
    \begin{figure}[t]\ContinuedFloat
          \renewcommand\thesubfigure{\roman{subfigure}}
    \begin{center}
    \begin{subfigure}[t]{0.33\textwidth}
        \includegraphics[width=\textwidth]{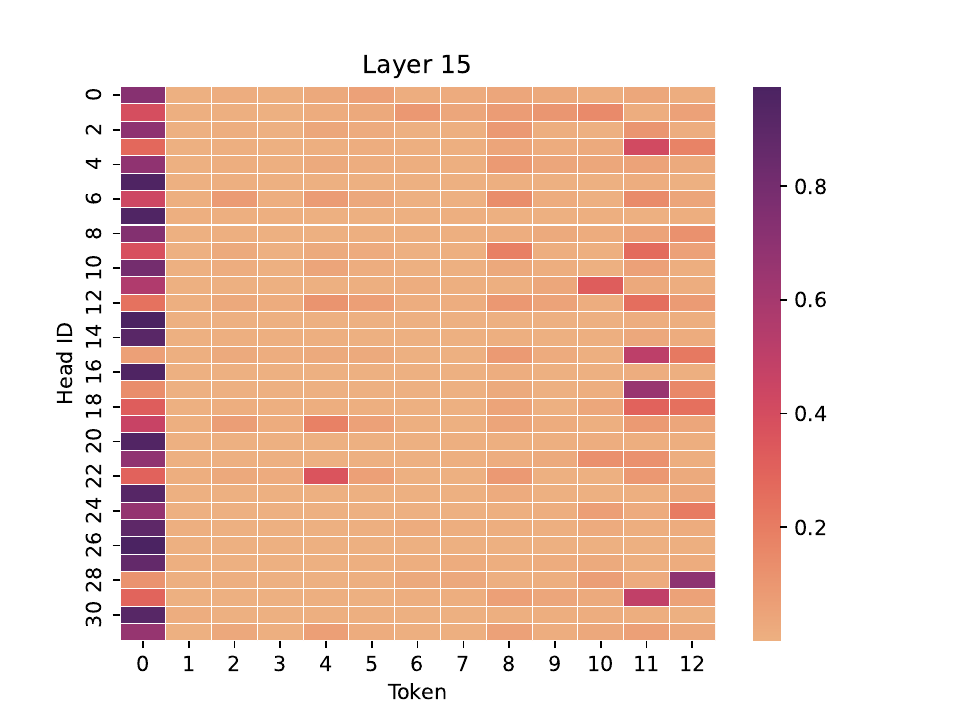}
        \caption{\small{Layer 15}}
    \end{subfigure}
    \begin{subfigure}[t]{0.33\textwidth}
        \includegraphics[width=\textwidth]{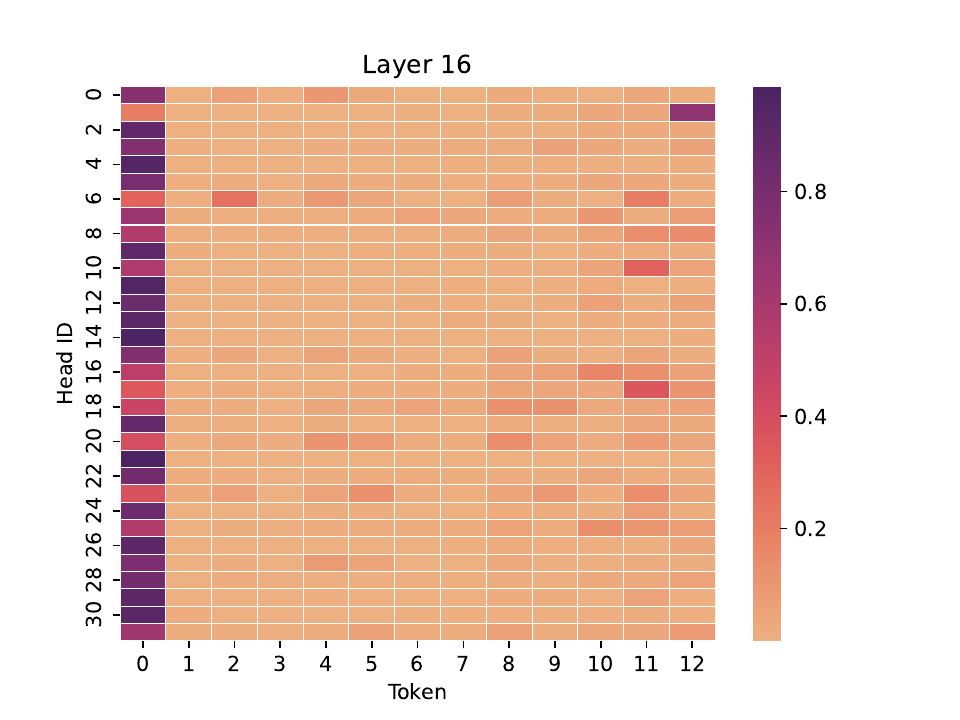}
        \caption{\small{Layer 16}}
    \end{subfigure}
\begin{subfigure}[t]{0.33\textwidth}
        \includegraphics[width=\textwidth]{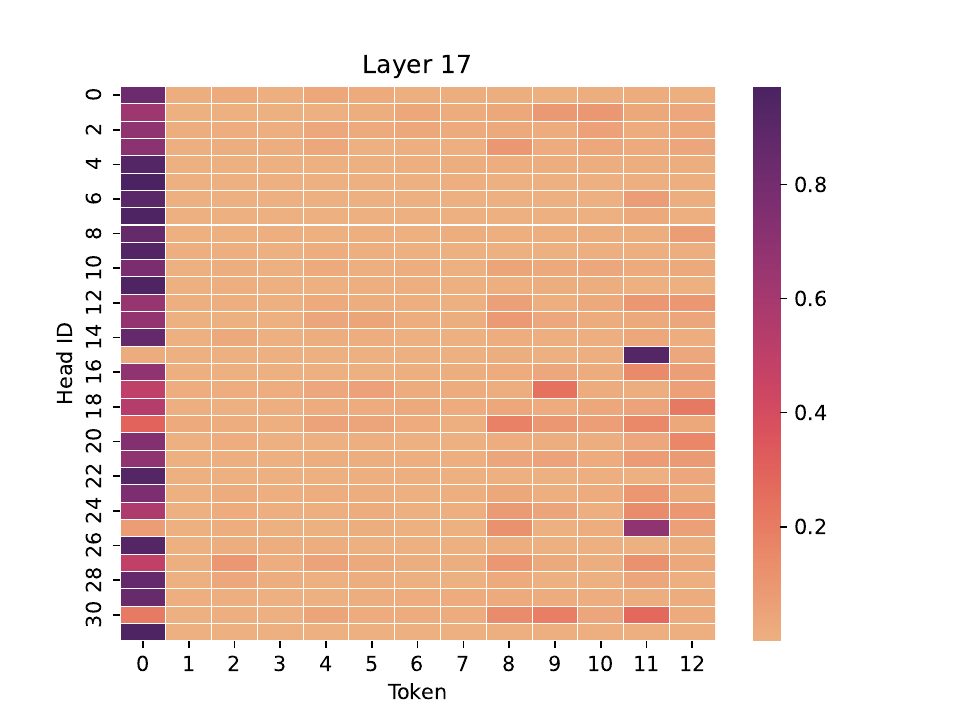}
        \caption{\small{Layer 17}}
    \end{subfigure}
    \begin{subfigure}[t]{0.33\textwidth}
        \includegraphics[width=\textwidth]{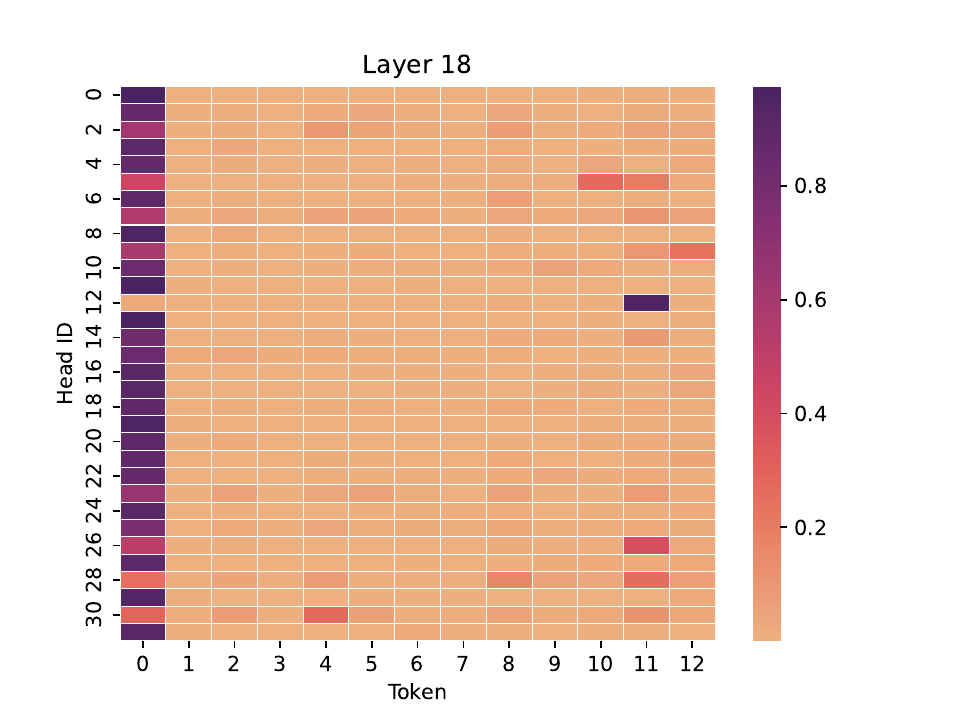}
        \caption{\small{Layer 18}}
    \end{subfigure}
    \begin{subfigure}[t]{0.33\textwidth}
        \includegraphics[width=\textwidth]{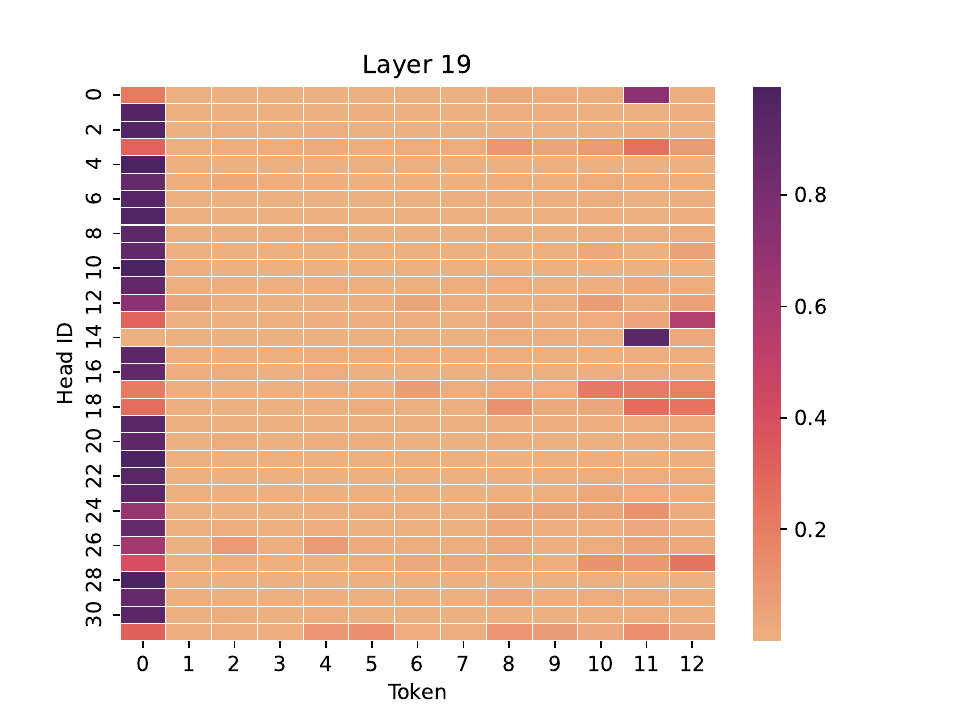}
        \caption{\small{Layer 19}}
    \end{subfigure}
    \begin{subfigure}[t]{0.33\textwidth}
        \includegraphics[width=\textwidth]{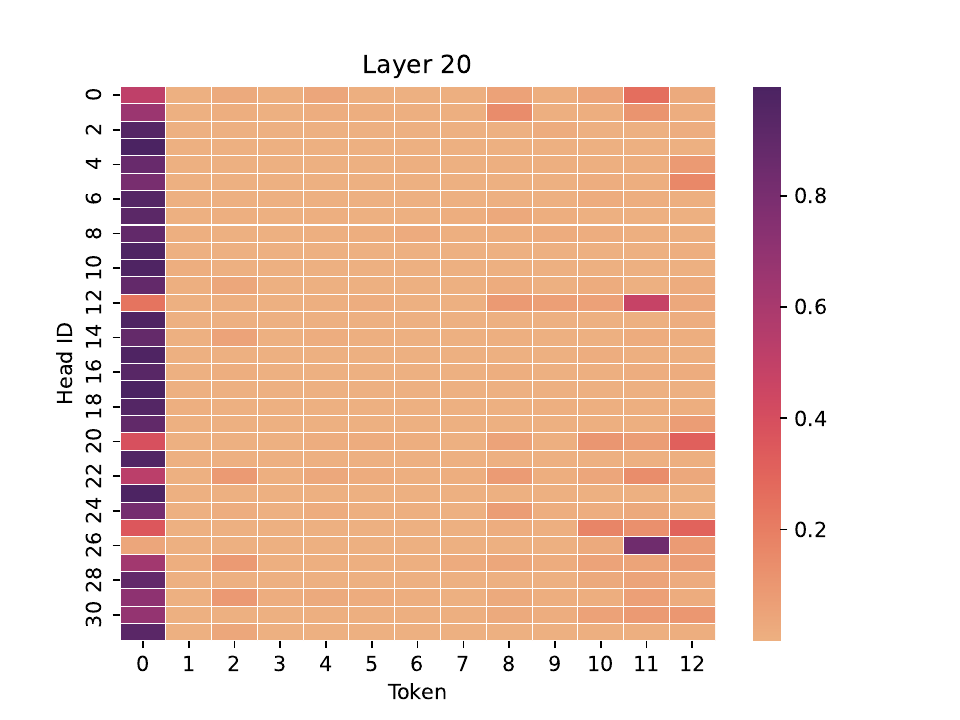}
        \caption{\small{Layer 20}}
    \end{subfigure}    
     \begin{subfigure}[t]{0.33\textwidth}
        \includegraphics[width=\textwidth]{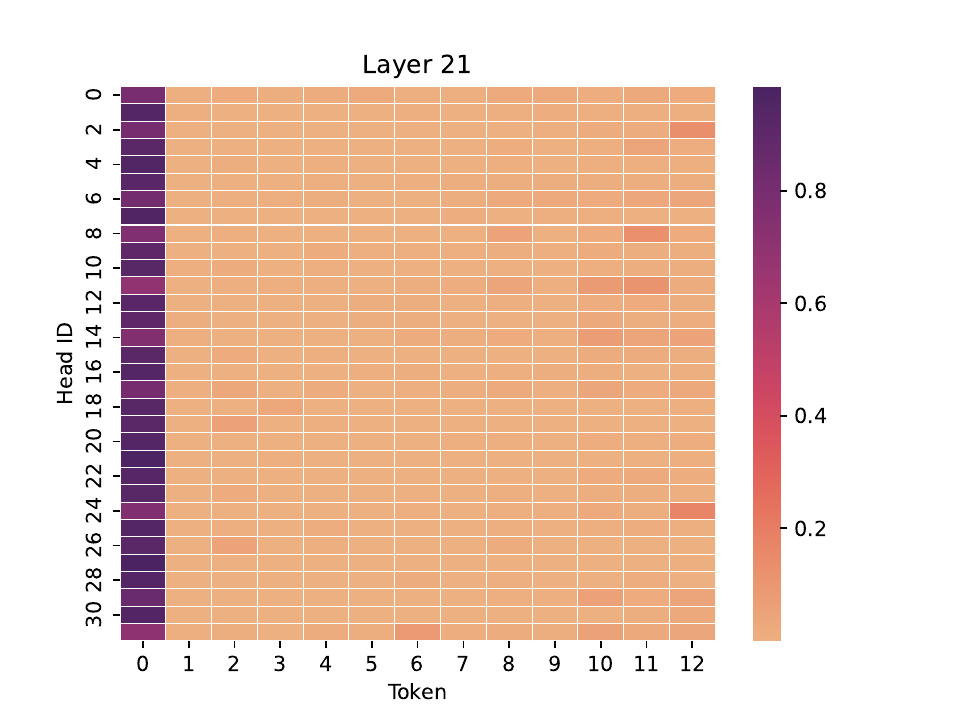}
        \caption{\small{Layer 21}}
    \end{subfigure}
 \begin{subfigure}[t]{0.33\textwidth}
        \includegraphics[width=\textwidth]{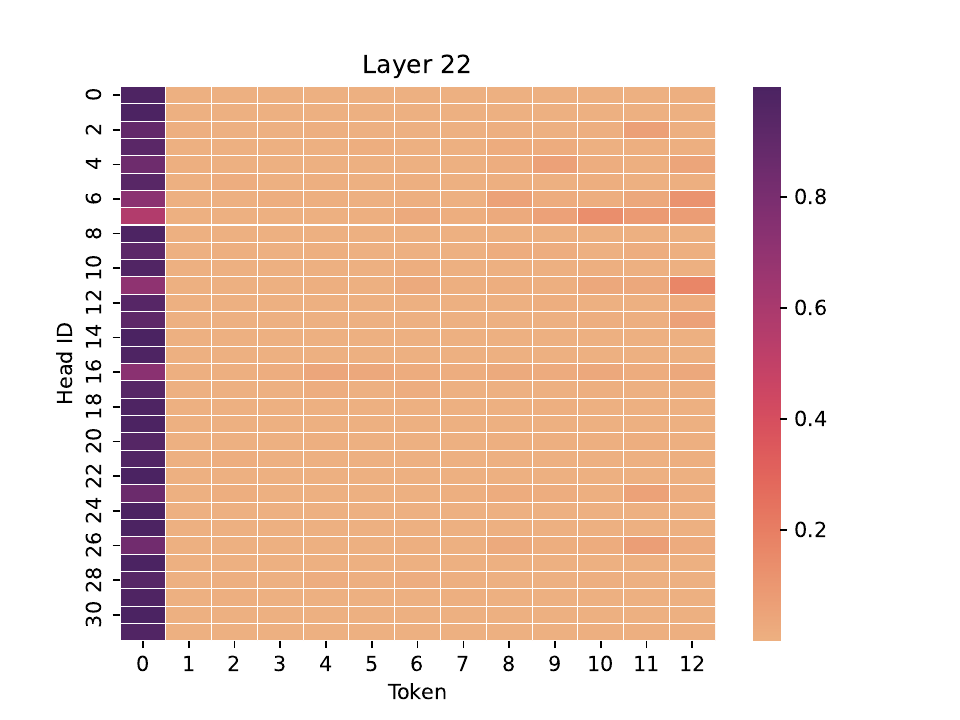}
        \caption{\small{Layer 22}}
    \end{subfigure}
     \begin{subfigure}[t]{0.33\textwidth}
        \includegraphics[width=\textwidth]{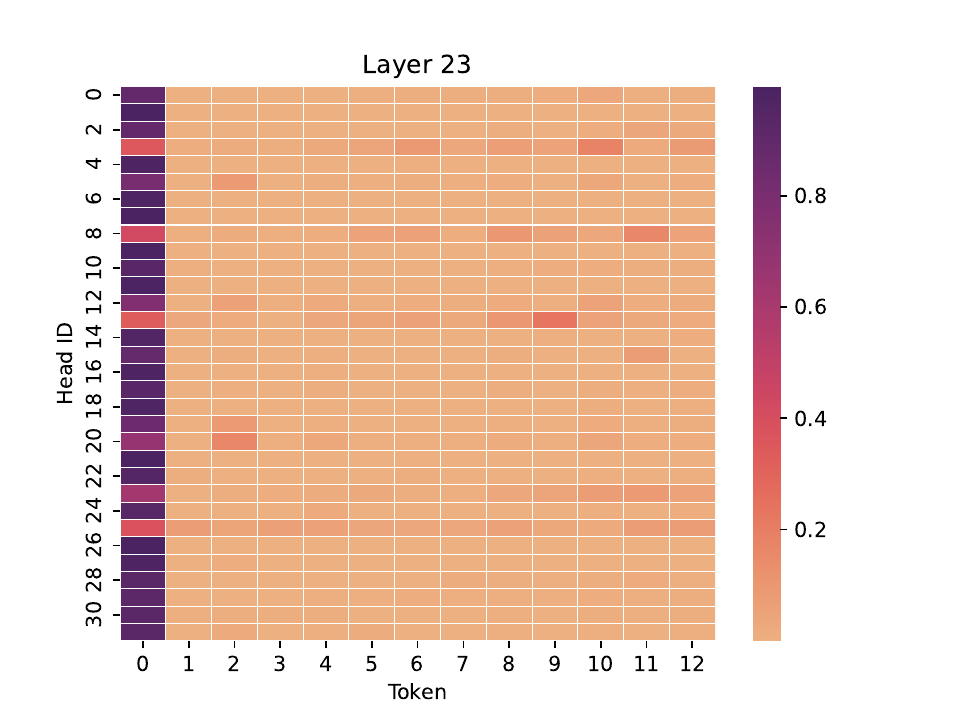}
        \caption{\small{Layer 20}}
    \end{subfigure}
 \begin{subfigure}[t]{0.33\textwidth}
        \includegraphics[width=\textwidth]{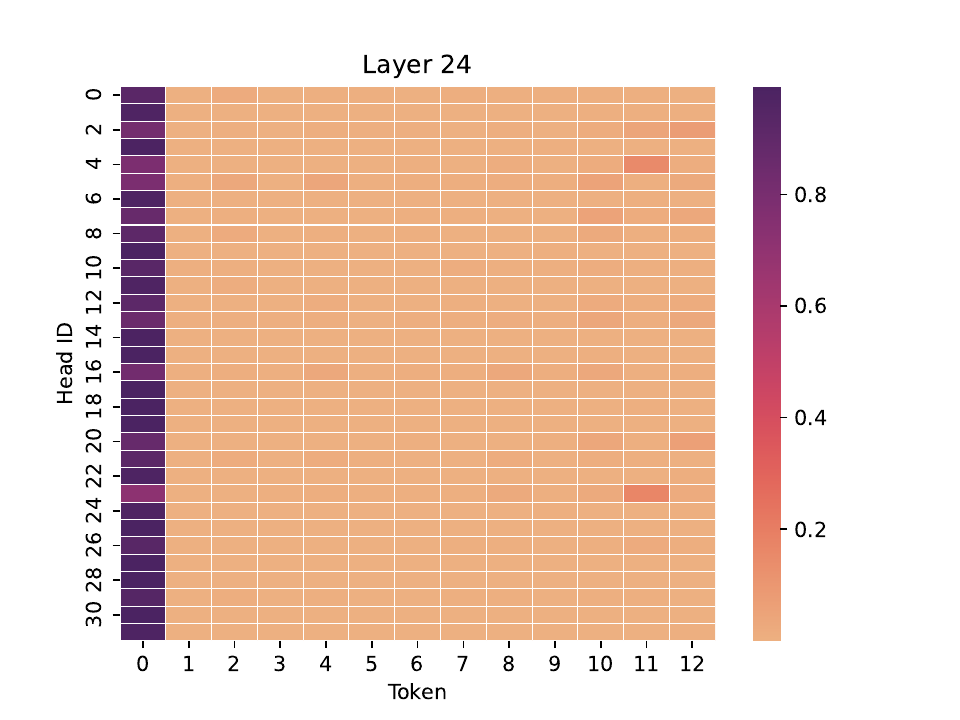}
        \caption{\small{Layer 24}}
    \end{subfigure}
 \begin{subfigure}[t]{0.33\textwidth}
        \includegraphics[width=\textwidth]{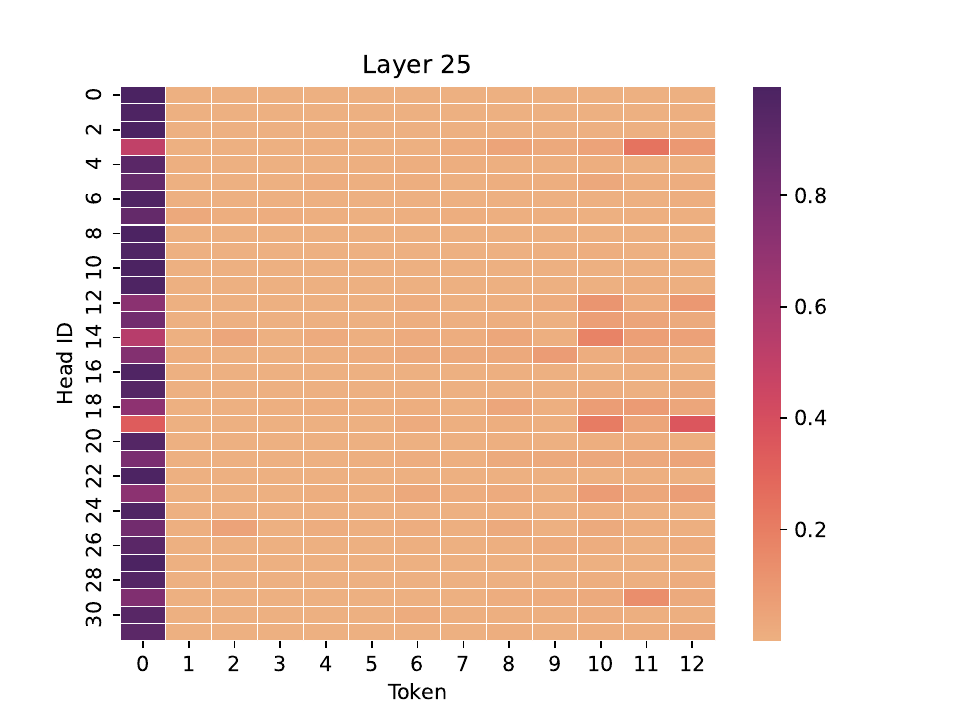}
        \caption{\small{Layer 25}}
    \end{subfigure}
 \begin{subfigure}[t]{0.33\textwidth}
        \includegraphics[width=\textwidth]{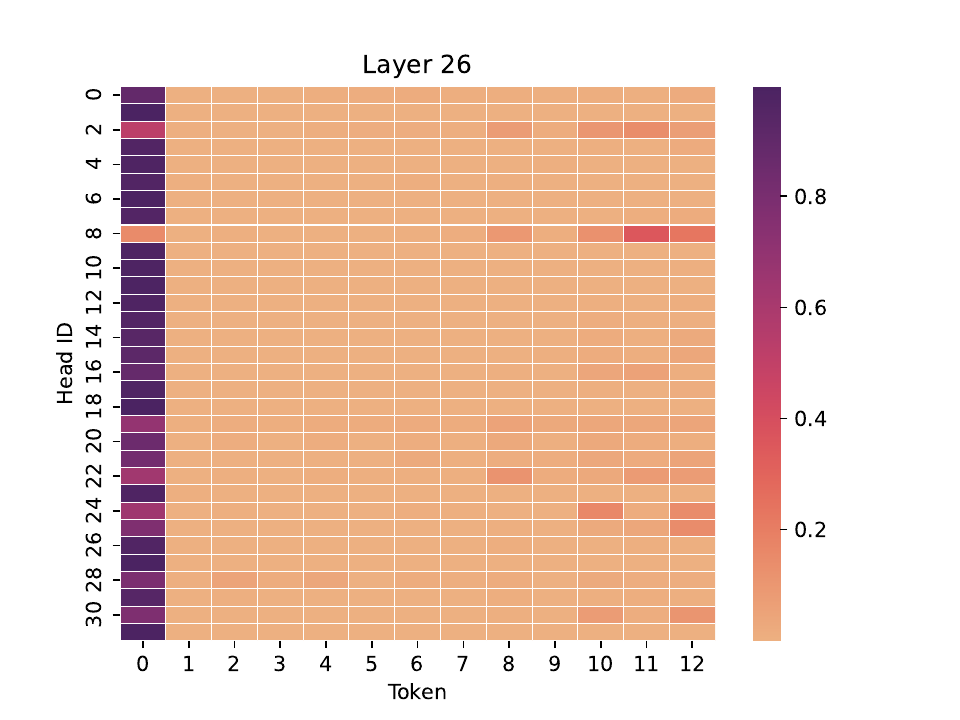}
        \caption{\small{Layer 26}}
    \end{subfigure}   
    \begin{subfigure}[t]{0.33\textwidth}
        \includegraphics[width=\textwidth]{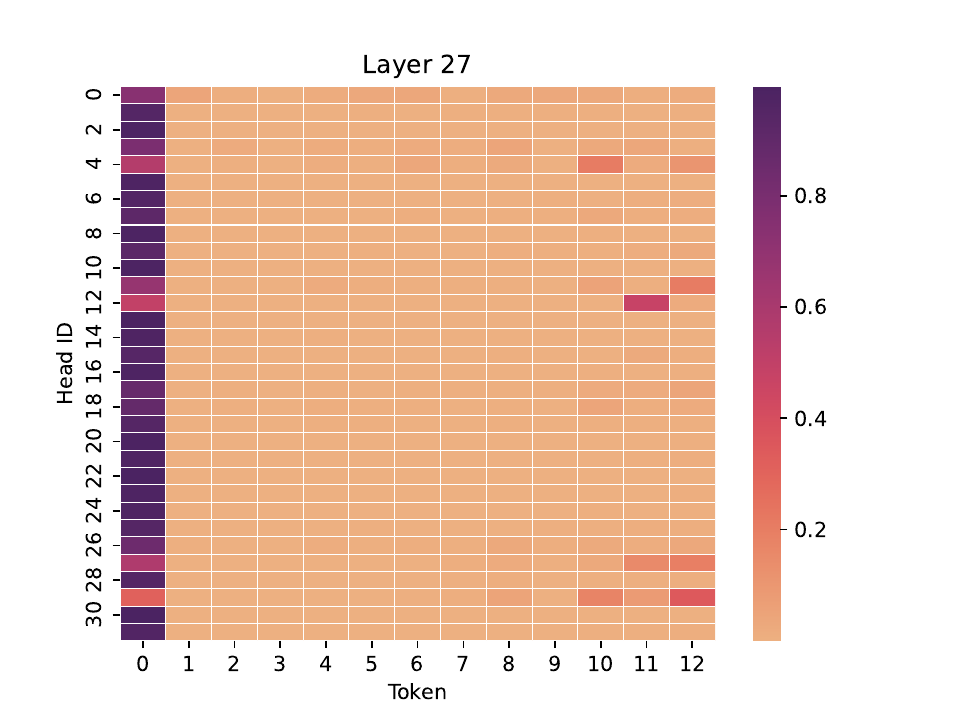}
        \caption{\small{Layer 27}}
    \end{subfigure}
    \begin{subfigure}[t]{0.33\textwidth}
        \includegraphics[width=\textwidth]{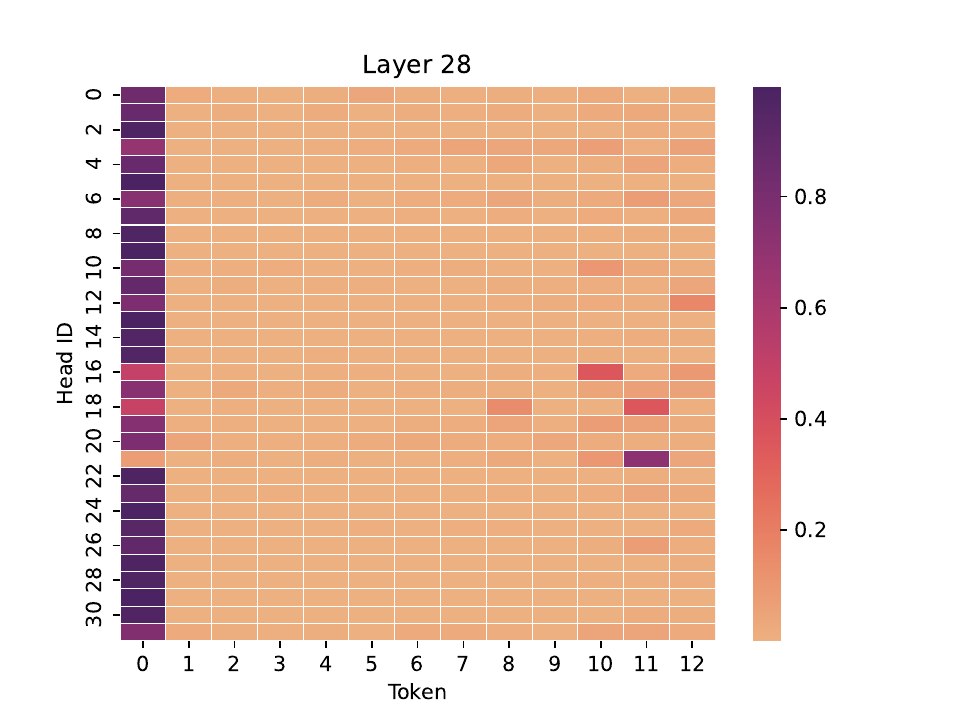}
        \caption{\small{Layer 28}}
    \end{subfigure}
    \begin{subfigure}[t]{0.33\textwidth}
        \includegraphics[width=\textwidth]{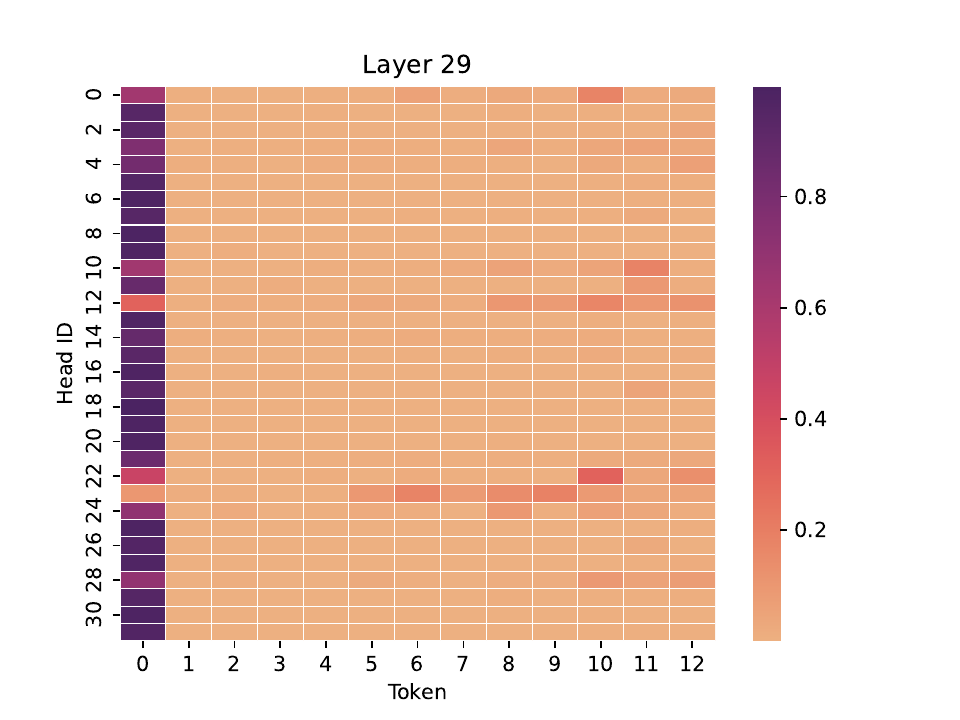}
        \caption{\small{Layer 29}}
    \end{subfigure}
        \end{center}
 \caption{Activations of \llama-7B}
    \end{figure}
    \begin{figure}[t]\ContinuedFloat
          \renewcommand\thesubfigure{\roman{subfigure}}
    \begin{center}
    \begin{subfigure}[t]{0.33\textwidth}
        \includegraphics[width=\textwidth]{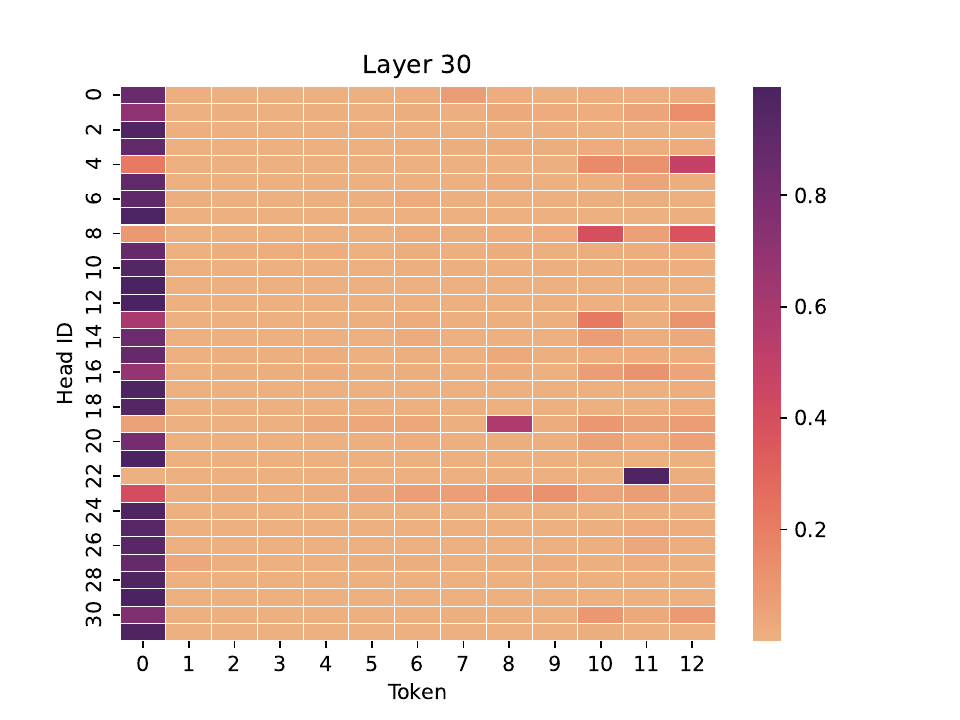}
        \caption{\small{Layer 30}}
    \end{subfigure}
        \begin{subfigure}[t]{0.33\textwidth}
        \includegraphics[width=\textwidth]{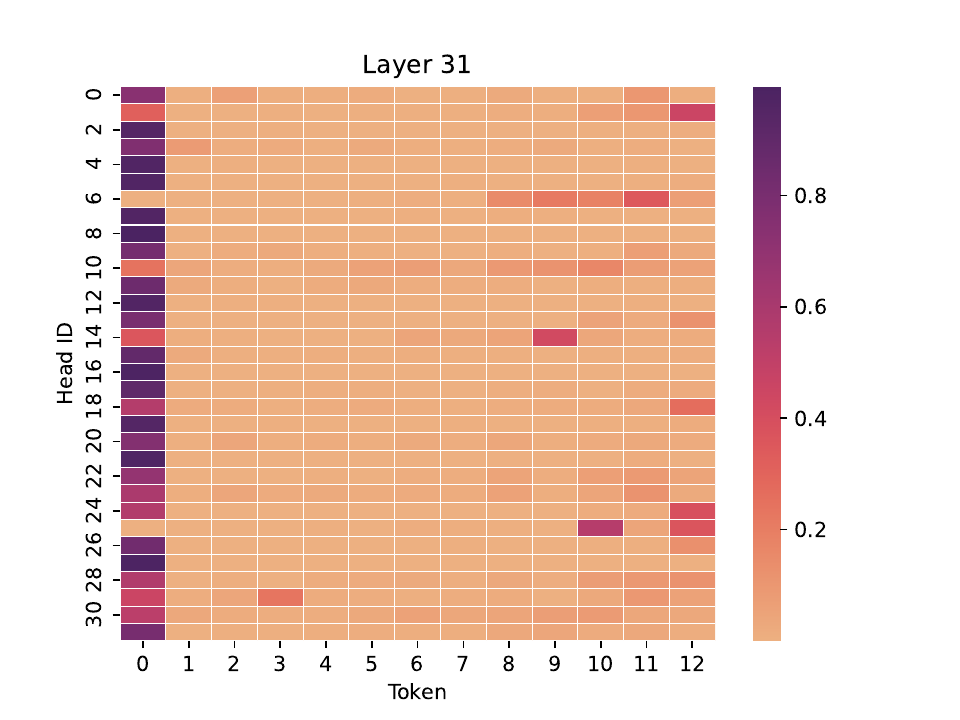}
        \caption{\small{Layer 31}}
    \end{subfigure}
\caption{Activations of \llama-7B}
\end{center}
\end{figure}

\end{document}